\documentclass[%
	paper=A4,	  				
	twoside,		   		    
	openright,					
	parskip=half,				
	chapterprefix=true,			
	10pt,						
	headings=normal,			
	bibliography=totoc,			
	listof=totoc,				
	titlepage=off,				%
	captions=tableabove,		
	draft=false,				
]{scrreprt}%

\newcommand{\thesisTitle}{ZynqNet:\\ An FPGA-Accelerated Embedded\\ Convolutional Neural Network}
\newcommand{\thesisName}{David Gschwend}
\newcommand{\thesisSubject}{Master Thesis}

\usepackage[utf8]{inputenc}		
\usepackage[english]{babel} 

\usepackage[					
	figuresep=colon,%
	sansserif=false,%
	hangfigurecaption=false,%
	hangsection=true,%
	hangsubsection=true,%
	colorize=full,%
	colortheme=bluegreen,%
	bibsys=biber,%
	bibfile=zynqnet_report.bib,%
	bibstyle=ieee,%
]{cleanthesis}
\usepackage{xfrac} 

\usepackage{algorithm}
\usepackage{algpseudocode}

\usepackage{tabularx} 
\usepackage{pdfpages} 
\usepackage{pdflscape} 
\usepackage{booktabs} 
\usepackage{graphicx} 
\usepackage[binary-units=true, detect-all]{siunitx} 
\usepackage{subcaption} 

\usepackage{tikz}
\usetikzlibrary{positioning}

\usepackage{amsmath}
\usepackage{amsfonts}
\usepackage{hyperref}

\hypersetup{					
	pdftitle={\thesisTitle},	
	pdfsubject={\thesisSubject},
	pdfauthor={\thesisName},	
	plainpages=false,			
	colorlinks=false,			
	pdfborder={0 0 0},			
	breaklinks=true,			
	bookmarksnumbered=true,		%
	bookmarksopen=true			%
}

\usepackage{multicol}

\usepackage{rotating}

\usepackage{microtype}

\usepackage{tocloft}
\SetExpansion
 [ context = sloppy,
 stretch = 20,
 shrink = 45,
 step = 1 ]
 { encoding = {OT1,T1,TS1} }
 { }

\usepackage{float}

\usepackage[frozencache]{minted}
\setminted{autogobble,breaklines,fontsize=\small,frame=lines}
\setmintedinline{breaklines=false,fontsize=\normalsize,frame=none}
\def\cppcode#1{\mintinline{cpp}{#1}}

\usepackage[inline]{enumitem}
\setlist{leftmargin=*} 
\setlist[1]{labelindent=\parindent} 
\setlist{itemsep=0.7ex, parsep=0pt} 

\def\cpp{C{}\texttt{++}\xspace}

\def\caffe{Caffe\xspace}

\def\hin{\ensuremath{h_\textrm{in}}\xspace}
\def\win{\ensuremath{w_\textrm{in}}\xspace}
\def\chin{\ensuremath{ch_\textrm{in}}\xspace}
\def\hout{\ensuremath{h_\textrm{out}}\xspace}
\def\wout{\ensuremath{w_\textrm{out}}\xspace}
\def\chout{\ensuremath{ch_\textrm{out}}\xspace}
\def\npe{\ensuremath{N_\textrm{PE}}\xspace}

\def\x{$\times$}

\usepackage[title,titletoc]{appendix}

\usepackage{cleveref}
\crefname{appsec}{appendix}{appendices}

\KOMAoptions{listof=totoc}

\makeatletter
\newcommand*{\algrule}[1][\algorithmicindent]{%
  \makebox[#1][l]{%
    \hspace*{.3em}
    \vrule height .75\baselineskip depth .25\baselineskip
  }
  \hspace*{-.6em}
}

\newcount\ALG@printindent@tempcnta
\def\ALG@printindent{%
    \ifnum \theALG@nested>0
    \ifx\ALG@text\ALG@x@notext
    \addvspace{-3pt}
    \else
    \unskip
    \ALG@printindent@tempcnta=1
    \loop
    \algrule[\csname ALG@ind@\the\ALG@printindent@tempcnta\endcsname]%
    \advance \ALG@printindent@tempcnta 1
    \ifnum \ALG@printindent@tempcnta<\numexpr\theALG@nested+1\relax
    \repeat
    \fi
    \fi
}
\patchcmd{\ALG@doentity}{\noindent\hskip\ALG@tlm}{\ALG@printindent}{}{\errmessage{failed to patch}}
\patchcmd{\ALG@doentity}{\item[]\nointerlineskip}{}{}{} 
\makeatother

\begin{document}


\pagenumbering{roman}			
\pagestyle{empty}				

\begin{titlepage}
	\pdfbookmark[0]{Titlepage}{Titlepage}
	\tgherosfont

 	\begin{center}
		\begin{minipage}[b]{0.45\linewidth}
			\begin{center}
 				\vspace{0pt}	
 				\includegraphics[width=0.8\linewidth]{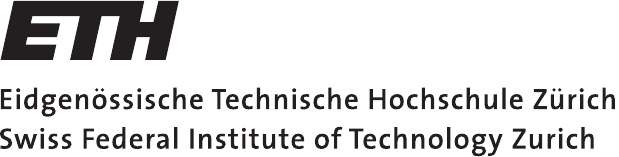}
				\hspace{1cm}
			\end{center}
		\end{minipage}
		\hfill
		\begin{minipage}{0.45\textwidth}
			\begin{center}
			\vspace{-1.4cm}
			\hspace{1cm}
 				\includegraphics[width=0.8\linewidth]{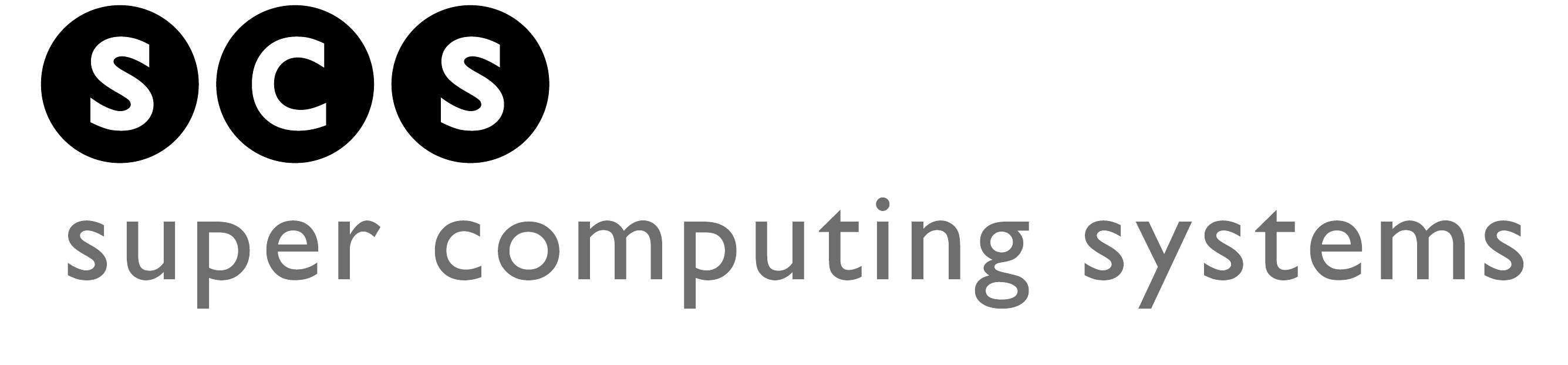}
			\end{center}
		\end{minipage}
		
		\vspace{0.1cm}
		\hspace*{0.15cm}\rule{0.985\textwidth}{0.4pt}
		\vspace{0.1cm}
				
		{Master Thesis Report}
		
		\vfill

		{\Huge\textbf{ZynqNet:\\[6pt] An FPGA-Accelerated Embedded\\[10pt] Convolutional Neural Network}}

		\vfill
		
		\includegraphics[width=0.94\linewidth]{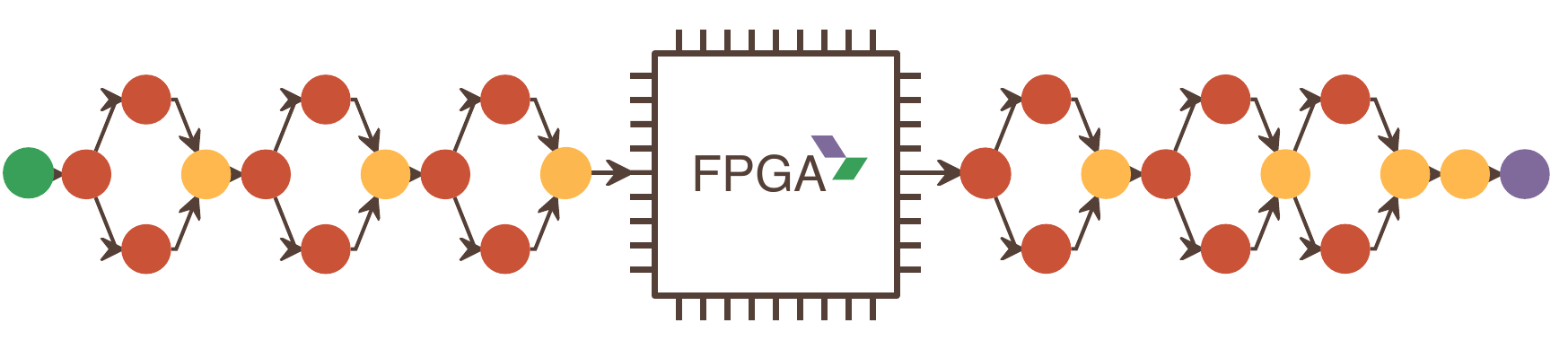}
		
		\vfill
		
		{\Large David Gschwend}\\
		{dgschwend@gmail.com}
		
		\vfill
		
		\begin{tabular}{ll}
		 Supervisors: & Emanuel Schmid \\
		& Felix Eberli \\
		 \rule{0pt}{3ex}Professor: & Prof.\ Dr.\ Anton Gunzinger \\
		\end{tabular}\\[2mm]
		
		\vfill

		{August 2016, ETH Zürich,}\\[2mm]
		{Department of Information Technology and Electrical Engineering} \\[2mm]

 \end{center}

\end{titlepage}
\cleardoublepage

\pagestyle{plain}				

\chapter*{Abstract}


Image Understanding is becoming a vital feature in ever more applications ranging from medical diagnostics to autonomous vehicles. Many applications demand for embedded solutions that integrate into existing systems with tight real-time and power constraints.
Convolutional Neural Networks (CNNs) presently achieve record-breaking accuracies in all image understanding benchmarks, but have a very high computational complexity.
Embedded CNNs thus call for small and efficient, yet very powerful computing platforms.

This master thesis explores the potential of FPGA-based CNN acceleration and demonstrates a fully functional proof-of-concept CNN implementation on a Zynq System-on-Chip.
The \emph{ZynqNet Embedded CNN} is designed for image classification on ImageNet and consists of \emph{ZynqNet CNN}, an optimized and customized CNN topology, and the \emph{ZynqNet FPGA Accelerator}, an FPGA-based architecture for its evaluation.

\emph{ZynqNet CNN} is a highly efficient CNN topology. Detailed analysis and optimization of prior topologies using the custom-designed \emph{Netscope CNN Analyzer} have enabled a CNN with \SI{84.5}{\%} top-5 accuracy at a computational complexity of only \num{530} million multiply-accumulate operations. The topology is highly regular and consists exclusively of convolutional layers, ReLU nonlinearities and one global pooling layer. The CNN fits ideally onto the FPGA accelerator.

The \emph{ZynqNet FPGA Accelerator} allows an efficient evaluation of ZynqNet CNN. It accelerates the full network based on a nested-loop algorithm which minimizes the number of arithmetic operations and memory accesses.
The FPGA accelerator has been synthesized using High-Level Synthesis for the Xilinx Zynq XC-7Z045, and reaches a clock frequency of \SI{200}{\MHz} with a device utilization of \SI{80}{\%} to \SI{90}{\%}.

\vspace{0.8cm}

\paragraph*{Organization of this report}
\Cref{chap:introduction} gives an overview of the current opportunities and challenges regarding {image understanding in embedded systems}. The following \cref{chap:background} introduces the central concepts of \emph{Convolutional Neural Networks} (CNNs) and \emph{Field-Programmable Gate Arrays} (FPGAs), as well as a number of CNN topologies and CNN accelerators from prior work. \Cref{chap:cnn-design} dives deep into the analysis, training and optimization of CNN architectures, and presents our customized \emph{ZynqNet CNN} topology. Next, \cref{chap:fpga-design} shifts the focus onto the design and implementation of our FPGA-based architecture for the evaluation of CNNs, the \emph{ZynqNet FPGA Accelerator}, and reports lessons learned from the application of High-Level Synthesis. Finally, \cref{chap:results} presents the performance results of the overall \emph{ZynqNet Embedded CNN} system, before the conclusion in \cref{chap:conclusion} puts these in a bigger perspective.
\cleardoublepage

%
%
\pdfbookmark[0]{Acknowledgement}{Acknowledgement}
\chapter*{Acknowledgement}
\label{sec:acknowledgement}

First and foremost, I would like to thank my supervisor Emanuel Schmid for the pleasant collaboration, the fruitful discussions, the helpful guidance and his excellent support during the project. You offered me full confidence and freedom, yet were always there when I needed feedback, a different point of view or new ideas. I also thank Felix Eberli for arranging this project, for involving me in various interesting meetings and discussions, and for his generous support.

Special thanks also go to professor Dr. Anton Gunzinger for giving me the chance to work on a fascinating project of practical relevance, and to the whole staff at Supercomputing Systems AG for the warm welcome and the pleasant stay.

Finally, I want to express my gratitude to my family, my friends and my fiancée. You've always had my back, and I could not have made it here without your constant and unconditional support. Thank you. 
\cleardoublepage

\setcounter{tocdepth}{2}	 	
\tableofcontents				

\cleardoublepage

\listoffigures 

\clearpage

\listoftables

\cleardoublepage

\pagenumbering{arabic}			
\setcounter{page}{1}			
\pagestyle{maincontentstyle} 	

\chapter{Introduction}
\label{chap:introduction}

\cleanchapterquote{It is clear that humans will soon outperform state-of-the-art image classification models only by use of significant effort, expertise, and time.}{Andrej Karpathy}{(Deep Learning Expert, OpenAI)}

%
\section{Motivation}






Image understanding is a very difficult task for computers. Nevertheless, advanced Computer Vision (CV) systems capable of {image classification}, {object recognition} and {scene labeling} are becoming increasingly important in many applications in robotics, surveillance, smart factories and medical diagnostics. Unmanned aerial vehicles and autonomous cars, which need to perceive their surroundings, are further key applications.

In the last few years, significant progress has been made regarding the performance of these advanced CV systems. The availability of powerful computing platforms and the strong market pull have shaped a very fast-paced and dynamic field of research. Former approaches to image understanding, which mainly relied on hand-engineered features and hard-coded algorithms, are increasingly being replaced by \emph{machine learning} concepts, where computers learn to understand images by looking at thousands of examples. These advanced learning algorithms, which are based on recent high-performance computing platforms as well as the abundance of training data available today, are commonly referred to as \emph{deep learning}.

\emph{Convolutional Neural Networks} (CNNs) currently represent the most promising approach to image understanding in CV systems. These brain-inspired algorithms
consist of multiple
layers of feature detectors and classifiers, which are adapted and optimized using techniques from machine learning~\cite{braininspired}.
The idea of neural networks has been around for almost 80 years~\cite{firstNN}, yet only the latest generations of high-performance computing hardware have allowed the evaluation and training of CNNs deep and wide enough for good performance in image understanding applications. 
The progress in these last years has been amazing though, and state-of-the-art convolutional neural networks already rival the accuracy of humans when it comes to the classification of images~\cite{ilsvrc_human}.

This exceptional performance of CNNs comes at the cost of an enormous computational complexity. The real-time evaluation of a CNN for image classification on a live video stream can require billions or trillions of operations per second. The effort for image segmentation and scene labeling is even significantly higher.
While this level of performance can be reached with the most recent Graphics Processing Units (GPUs), there is the simultaneous wish to embed such solutions into other systems, such as cars, drones, or even wearable devices, which exhibit strict limitations regarding physical size and energy consumption.
Future embedded CNNs thus call for small and efficient, yet very powerful computing platforms.

Different platforms have been considered for efficient high-performance implementations of CNNs, and \emph{Field-Programmable Gate Arrays} (FPGAs) are among the most promising of them. These versatile integrated circuits provide hundreds of thousands of programmable logic blocks and a configurable interconnect, which enables the construction of custom-tailored accelerator architectures in hardware.
These have the potential to deliver the computational power required by embedded CNNs within the size and power envelopes dictated by their respective applications.

\section{Contribution}

Initially, this master aimed to explore, benchmark and optimize one or more commercial approaches to the acceleration of convolutional neural networks on FPGAs, with a focus on embedded systems. Multiple FPGA and intellectual property vendors have announced frameworks and libraries that target the acceleration of deep learning systems.%
\footnote{%
The following commercial frameworks and libraries target the acceleration of CNNs using FPGAs:
\begin{itemize}[labelindent=2mm,leftmargin=*]
\item Auviz Systems AuvizDNN Framework \cite{auviz-info,auviz-info2,auviz-xcell}
\item Falcon Computing Solutions machine learning libraries based on OpenCL  \cite{falcon-cnn}
\item MulticoreWare machine learning libraries based on SDAccel  \cite{multicoreware-cnn}
\end{itemize}
Additionally, Altera OpenCL \cite{altera-cnn} and Xilinx SDAccel \cite{xilinx-sdaccel} are generic frameworks which allow computation kernels to be offloaded from a host processor onto FPGA-based accelerators. However, these frameworks do not directly accelerate CNNs and were therefore not considered ready-to-use, although both companies mention the acceleration of deep learning algorithms as a major use case.
}
However, none of these solutions turned out to be ready and available for testing.

Nevertheless, we decided to further pursue this promising approach by building our own proof-of-concept FPGA-based CNN implementation from scratch, with a special focus on the optimized co-operation between the underlying hardware architecture and the convolutional neural network. The result is the \emph{ZynqNet Embedded CNN}, an FPGA-based convolutional neural network for image classification. The solution consists of two main components:
\begin{enumerate}
	\item The \emph{ZynqNet CNN}, a customized convolutional neural network topology, specifically shaped to fit ideally onto the FPGA. The CNN is exceptionally regular, and reaches a satisfying classification accuracy with minimal computational effort.
	\item The \emph{ZynqNet FPGA Accelerator}, a specialized FPGA architecture for the efficient acceleration of ZynqNet CNN and similar convolutional neural networks.
\end{enumerate}
ZynqNet CNN is trained offline on GPUs using the \caffe framework, while the ZynqNet FPGA Accelerator employs the CNN for image classification, or \emph{inference}, on a Xilinx Zynq XC-7Z045 System-on-Chip (SoC).
Both components have been developed and optimized within the six month time frame of this master thesis, and together constitute a fully functional convolutional neural network implementation on the small and low-power Zynq platform.

This report documents the ZynqNet CNN and the ZynqNet FPGA Accelerator and gives insight into their development.
In addition, the \emph{Netscope CNN Analyzer} is introduced, a custom tool for visualizing, analyzing and editing convolutional neural network topologies. Netscope has been used to analyze a number of different CNN architectures, and the findings are presented in the form of a \emph{Design Space Exploration} (DSE) of CNN topologies from prior work.
Finally, the performance of the ZynqNet Embedded CNN is evaluated and its performance is compared to other platforms.

\chapter{Background and Concepts}
\label{chap:background}

\cleanchapterquote{If I have seen further than others, it is by standing upon the shoulders of giants.}{Isaac Newton}{}

This chapter introduces two of the main concepts behind this thesis: Convolutional Neural Networks (CNNs, \cref{sec:cnn}) and Field-Programmable Gate Arrays (FPGAs, \cref{sec:fpga-intro}).
In addition, we present a number of CNN topologies (\cref{sec:cnn-topologies}) and embedded CNN implementations from prior work (\cref{sec:prior-work}), before the final section compiles the requirements and specifications for our own FPGA-accelerated embedded CNN (\cref{sec:project-specs}).

%
\section{Convolutional Neural Networks}
\label{sec:cnn}

The following sections give a brief overview of neural networks in general, and of convolutional neural networks in particular. First, an intuitive explanation for the inner workings of neural networks is presented, including a high-level description of the network training process (\cref{sec:neural-network-intro}). The next section dives into convolutional neural networks, an architecture particularly suited for processing images, and gives an overview of the construction of these networks (\cref{sec:cnn-intro}).
Finally, the most important CNN topologies for image classification are introduced and characterized (\cref{sec:cnn-topologies}).

For a more conclusive introduction to this rapidly expanding field, the excellent course \emph{CS231n: Convolutional Neural Networks for Visual Recognition} by Andrej Karpathy is highly recommended and publicly available \cite{cs231n}. Further starting points include the \emph{Deep Learning Book} by Goodfellow, Bengio et al. \cite{dlbook}, the online course by Nielsen \cite{mnielsen} as well as the \caffe tutorials \cite{caffe_tutorials}.

%
\subsection{Introduction to Neural Networks}
\label{sec:neural-network-intro}

\paragraph{Biological Inspiration} 
\label{par:inspiration}
Neural networks are a family of computation architectures originally inspired by biological nervous systems. The human brain contains approximately 86 billion neurons connected by $10^{14}$--$10^{15}$ synapses. Each neuron receives input signals at its dendrites and produces output signals along its axon, which branches out and connects to the dendrites of other neurons via synapses. These synapses influence the transfer of information from one neuron to the other by amplifying or attenuating the signals, or even inhibiting the signal transfer at other synapses. Together, the billions of conceptually simple neurons form an incredibly complex interacting network which enables us humans to see, hear, move, communicate, remember, analyze, understand and even to fantasize and dream~\cite{brain_neurons,cs231n}.

\begin{figure}[tb]
  \centering
  \includegraphics[width=0.44\linewidth]{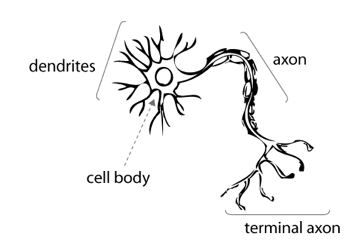}\quad
  \includegraphics[trim=0 -8mm 0 0, width=0.44\linewidth]{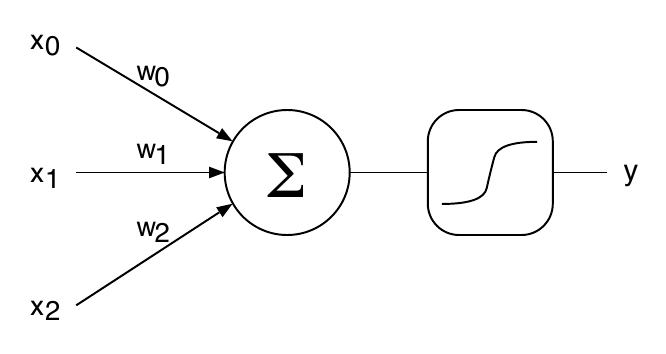}
  \caption[Illustration of Biological and Aritificial Neurons]{A Biological Neuron and its Artificial Counterpart. (Image adapted from \cite{intechopen})}
  \label{fig:artificial-neuron}
\end{figure}

\paragraph{Artificial Neurons} 
\label{par:artificial_neurons}

The basic building block in artificial neural networks is the \emph{artificial neuron}, depicted in \cref{fig:artificial-neuron}. The artificial neuron receives a number of input signals $x_i$ from other neurons. These input signals are multiplied with weights $w_i$ to simulate the synaptic interaction at the dendrites. The weighed input signals are summed up, biased with a fixed $w_b$ and fed into a non-linear \emph{activation function}, which produces the neuron's output signal $y = f\left(\sum \left[ x_i\cdot w_i \right] + w_b \right)$ ~\cite{cs231n}. The weights $w$ can be seen as the tuning knobs that define the neuron's reaction to a given input signal, and their values can be adjusted in order to learn to approximate a desired output signal~\cite{cs231n,origami_sa}.\footnote{A single artificial neuron can naturally only give a simple approximation. Interestingly, already a two-layer neural network is a \emph{universal approximator} that can approximate any continuous functions to an arbitrary degree of precision using a finite amount of neurons. An intuitive explanation for this theorem is given in~\cite[ch. 4]{mnielsen}, which also justifies the application of a non-linear activation function.}

\paragraph{Neural Network Organization} 
\label{par:network_organization}

\begin{figure}[tb]
  \centering
  \includegraphics[width=0.5\linewidth]{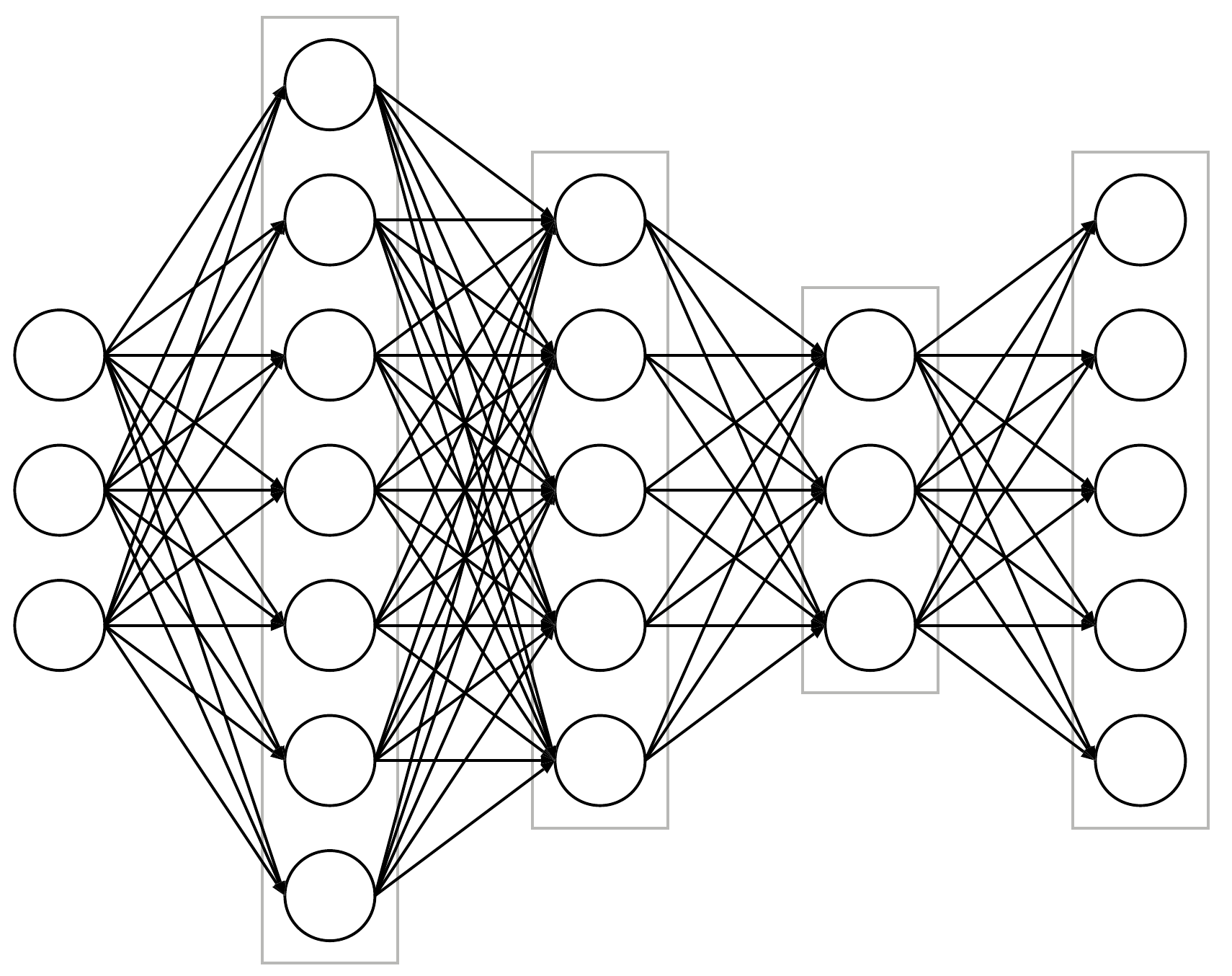}
  \caption[Illustration of a Fully Connected Neural Network]{Example of a Neural Network with 4 Fully-Connected Layers, 3 Inputs and 5 Outputs.}
  \label{fig:feedforward-neural-net}
\end{figure}

A \emph{neural network} is formed by interconnecting many artificial neurons. Usually, the neurons are arranged in a directed acyclic graph to form a \emph{feed-forward neural network}.\footnote{Neural networks with directed cycles in their structure are called \emph{Recurrent Neural Networks} (RNNs) and play an important role in speech recognition, text-to-speech synthesis and natural language processing. Thanks to their inherent memory, they are well suited for processing time-dependent signals. However, RNNs are usually difficult to train and have problems with scaling. Some of these difficulties can be mitigated by using \emph{Long Short-Term Memory} (LSTM) networks, which are currently the most popular type of RNNs~\cite{lstm_paper,lstm_acoustics}.}
The neurons are further grouped into layers, and connections are only allowed between neurons of adjacent layers. \Cref{fig:feedforward-neural-net} shows an example of a four-layer feed-forward neural network with fully-connected layers%
\footnote{In a fully-connected layer, each output from the previous layer is connected to every neuron in the current layer. A feed-forward neural network consisting of fully-connected layers is also called \emph{Multilayer Perceptron} (MLP)~\cite{cs231n}.} and five outputs.


\paragraph{Network Training} 
\label{par:network_training}
The parameters in a neural network are not manually chosen, but \emph{learned} during a training phase. The most popular training approach is called \emph{supervised learning} and requires a set of \emph{labeled training examples}.
One optimization pass through all training examples is called a \emph{training epoch}. Depending on the type of data and the capacity of the neural network, a complete training session can take anywhere from one to a few hundred epochs.
The training starts with small, randomly initialized weights.\footnote{Initialization with a constant (e.g. zero) would make all neurons compute exactly the same outputs and would prevent any learning. The exact initialization strategy can be quite important.%
}
One by one, the examples are fed through the network (so-called \emph{forward pass}). The resulting outputs are compared to the \emph{ground truth labels} using a \emph{loss function}, which measures how much the output deviates from the expected result. The goal of the learning process is then to minimize this loss (or error) on the training set by optimizing the weight parameters.
\emph{Stochastic Gradient Descent} is the most popular optimization method currently used for training neural networks. The gradient descent algorithm computes a gradient vector that describes each weight's influence on the error. These gradients can be efficiently calculated by \emph{backpropagation of the output error} through the network (so-called \emph{backward pass}).
The optimization loop repeatedly takes a training example, calculates the current loss (forward pass), derives the gradient vector (backward pass), and adjusts all weights by a small amount in the opposite direction of their respective gradient (update phase).
The magnitude of these updates is determined by the so-called \emph{learning rate}.
An alternate version of the algorithm, called \emph{Batch Gradient Descent}, defers the weight updates, and first computes and averages the gradients of a \emph{batch} of training examples.
This allows the computation to be vectorized and executed more efficiently on platforms which support vector instructions, including GPUs, DSPs and most CPUs~\cite{dl_nature,dlbook,cs231n}.

\paragraph{Performance Validation} By iteratively adjusting the weights, the network ideally converges towards a solution with minimal loss and thus with a good approximation of the desired output on the training set.
Every few epochs, the performance of the model is verified with an array of \emph{validation examples} which were not used during training.\footnote{The validation examples are usually a fraction of the labeled examples which are set aside from the beginning. It is common to use around \SIrange{20}{25}{\%} of the labeled examples as validation set.}
If the training set is representative for the actual ``real-world'' data, the network also delivers good estimations for previously unseen examples. If however the training set is too small, or the network's learning capacity too high, the neural network can memorize examples ``by heart'' and lose its ability to generalize. Such \emph{overfitting} can be counteracted with enlarged training sets (possibly using \emph{data augmentation} strategies such as mirroring, rotation and color transformations) as well as changes to the network structure (such as the addition of regularization methods)~\cite{cs231n}.

%
\subsection{Introduction to Convolutional Neural Networks}
\label{sec:cnn-intro}

Convolutional Neural Networks (CNNs) are a special class of neural networks particularly suited for operation on 2D input data such as images. They are widely used for \emph{image classification}, \emph{object recognition} and \emph{scene labeling} tasks.

\begin{figure}[tb]
  \centering
  \includegraphics[width=0.49\linewidth]{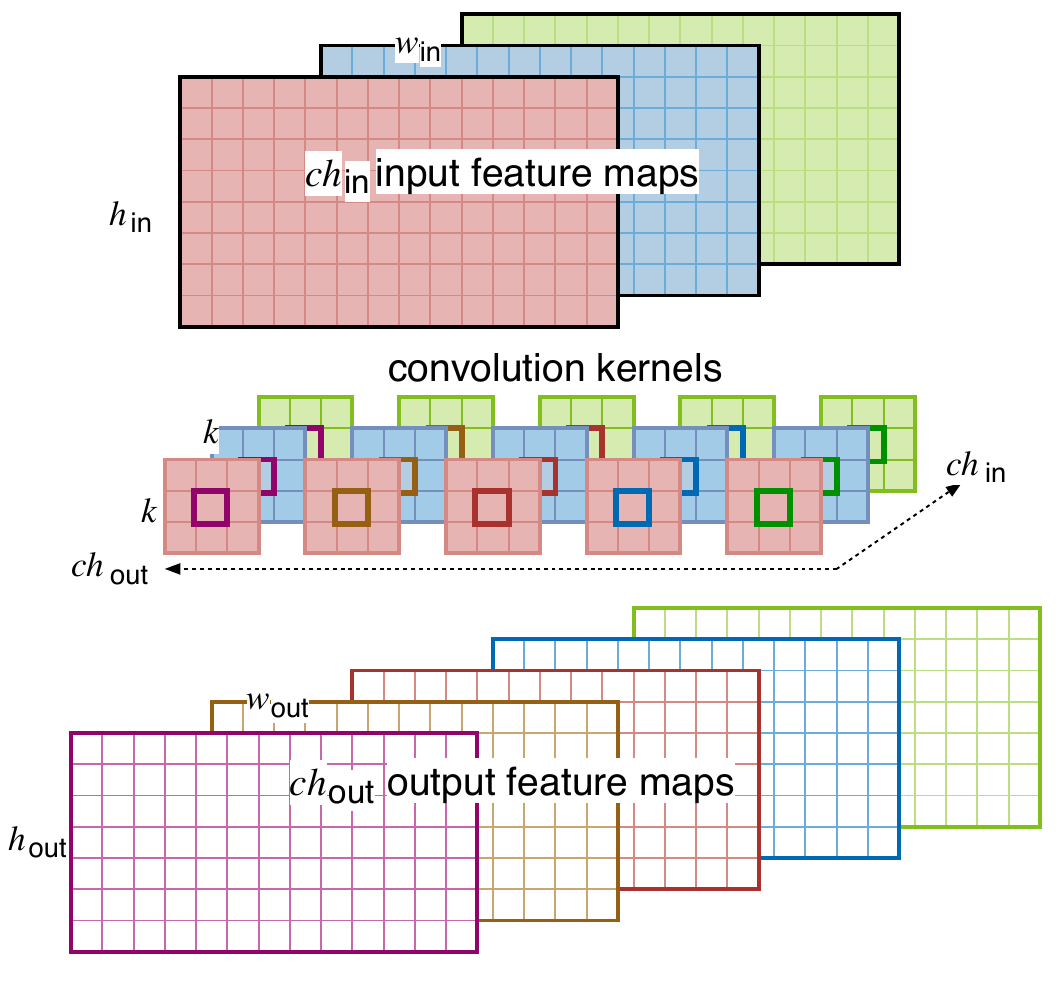}\hfill
  \includegraphics[width=0.46\linewidth, trim=0cm -1cm 0 0]{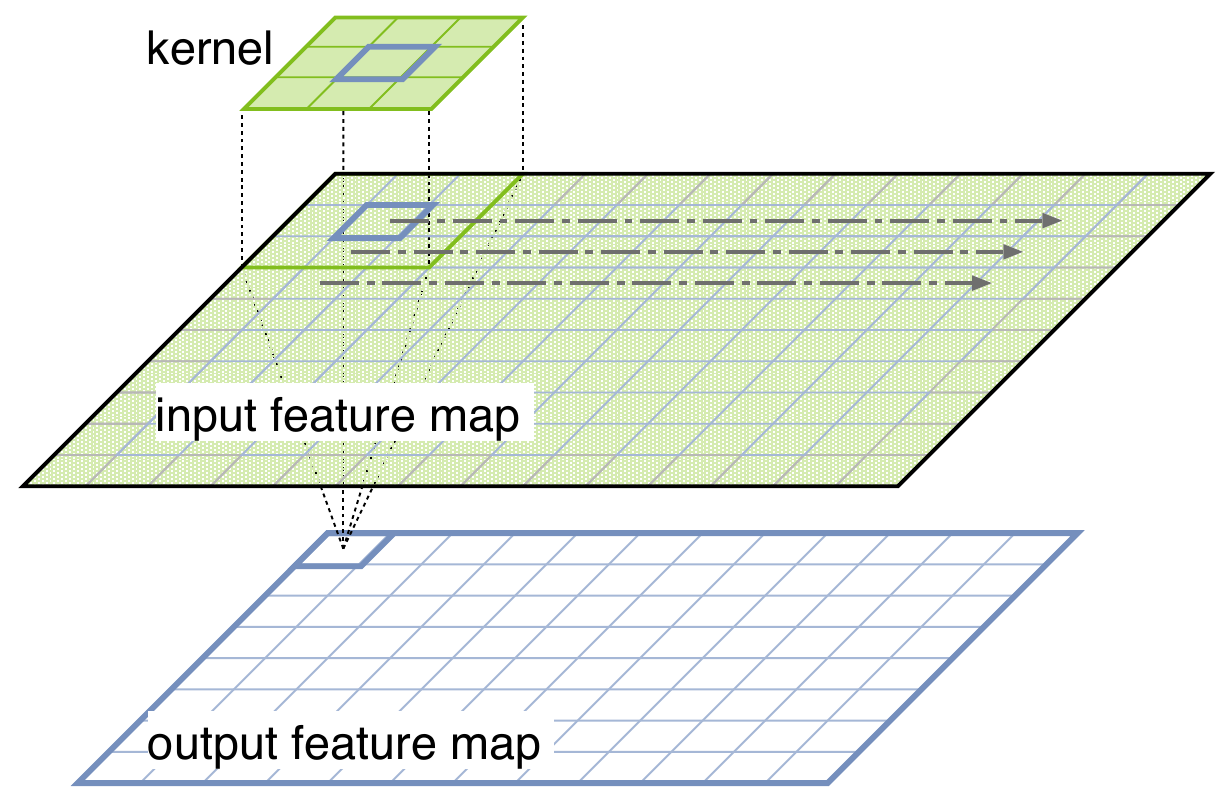}
  \caption[Convolutional Layer Nomenclature and Illustration]{
  Left: Illustration of the CNN Layer Nomenclature. The \chin input feature maps (solid R, G, B) are transformed into \chout output feature maps (outlined P, B, R, B, G) by applying \chin\x\chout filter kernels of size $k$\x$k$. Right: Illustration of the 2D convolution between a 3\x3 kernel and an input feature map by sliding the kernel over the input pixels and performing multiply-accumulate operations at each pixel position.
  }
  \label{fig:convolution-kernel}
\end{figure}

\paragraph{Nomenclature}
The input to each layer in a convolutional neural network consists of a stack of \chin 2D images of dimension \hin\x\win, the so-called \emph{input feature maps}.
Each layer produces a stack of \chout 2D images of dimension \hout\x\wout, called \emph{output feature maps}. An illustration can be found in \cref{fig:convolution-kernel}.

\paragraph{Motivation}
When neural networks are employed for image-related tasks, their input usually consists of pixel data. Even for an image with a modest resolution of 256\x256 RGB pixels, the resulting
input consists of $256\times256\times3\approx\SI{200000}{}$ elements, and a subsequent fully-connected neural network layer would need billions of weights. Luckily, there is no need for full connectivity when dealing with pixel data thanks to the locality of information in images. In order to decide whether there is a car in the center of an image one does not need to consider the color of the top-right corner pixel --- and the bottom-right pixels usually do not influence the class assigned to the top-left pixels. The important information in images can be captured from local neighborhood relations. Strong contrasts indicate edges, aligned edges result in lines, combined lines can result in circles and contours, circles can outline a wheel and multiple nearby wheels can point to the presence of a car~\cite{farabet_2010,cs231n}.
This locality of information in images is exploited in convolutional neural networks by replacing the fully-connected layers with \emph{convolutional layers}.

\paragraph{Weight Sharing by Convolution}
\label{par:cnn-weight-sharing}
A convolutional layer contains a \chin\x\chout array of kernels, which are small filters of size $k$\x$k$ (typically 1\x1, 3\x3, 5\x5, 7\x7 or 11\x11).
These kernels are applied to the input feature maps by means of 2D convolution. Each output pixel is thus generated from just a small \emph{local receptive field} in the input image.
\chout filter kernels are slid over each input feature map. For each input feature map, this results in \chout partial output feature maps.
The final output feature maps are formed by summing the partial output feature maps contributed by all \chin input channels (see \cref{fig:convolution-kernel} for an illustration, a mathematical formulation follows in \cref{eq:convlayer-formula} in \cref{sec:math-formulation-convolution}).
Instead of requiring $(\hin \times \win \times \chin) \times (\hout \times \wout \times \chout)$ weights, the number of parameters in a convolutional layer is thus reduced to $(k\cdot k) \times (\chin \times \chout)$.
The independence from the input image dimensions also enables large images to be processed without an exploding number of weights~\cite{cs231n,origami_sa,mnielsen}.

\begin{figure}[tb]
  \centering
  \includegraphics[width=\linewidth]{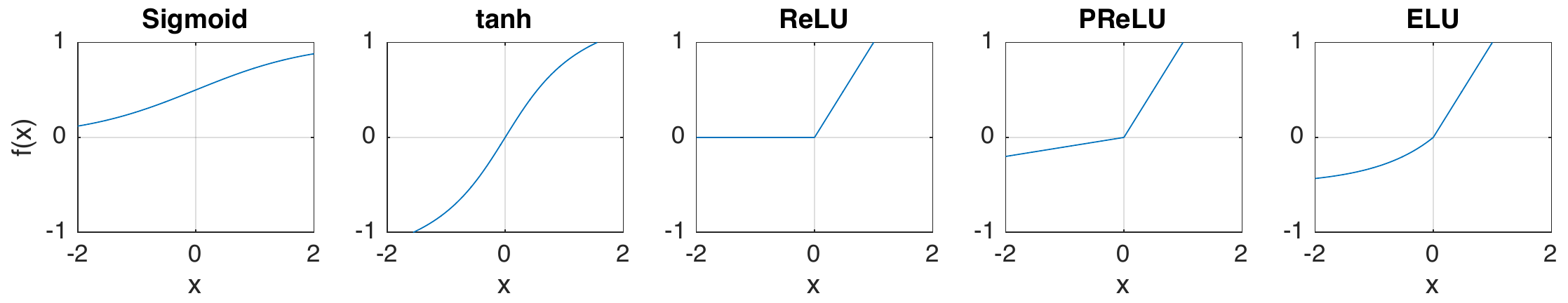}
  \caption[Comparison of Non-Linear Activation Functions]{The Non-Linear Activation Functions Sigmoid, tanh, ReLU, PReLU and ELU.}
  \label{fig:nonlinearities}
\end{figure}

\clearpage\paragraph{Layer Types}
Convolutional neural networks are constructed by stacking a number of generic network layers, which transform the input feature maps of dimension $(\hin\times\win\times\chin)$ into output feature maps of dimension $(\hout\times\wout\times\chout)$ \cite{cs231n,caffe_tutorials}. A typical CNN consists of the following layer types:

\begin{description}
  \item[Convolutional Layers] apply $(\chin \times \chout)$ filters of size $(k \times k)$ to generate the output feature maps.
  For filters larger than 1\x1, border effects reduce the output dimensions. To avoid this effect, the input image is typically \emph{padded} with $p = \lfloor k/2 \rfloor$ zeros on each side.
  The filters can be applied with a \emph{stride} $s$, which reduces the output dimensions to $\wout = \win/s$ and $\hout = \hin/s$.

  \item[Nonlinearity Layers] apply a non-linear activation function to each input pixel. The most popular activation function is the \emph{Rectified Linear Unit (ReLU)} which computes $f(x) = \max(0,x)$ and clips all negative elements to zero. Early networks used sigmoidal functions such as $f(x) = 1/(1+e^{-x})$ or $f(x)=\tanh(x)$, but these are no longer used because of their computational complexity and their slowing effect on convergence during training. More recent ideas include the \emph{Parametric ReLU (PReLU)} $f(x) = \max(\alpha\cdot x,x)$ with learnable parameter $\alpha$ \cite{relu_init}, \emph{Maxout} \cite{maxout} and \emph{Exponential Linear Units (ELU)} \cite{elu}. \Cref{fig:nonlinearities} shows a comparison of some of these options.

  \item[Pooling Layers] reduce the spatial dimensions of the input by summarizing multiple input pixels into one output pixel.
  Two popular choices are \emph{max-pooling} and \emph{avg-pooling}, which summarize their local receptive field by taking the maximum or the average value of the pixels, respectively.
  They are usually applied to a patch of 2\x2 or 3\x3 input pixels with a stride $s=2$,
  but can also be applied as \emph{global pooling} to the whole input image, in order to reduce the spatial output dimensions to 1\x1 pixels.

  \item[Fully-Connected Layers] are often used as the last layers in a CNN to compute the class scores in image classification applications. Even though the spatial dimensions \hin and \win in the last layers are typically heavily reduced, the fully-connected layers often account for most of the weights in these CNNs.

  \item[Local Response Normalization (LRN) Layers] introduce competition between the neurons of adjacent output channels by normalizing their responses with respect to a certain neighborhood of $N$ channels. LRN layers were introduced in the famous \emph{AlexNet} architecture \cite{alexnet}, but are used less often in recent CNNs.

  \item[Batch Normalization (BN) Layers] were introduced in 2015 by researchers at Google \cite{batchnorm}. Batch Normalization is applied after every training batch and normalizes the layer's output distribution to zero-mean, unit-variance. The uniform input distribution to subsequent layers should allow higher learning rates and thereby accelerate the training and improve the accuracy of the network. However, as of this writing, BN layers are not fully supported on all training platforms and can be difficult to employ in practice.

  \item[Dropout Layers] are a popular method to combat overfitting in large CNNs. These layers randomly drop a selectable percentage of their connections during training, which prevents the network from learning very precise mappings, and forces some abstraction and redundancy to be built into the learned weights.

  \item[Softmax Layers] are the most common \emph{classifiers}. A classifier layer is added behind the last convolutional or fully-connected layer in each image classification CNN, and squashes the raw class scores $z_i$ into class probabilities $P_i$ according to $P_i = {e^{z_i}} / {\sum_k{e^{z_k}}}$, which results in a vector $P$ that sums up to $1$.
\end{description}


\paragraph{Neural Network Training Frameworks}

There are many popular software frameworks specifically built for the design and training of neural networks, including, among others, the \emph{Neural Network Toolbox for MATLAB} \cite{matlab_nn_toolbox}, \emph{Theano} \cite{theano} with the extensions \emph{Lasagne} \cite{lasagne} and \emph{Keras} \cite{keras}, \emph{Torch} \cite{torch}, \emph{TensorFlow} \cite{tensorflow} and \emph{\caffe} \cite{caffe}. Most of these frameworks can utilize one or multiple GPUs in order to heavily accelerate the training of neural networks.
For this thesis, the \caffe framework has been used due to its maturity, its support in the GPU-based training system \emph{NVidia DIGITS}, \cite{digits} and most importantly because of the excellent availability of network descriptions and pretrained network topologies in native \caffe format.

\paragraph{Network Specification} In order to fully describe a convolutional neural network, the following information is required:
\begin{enumerate}
  \item a topological description of the network graph
  \item a list of layers and their settings
  \item the weights and biases in each layer
  \item (optionally) a training protocol
\end{enumerate}

In \caffe, the network description and the layer settings are stored in a JSON-like, human-readable text format called \texttt{.prototxt}. The weights are saved in binary \texttt{.caffemodel} files.
The training protocol is also supplied in \texttt{.prototxt} format and includes settings such as the base learning rate, the learning rate schedule, the batch size, the optimization algorithm, as well as the random seeds for training initialization.
These settings are only needed if the network is to be trained from scratch or \emph{finetuned}, which refers to the process of adapting a trained network to a different dataset. For \emph{inference}, where a fully trained network is utilized for forward-computation on new input data, the network description and the trained weights are sufficient.


\clearpage\subsection{Network Topologies for Image Classification}
\label{sec:cnn-topologies}

One of the most interesting, yet also one of the hardest problems in Computer Vision is \emph{Image Classification}: The task of correctly assigning one out of several possible labels to a given image. Examples for this problem include yes-or-no decisions (Is there a person in front of the car? Is this tissue sample cancerous?) but also recognition tasks with a large number of labels (What breed of dog is this? Who is on this photo?). As an extension of image classification, \emph{scene labeling} assigns a class to every pixel of the input image.

\begin{figure}[tb]
  \centering
  \includegraphics[width=\linewidth/(8)]{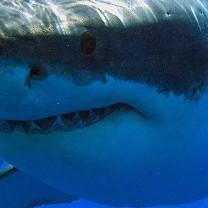}
  \includegraphics[width=\linewidth/(8)]{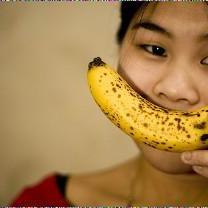}
  \includegraphics[width=\linewidth/(8)]{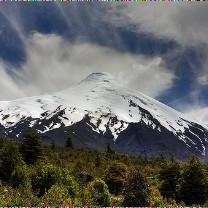}
  \includegraphics[width=\linewidth/(8)]{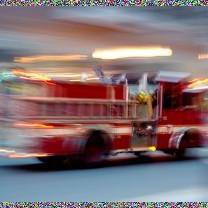}
  \includegraphics[width=\linewidth/(8)]{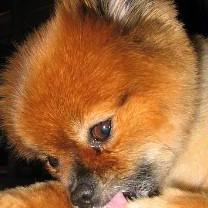}
  \includegraphics[width=\linewidth/(8)]{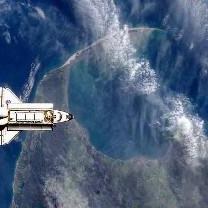}
  \includegraphics[width=\linewidth/(8)]{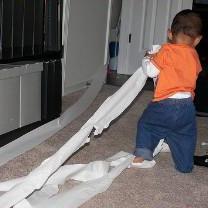}
  \caption[Example Images from the ImageNet Dataset]{Sample Images from the ImageNet Challenge (white shark, banana, volcano, fire engine, pomeranian, space shuttle, toilet paper)}
  \label{fig:ilsvrc}
\end{figure}

\paragraph{ImageNet Challenge}
The \emph{ImageNet Large Scale Visual Recognition Challenge (ILSVRC)} is an annual competition where participants develop algorithms to classify images from a subset of the \emph{ImageNet} database. The ImageNet database consists of more than 14 million photographs collected from the Internet, each labeled with one groundtruth class. The ILSVRC training set consists of approximately 1.2 million images in 1000 different classes, covering a huge variety of objects (from toilet paper, bananas and kimonos to fire trucks, space shuttles and volcanoes), scenes (from valleys and seashores to libraries and monastries) and animals (120 breeds of dogs, but also axolotls, sharks and triceratop dinosaurs). Some sample images from the challenge are shown in \cref{fig:ilsvrc}.
Participants are allowed to make five predictions. The \emph{top-1 accuracy} tracks the percentage of correct labels assigned at first guess, and the \emph{top-5 accuracy} takes all five predictions into account.
Humans can reach approximately \SI{5}{\%} top-5 error rate with explicit training and concentrated effort~\cite{ilsvrc_human,ilsvrc}.

\paragraph{CNN Topologies for Image Classification on ImageNet} The huge number of training samples and the difficulty of the problem make the ImageNet challenge an ideal playground for machine learning algorithms. Starting with \emph{AlexNet} in 2012, convolutional neural networks have taken the lead in the ILSVRC competition, and the top-1 and top-5 error rates of the winning entries have dropped significantly every since. The most important topologies are summarized in \cref{tab:topology-summary}, visualized in \cref{fig:cnn-topologies-visualization} and quickly introduced in the following list.%
\footnote{%
  The top-5 error rates reported in this list correspond to the performance of the ILSVRC submissions, unless otherwise stated. The participants often use multi-net fusion (fusing the predictions of multiple separately trained networks) and multi-crop evaluation (averaging the predictions made on different crops of the input image) to boost their accuracy. The single-net single-crop error rate of these CNNs can differ significantly.
}

\begin{table}[tbp]
\centering
\caption[Comparison of CNN Topologies for ImageNet Classification from Prior Work]{Comparison of Different CNN Topologies for Image Classification on ImageNet. The top-5 error rate is listed for single-net, single-crop evaluation. \#MACCs is the number of multiply-accumulate operations in one forward pass. \#activations is the total pixel count in all output feature maps.}
\label{tab:topology-summary}
\begin{tabular}{@{}lrrrrr@{}}
\toprule
      & \begin{tabular}[c]{@{}r@{}}\#conv.\\ layers\end{tabular}
      & \begin{tabular}[c]{@{}r@{}}\#MACCs\\{[}millions{]}\end{tabular}
      & \begin{tabular}[c]{@{}r@{}}\#params\\{[}millions{]}\end{tabular}
      & \begin{tabular}[c]{@{}r@{}}\#activations\\{[}millions{]}\end{tabular}
      & \begin{tabular}[c]{@{}r@{}}ImageNet\\top-5 error\end{tabular} \\
\midrule 
AlexNet             & \num{5}    & \num[group-minimum-digits = 4]{1140}  & \num{62.4}  & \num{2.4}   & \num{19.7}{\%} \\  
Network-in-Network  & \num{12}   & \num[group-minimum-digits = 4]{1100}  & \num{7.6}   & \num{4.0}   & \textasciitilde\num{19.0}{\%} \\  
VGG-16              & \num{16}   & \num[group-minimum-digits = 4]{15470} & \num{138.3} & \num{29.0}  & \num{8.1}{\%}  \\  
GoogLeNet           & \num{22}   & \num[group-minimum-digits = 4]{1600}  & \num{7.0}   & \num{10.4}  & \num{9.2}{\%}  \\  
ResNet-50           & \num{50}   & \num[group-minimum-digits = 4]{3870}  & \num{25.6}  & \num{46.9}  & \num{7.0}{\%}  \\  
Inception v3        & \num{48}   & \num[group-minimum-digits = 4]{5710}  & \num{23.8}  & \num{32.6}  & \num{5.6}{\%}  \\  
Inception-ResNet-v2 & \num{96}   & \num[group-minimum-digits = 4]{9210}  & \num{31.6}  & \num{74.5}  & \num{4.9}{\%}  \\  
SqueezeNet          & \num{18}   & \num[group-minimum-digits = 4]{860}   & \num{1.2}   & \num{12.7}  & \num{19.7}{\%} \\  
\bottomrule
\end{tabular}
\end{table}

\begin{description}
  \item[AlexNet] by Alex Krizhevsky et al. from the University of Toronto was the first CNN to win the ILSVRC in 2012. AlexNet consists of 5 convolutional layers, has 60 million parameters and requires approximately 1.1 billion \emph{multiply-accumulate} (MACC) operations for one forward pass. The network achieved a groundbreaking top-5 error rate of \SI{15.3}{\%} on ILSVRC 2012, with the second-best entry left behind at \SI{26.2}{\%}~\cite{alexnet}.

  \item[Network-in-Network (NiN)] by Min Lin et al. from the National University of Singapore was published as a novel CNN architecture in 2013. The NiN architecture consists of small, stacked multilayer perceptrons which are slid over the respective input just like convolutional filters. Additionally, the authors use global average pooling in the classifier instead of fully-connected layers. This makes the network much smaller in terms of parameters. NiN never officially participated in ILSVRC, but can be trained on the ImageNet dataset and reaches approximately AlexNet-level accuracy~\cite{firecaffe,nin}.

  \item[VGG] stands for Visual Geometry Group, University of Oxford, and also names this group's CNN architecture which won part of the ILSVRC 2014 challenge. The researchers experimented with deep CNNs containing up to 19 convolutional layers. The most popular variant \emph{VGG-16} has a depth of 16 layers, and a very regular structure, consisting exclusively of 3\x3 convolution and 2\x2 max-pooling layers. The spatial dimensions are steadily reduced from 224\x224 pixels to 7\x7 pixels, while the number of channels is simultaneously increased from 3 to 4096. The network reached a top-5 error of \SI{7.3}{\%}. However, VGG-16 contains almost 140 million weights and one forward pass requires nearly 16 billion MACC operations~\cite{vgg}.

  \item[GoogLeNet] by Christian Szegedy et al. from Google is a milestone CNN architecture published just a few days after the VGG architecture.
  The 22-layer GoogLeNet set a new ILSVRC classification record with a top-5 error rate of \SI{6.67}{\%}, while requiring only 1.2 million parameters and 0.86 billion MACC operations.\footnote{The ILSVRC-2014 winning entry used multi-crop evaluation on 144 crops for this result. Single-crop performance is rather in the order of \SI{9}{\%} top-5 error \cite{caffe-googlenet}.}
  The savings are achieved by a more complex architecture which employs so-called \emph{Inception modules}.
  These modules are a \emph{network-in-network} sub-architecture which first uses a 1\x1 convolutional layer to reduce the number of channels, before expanding this compressed representation again using parallel convolutional layers with kernel sizes 1\x1, 3\x3 and 5\x5.
  The reduction in the channel dimension decreases the number of parameters and MACC operations in both the reducing and the expanding layers, and the composition of multiple layers increases the non-linear expressiveness of the network. To improve training convergence, GoogLeNet makes use of LRN layers~\cite{googlenet}.

  \item[ResNet] by Kaiming He et al. from Microsoft Research won the ILSVRC in 2015. Their very deep ResNet-152 model achieved a top-5 error rate of less than \SI{5.7}{\%} by using 152 convolutional layers. Models with a depth of more than 20 convolutional layers were previously very hard to train. The researchers solved this problem by including detours around each batch of two subsequent convolutional layers, summing both the detoured original and the filtered representation together at the junction points. This topology resembles a function $y = F(x) + x$ where the network only needs to learn the \emph{residual function} $F(x)$, merely ``adding information'' rather than reinventing the wheel every two layers. The smaller version ResNet-50 uses 50 convolutional layers and Batch Normalization, has 47 million parameters and needs 3.9 billion MACC operations per forward pass to reach a top-5 error of \SI{6.7}{\%}~\cite{resnet}.

  \item[Inception v3 and v4] by Christian Szegedy et al. are Google's latest published image classification CNNs. The GoogLeNet architecture has been thoroughly studied and optimized in the Inception v3 paper \cite{inceptionv3}, with valuable hints on how to design and modify CNNs for efficiency.
  The Inception v4 paper \cite{inceptionv4}, published in February 2016, studies the positive effects of residual connections in Inception module-based architectures and presents \emph{Inception-ResNet-v2} which reaches a \SI{4.1}{\%} top-5 error rate on the ILSVRC dataset. All recent Inception architectures make heavy use of Batch Normalization layers~\cite{inceptionv3,inceptionv4}.

  \item[SqueezeNet] by Forrest Iandola et al. from UC Berkeley, also published in February 2016, differs from the other CNN architectures in this list because the design goal was not record-breaking accuracy. Instead, the authors developed a network with an accuracy similar to AlexNet, but with 50\x\ less parameters. This parameter reduction has been achieved by using \emph{Fire modules}, a reduce-expand micro-architecture comparable to the Inception modules, and careful balancing of the architecture. The 18-layer SqueezeNet uses 7\x7, 3\x3 and 1\x1 convolutions, 3\x3 max-pooling, dropout and global average pooling, but neither fully-connected, nor LRN, nor Batch Normalization layers. One forward pass requires only 860 million MACC operations, and the 1.24 million parameters are enough to achieve less than \SI{19.7}{\percent} single-crop top-5 error~\cite{squeezenet}.\footnote{AlexNet also has a top-5 error of \SI{19.7}{\percent} with single-crop evaluation. The \SI{15.3}{\percent} top-5 error on ILSVRC has been achieved using multi-net fusion and multi-crop evaluation.}
\end{description}

\subsection{Compression of Neural Network Models}
\label{sec:network-compression}
State-of-the-art CNNs require significant amounts of memory for their weights (e.g. \SI{560}{\mega\byte} for VGG-16 with 32-bit weights) which can be problematic for example regarding over-the-air updates or the deployment on embedded systems. Researchers have been looking for ways to reduce both the number of weights, and the memory required per weight.

\paragraph{Kernel Decomposition and Pruning} Denil et al. demonstrate that up to \SI{95}{\%} of all weights in their CNN can be predicted instead of learned, without a drop in accuracy \cite{compression_denil}. Denton et al. approximate fully trained convolution kernels using singular value decomposition (SVD) \cite{compression_denton}, while Jin et al. replace the 3D convolution operation by three consecutive one-dimensional convolutions (across channel, horizontal, vertical) \cite{compression_jin}. Similar methods have been used to efficiently deploy CNNs on smartphones~\cite{compression_yong}. A final idea is \emph{network pruning}, where small or otherwise unimportant weights are set to zero, which effectively removes the corresponding connections~\cite{compression_pruning,compression_structuredpruning}.

\paragraph{Limited Numerical Precision} Reducing the memory consumption of each weight is possible by replacing the typical \emph{32-bit floating-point} weights either with \emph{16-bit floating-point} weights \cite{google-fp16-results, nvidia-jetson-whitepaper} or with \emph{fixed-point approximations} of less than 32 bits \cite{limitedprecision}.
Neural networks have been shown to tolerate this type of quantization very well.
Hwang et al. successfully quantized most layers in their CNN to three bits \cite{fp_optimization}, Sung et al. restricted their network to ternary values ($-1$,$0$,$1$) with a negligible drop in accuracy \cite{resiliencyquantization}, and Courbariaux et al. even train CNNs with binary weights and activations \cite{binarynet,binaryconnect}.
With Ristretto, Gysel et al. recently published an automated CNN approximation tool which analyzes floating-point networks and condenses their weights to compact fixed-point formats, while respecting a maximally allowed accuracy drop \cite{ristretto}.

\paragraph{Deep Compression} Finally, Han et al. combine pruning, trained quantization and Huffman coding to reduce the storage requirement of AlexNet by a factor of $39\times$, and that of VGG-16 even $49\times$ without any drop in accuracy \cite{compression_deepcompression}. For all methods mentioned, finetuning the network with the compressed weights helps to recover most of the initial accuracy loss.

%
\clearpage\section{Field-Programmable Gate Arrays}
\label{sec:fpga-intro}

This section gives a high-level introduction to \emph{Field-Programmable Gate Arrays} (FPGAs). The first part highlights characteristics, strengths and weaknesses of this hardware platform, before the second part focuses on \emph{High-Level Synthesis} (HLS), a relatively new methodology which makes it possible to program FPGAs in high-level languages such as C and \cpp.

%
\subsection{Introduction to Field-Programmable Gate Arrays}

Field-Programmable Gate Arrays (FPGAs) are semiconductor devices consisting of a 2D array of configurable logic blocks (CLBs, or \emph{logic slices}), which are connected via programmable interconnects. The interconnect can be thought of as a network of wire bundles running vertically and horizontally between the logic slices, with switchboxes at each intersection.
Modern high-end FPGA generations feature hundreds of thousands of configurable logic blocks, and additionally include an abundance of hardened functional units which enable fast and efficient implementations of common functions.\footnote{This includes on-chip SRAM (Block RAM), USB, PCIe and Ethernet Transceivers, Serializer-Deserializer circuits, Digital Signal Processor (DSP) Slices, Cryptographic Accelerators, PLLs, Memory Interfaces and even full ARM processor cores.}
The logic blocks, the fixed-function units as well as the interconnect are programmed electronically by writing a \emph{configuration bitstream} into the device. The configuration is typically held in SRAM memory cells, and the FPGAs can be reprogrammed many times~\cite{kaeslin,xilinx-what-is-fpga}.

\paragraph{FPGAs versus General-Purpose Processors}
The advantage of FPGA-based systems over traditional processor-based systems such as desktop computers, smartphones, most embedded systems, and also over GPUs, is the availability of freely programmable general-purpose logic blocks. These can be arranged into heavily specialized accelerators for very specific tasks, resulting in improved processing speed, higher throughput and energy savings. This advantage comes at the price of reduced agility and increased complexity during the development, where the designer needs to carefully consider the available hardware resources and the efficient mapping of his algorithm onto the FPGA architecture. Further, some algorithmic problems do not map well onto the rigid block structures found on FPGAs~\cite{kaeslin,xilinx-hls-intro}.

\paragraph{FPGAs versus ASICs}
\emph{Application-Specific Integrated Circuits} (ASICs) are custom-tailored semiconductor devices.
In contrast to FPGAs, they do not suffer any area or timing overhead from configuration logic and generic interconnects, and therefore typically result in the smallest, fastest and most energy-efficient systems.
However, the sophisticated fabrication processes for ASICs results in lengthy development cycles and very high upfront costs, which demands a first-time-right design methodology and very extensive design verification.
Therefore ASICs are mostly suited for very high-volume, cost-sensitive applications where the non-recurring engineering and fabrication costs can be shared between a large number of devices.
FPGAs with their reprogrammability are better suited for prototyping and short development cycles~\cite{kaeslin}.

%
\clearpage\subsection{Introduction to High-Level Synthesis}
\paragraph{Hardware Description Languages and Register Transfer Level Design} Traditionally, FPGAs are programmed using a \emph{Hardware Description Language} (HDL) such as VHDL or Verilog.
Most designs are described at \emph{Register Transfer Level} (RTL), where the programmer specifies his algorithm using a multitude of parallel processes which operate on vectors of binary signals and simple integer data types derived from them.
These processes describe combinational logic, basic arithmetic operations as well as registers, and are driven by the rising and falling edges of a clock signal.
RTL descriptions are very close to the logic gates and wires that are actually available in the underlying FPGA or ASIC technology, and therefore the hardware that results from \emph{RTL synthesis} can be closely controlled.
However, the process of breaking down a given algorithm into logic blocks, processes and finite state machines on the register transfer level is very tedious and error-prone. Many design decisions have to be made before writing any code, and later changes are difficult and costly. This prevents iterative optimizations and demands a lot of intuition, experience and expert knowledge from designers~\cite{kaeslin}.

\paragraph{Increasing the Level of Abstraction with HLS}
\emph{High-Level Synthesis} (HLS) tries to lower this barrier to entry by enabling designers to specify their algorithms in a high-level programming language such as C, \cpp or SystemC. Many implementation details are abstracted away and handled by the \emph{HLS compiler}, which converts the {sequential software description} into a {concurrent hardware description}, usually at RTL level.

\paragraph{Vivado High-Level Synthesis} \emph{Vivado High Level Synthesis} (VHLS) by Xilinx Inc. is one of the most popular commercial HLS compilers.
With VHLS, designers can use loops, arrays, structs, floats, most arithmetic operations, function calls, and even object-oriented classes. These are automatically converted into counters, memories, computation cores and handshake protocols as well as accompanying state machines and schedules. The compilation can be influenced using scripted \emph{compiler directives} or embedded \emph{compiler pragmas}, which are meta-instructions interpreted directly by the VHLS compiler. Operations are by default scheduled to be executed concurrently and as early as possible. Using the compiler pragmas, the designer can further influence the inference of memories and interfaces, the parallelization of loops and tasks, the synthesis of computation pipelines, etc.~\cite{xilinx-hls-intro, xilinx-ug902}

\paragraph{Promises and Difficulties} The increased abstraction level in High-Level Synthesis promises faster development cycles, flexible optimization strategies and much higher productivity at the cost of slightly less control on the end result. Especially with regard to every-increasing design complexities, shrinking time-to-market requirements and the abundant resources in modern FPGAs, such a compromise would be very welcome.
However, HLS tools have been on the market for more than 12 years now, yet most engineers  still use RTL descriptions for their FPGA and ASIC designs. The task of converting sequential, high-level software descriptions into fully optimized, parallel hardware architectures is tremendously complex. Although companies have invested hundreds of millions of dollars and years of research into HLS \cite{xilinx-buys-autoesl,cadence-buys-forte,synopsys-buys-synfora}, the results attained are still highly dependent on the coding style and intricate design details. Because flaws and deficiencies in the compiler are only discovered during the design, the decision for HLS is associated with a non-negligble risk~\cite{hls-bluebook}.

%
\clearpage\section{Embedded Convolutional Neural Networks}
\label{sec:prior-work}

The following sections give a short overview of different options for the implementation of convolutional neural networks in embedded systems. All of these embedded implementations focus on \emph{inference} using the CNN, and assume that the training is done offline using e.g. GPU-based training systems. \Cref{sec:hw-platforms-embedded} introduces the possible hardware platforms for the computation of CNNs, before \cref{sec:prior-embedded-cnns} presents a number of existing CNN implementations from prior work.

\subsection{Potential Hardware Platforms}
\label{sec:hw-platforms-embedded}

Embedded systems typically have very specific requirements and constraints such as limited power and energy budgets, finite battery capacities, small physical sizes resulting in limited heat dissipation capabilities, as well as high reliability requirements and hard real-time constraints.
These characteristics make the development of algorithms and systems for the embedded market different from the scientific playground where many neural networks are currently researched.
Still, there are a number of different options for the implementation of convolutional neural networks in embedded systems:

\begin{description}

\item[Central Processing Units (CPUs)] are the processor cores found in most of today's devices, including desktop computers and smartphones. Most of these CPUs are general-purpose, flexibly programmable and built for good performance on a maximally wide range of computational workloads. There exist many different types of processors suitable for embedded systems, with different tradeoffs regarding speed and power requirements. However, CPUs compute results sequentially\footnote{High-end CPUs can include multiple cores and SIMD instructions to attain a certain level of parallelization, but they are still primarily destined for sequential computation.} and are thus not ideally suited for the highly parallel problem presented by convolutional neural networks.

\item[Digital Signal Processors (DSPs)] are highly specialized microprocessors. They are optimized for processing floating-point signals fast and efficiently (especially multiply-accumulate operations) and they typically include \emph{Very Long Instruction Word} (VLIW) instructions to increase parallelism. Modern DSPs such as the Texas Instrument C6678 include eight cores, run at \SI{1.25}{\giga\hertz} and compute up to \SI{160}{GFLOP/s} at less than \SI{15}{\watt}. Specialized vision processors such as Cadence Tensilica Vision DSP \cite{cadence-vision-dsp, cadence-press-release} or the Movidius Myriad 2 \cite{movidius-myriad2} even promise teraflops of performance at just \SI{1}{\watt}. However, DSPs are still primarily ``few-core'' processors which are optimized for fast sequential operation and thus cannot fully exploit the parallelism present in CNNs.

\item[Graphics Processing Units (GPUs)] are many-core processors which were originally designed for highly parallel graphical workloads.
GPUs have recently been discovered for general-purpose computing tasks, referred to as \emph{General-Purpose Computing on GPUs} (GPGPU), which is supported by the OpenCL and CUDA programming frameworks.
High-end GPUs such as the NVidia GeForce GTX Titan X \cite{titan-x} contain more than 3000 floating-point processing cores running at \SI{1}{\giga\hertz}, and offer more than \SI{330}{\giga\byte/\second} memory bandwidth. They compute up to \SI{6600}{GFLOP/s}, but also consume up to \SI{250}{\watt}.
Mobile GPUs such as the NVidia Tegra X1 \cite{tegra-x1} (which is also used in the NVidia Jetson TX1 modules and the NVidia Drive PX platform) include up to \num{256} processing cores running at \SI{1}{\giga\hertz} and a memory bandwidth of roughly \SI{25}{\giga\byte/\second}. They compute up to \SI{512}{GFLOP/s} while consuming less than 10 watts~\cite{tegra-x1-watt}. GPUs are well suited for the parallel workloads presented by CNNs and are fully supported by most deep learning frameworks. They constitute the primary platform for research in the area of CNNs.

\item[Field-Programmable Gate Arrays (FGPAs)] have been introduced in \cref{sec:fpga-intro}. The largest devices, such as the Xilinx Virtex UltraScale+ XCVU13P, include more than \num{3}~million logic cells, \num{12} thousand DSP slices and \SI{56}{\mega\byte} of on-chip SRAM~\cite{virtex-ultrascaleplus}. Estimating the floating-point performance of FPGAs is not straight forward \cite{fpga-gflops}, but a conservative estimate for the XCVU13P with 3 DSP slices per multiplication and $f=\SI{300}{\mega\hertz}$ results in more than \SI{1000}{GFLOP/s} at a few tens of watts~\cite{xilinx-gflops,xilinx-fp-resource-utilization,virtex-ultrascaleplus-power}. 
FPGA designs work best for very regular calculations which can be heavily parallelized by building custom processing engines using the programmable logic blocks. Algorithms that require data-dependent branching and decisions are less suited for this type of parallelization and result in a poor utilization of the computational power. The performance of FPGA designs can be further increased by utilizing fixed-point or half-precision floating-point data formats.

\item[Application-Specific Integrated Circuits (ASICs)] are the ideal solution when it comes to maximum performance and maximum energy efficiency. However, ASICs are even less suited for irregular computation than FPGAs, and they further require much of the algorithm to be freezed at design time. For this reason, ASICs are typically only built to accelerate a certain aspect of CNNs, such as the partial calculation of a convolutional or fully-connected layer, but seldomly to calculate entire neural networks. A prominent exception are neuromorphic integrated circuits, which use analog electronic circuits to mimic neurons and neural networks on custom-designed ICs~\cite{neuromorphic-book}.

\end{description}

Besides these options for \emph{local evaluation} of the CNN, a popular approach is to delegate the energy and resource intensive computation to remote datacenters. However, this method requires a permanent high-bandwidth network connection and introduces additional latency which might not be acceptable, e.g. in mobile, safety-relevant or real-time systems.

%

\subsection[Existing CNN Implementations on Embedded Platforms]{\scalebox{0.99}{Existing CNN Implementations on Embedded Platforms}}
\label{sec:prior-embedded-cnns}

This section introduces some of the most important milestones in the field of non-GPU-powered CNN implementations, with a special focus on FPGA-based solutions.

\paragraph{The Design Space of Neural Network Accelerators} In his mid-2015 research proposal \cite{drumond-epfl}, M. Drumond from EPFL Lausanne provides a survey of the design space of neural network accelerators on the platforms GPU, ASIC and FPGA. He focuses on the tradeoffs involved (in terms of energy-efficiency, flexibility and scalability) and the performance achievable. The paper provides an excellent overview of implementation options (albeit with a focus towards data center applications), and concludes that FPGAs can be much more energy efficient and scalable compared to GPUs, while maintaining a reasonable level of flexibility.

\paragraph{Deep Learning on FPGA: Past, Present and Future} Lacey et al. also investigate the suitability of FPGAs for accelerating CNNs in their 2016 paper \cite{deep-learning-fpga-past-present-future}. Besides presenting an overview of prior FPGA-based neural network accelerators, they propose to explore model-level optimizations on Convolutional Neural Networks to fully leverage the advantages of FPGAs. The paper identifies OpenCL and High-Level Synthesis as important steps towards the widespread acceptance of FPGAs as deep learning accelerators, and suggests that datacenters would especially profit from this platform's attractive scalability and performance per watt.

\paragraph{Accelerating Datacenter Workloads using FPGAs} Both Microsoft and Baidu seem to have come to the same conclusion, and have built FPGA-based accelerators for their datacenters. Microsoft's Catapult platform \cite{microsoft-catapult} (2014) was originally conceived to double the speed of the Bing ranking algorithm. It has been utilized to implement a record-breaking AlexNet accelerator in 2015 \cite{microsoft-cnn}, achieving $\sfrac{1}{2}$ of the throughput of a modern GPU at $\sfrac{1}{10}$ of the power budget (\cref{fig:fpga-accelerator-ideas} depicts a top-level overview of the accelerator architecture).
Chinese search giant Baidu has announced similar plans and a strategic partnership with FPGA manufacturer Altera~\cite{baidu-altera}.
Google also considered an FPGA-based accelerator for deep learning, but recently decided to go one step further and developed a custom ASIC solution~\cite{google-asic}.

\begin{figure}[tb]
  \centering
  \includegraphics[width=0.6\linewidth]{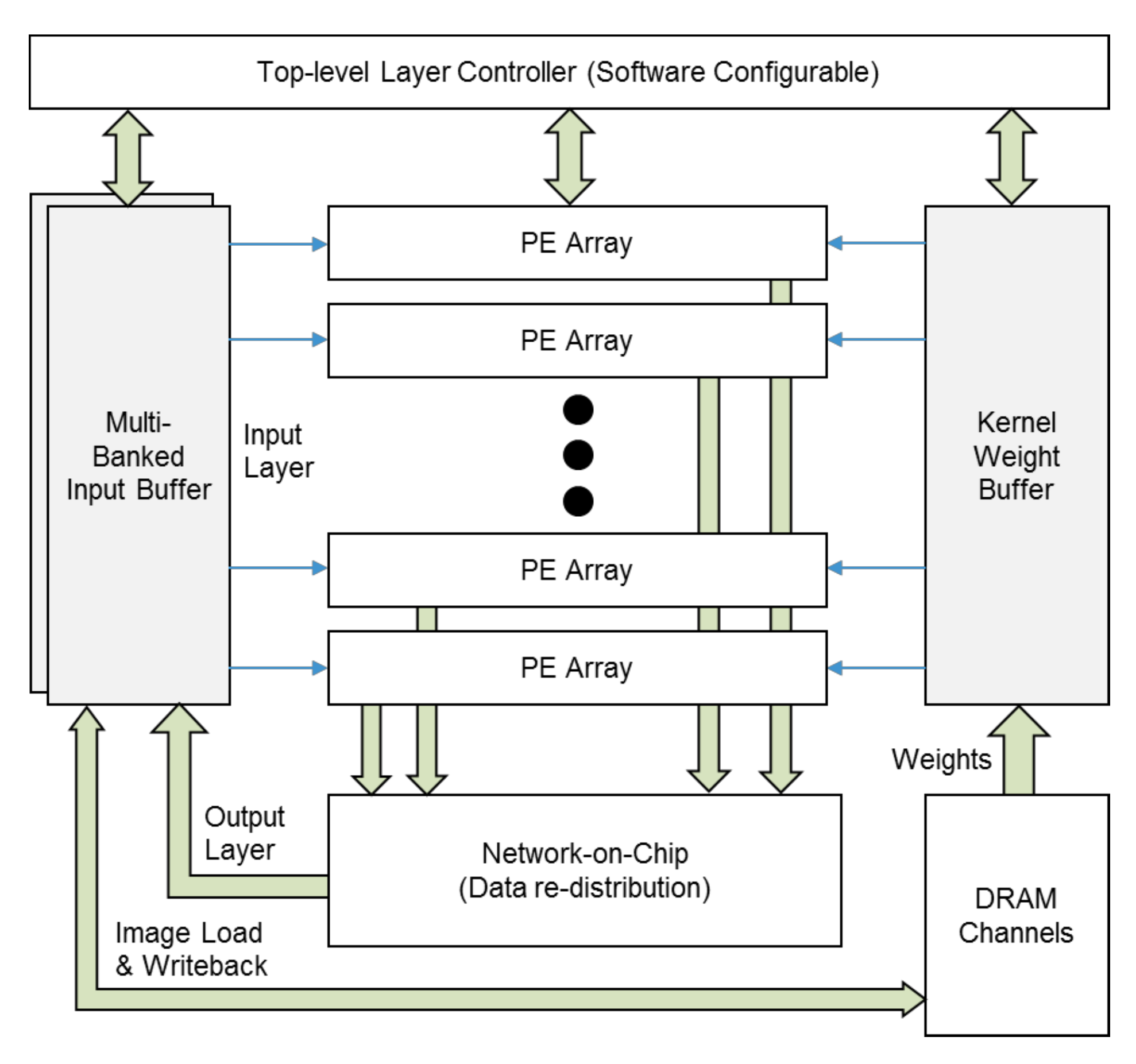}
  \caption[Block Diagram for Microsoft's FPGA-based CNN Accelerator]{Top-Level Overview of the FPGA-based CNN Accelerator developed by Microsoft. The architecture contains a number of generic Processing Elements as well as a Network-on-Chip, which feeds computation results back into the Input Buffer for reuse in the calculation of the next layer. \cite{microsoft-cnn}}
  \label{fig:fpga-accelerator-ideas}
\end{figure}

\paragraph{ASIC Implementations} \emph{DaDianNao} (2014) is a multi-chip accelerator system consisting of 64 ASIC nodes with large on-chip memories to save off-chip memory accesses and thereby optimize energy efficiency.
Based on their synthesis results, the authors claim up to 450\x\ higher performance and 150\x\ lower energy consumption with respect to a GPU implementation~\cite{dadiannao}.
\emph{Origami} (2015) is an accelerator ASIC co-developed by the author.
The IC has been designed as a co-processor to speed up the computationally intensive 2D convolutions in CNNs, with a focus on minimizing external memory bandwidth and maximizing energy efficiency.
The accelerator has been manufactured in 65nm technology and achieved new records in terms of area, bandwidth and power efficiency~\cite{origami_sa,origami_paper}.
An FPGA-based implementation is in progress as of 2016~\cite{origami-fpga}.
Finally, \emph{EyeRiss} (2016) is another accelerator ASIC for energy efficient evaluation of CNNs. The IC has been developed at the Massachusetts Institute of Technology and provides maximum flexibility regarding the network dimensions by using an array of 168 generic processing elements and a flexible network-on-chip interconnect.
Additionally, this IC features run-length compression of the off-chip memory and automatic zero skipping to conserve energy~\cite{eyeriss}

\paragraph{Optimizing FPGA-based CNN accelerators through automated Design Space Exploration}
In their early-2015 paper, Zhang et al. observe that most previous FPGA-based CNN accelerators do not achieve best performance due to underutilization of either logic resources or memory bandwidth.
The researchers use a polyhedral-based optimization framework to identify all legal permutations and tilings of the nested loops which form the algorithmic basis of a convolutional layer.
All these potential schedules are then analyzed with respect to their memory bandwidth and computational throughput requirements.
Using a roofline model (FLOPS vs. Computation-to-Communication Ratio), the accelerator with best performance and lowest memory bandwidth requirement is then selected (see \cref{fig:fpga-accelerator-ideas} for an illustration).
Zhang et al. successfully implement a proof-of-concept AlexNet accelerator with Vivado High-Level Synthesis on a Xilinx Virtex-7 485T FPGA~\cite{zhang-fpga}.%
\footnote{
At the time of publication, Zhang et al. set a new record by running inference on AlexNet at \SI{46}{FPS} drawing only \SI{18.6}{\watt}.
However, Microsoft's accelerator \cite{microsoft-cnn} soon broke the record, reaching almost 3\x\ the performance.
}
A very similar approach has been taken by Motamedi et al. in 2016 \cite{ucdavis_accelerator_dse}.
They identify four sources of parallelism in convolutional layers: inter-layer (independence of layers for different input images), inter-output (independence of output feature maps), inter-kernel (independence of convolutions at different image positions) and intra-kernel (independence of multiplications in convolution kernels).
The authors determine the ideal combination of these sources of parallelism by enumerating the design space of possible accelerators analytically.
By additionally utilizing the opportunity for tiling at kernel level, they achieve a speedup of almost 2\x\ compared to the accelerator proposed by Zhang et al.

\begin{figure}[tb]
  \centering
  \includegraphics[width=0.75\linewidth]{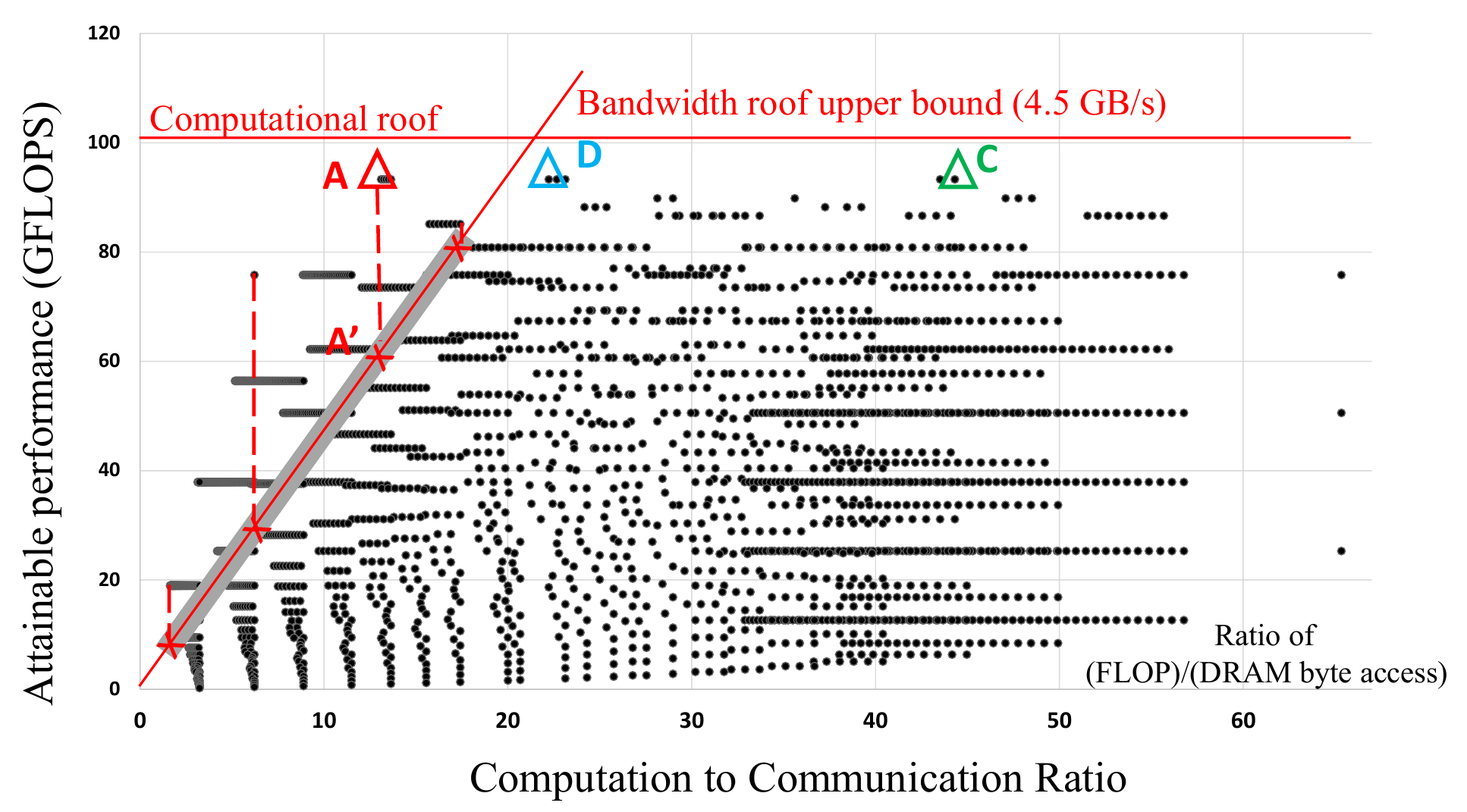}
  \caption[Illustration of DSE and Roofline Model by Zhang et al.]{Illustration of the Design Space Exploration and Roofline Method developed by Zhang et al. Their algorithm calculates the communication and computation requirements for a large number of implementation variants (shown as dots), draws the roof lines dictated by the platform (computational and memory bandwidth limits, red lines) and then selects the implementation with the highest throughput, yet lowest communication requirements, which still fits the platform's capacity (in this case, implementation C)~\cite{zhang-fpga}.}
  \label{fig:fpga-accelerator-ideas}
\end{figure}


\section{Project Goals and Specifications}
\label{sec:project-specs}

After consideration of the prior work introduced above and the evaluation of several alternatives (e.g. the design of a binary- or ternary-valued CNN), the project goal for this master thesis has been defined as ``[... to] build a demonstrator device that shows a convolutional neural network in operation'', with focus on the optimized co-operation of the neural network and the underlying hardware platform. The hardware platform has been fixed to the \emph{SCS Zynqbox}, an embedded systems platform based on the Xilinx Zynq XC-7Z045 System-on-Chip.\footnote{See \cref{sec:task description} for the original task description.}

\begin{figure}[btp]
	\centering
	\setlength{\fboxrule}{0.5pt}
	\scalebox{0.85}{\framebox{
		\begin{tikzpicture}[->,>=stealth,auto,thick,
							txt/.style={text width=2.3cm, align=center}]
		 \node [txt] (topo) {\textbf{CNN\\Topology}};   						 
		 \node [txt, below = 8mm of topo] (train) {Training};
		 \node [txt, below = 8mm of train] (perf) {Performance/\\Accuracy};
		 \node [txt, right = 1cm of topo] (arch) {\textbf{FPGA\\Architecture}};  
		 \node [txt, right = 1cm of train] (hls) {High-Level Synthesis};
		 \node [txt, right = 1cm of perf] (util) {Device\\Utilization};
		\path  (topo) 	edge[<->] (arch)  
		 	   (topo) 	edge (train)      
		 	   (train) edge (perf)
		 	   (perf.north west) 	edge[bend left=45] (topo.south west)
		 	   (arch) 	edge (hls)	      
		 	   (hls) 	edge (util)
		 	   (util.north east) 	edge[bend right=45] (arch.south east);
		\end{tikzpicture}
	}}
	\caption[Illustration of the System-Level Design Approach to this Project]{Illustration of the System-Level Design Approach for the Project, involving Optimization of both the CNN Topology and the FPGA Accelerator Architecture.}
	\label{fig:project-approach}
\end{figure}
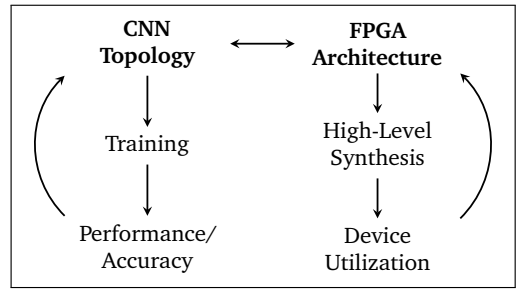

\paragraph{Design Approach} We decided to take a system-level design approach as illustrated in \cref{fig:project-approach}. The emphasis has been put equally on the \emph{design and optimization of a Convolutional Neural Network} and the \emph{design and optimization of an FPGA-based accelerator}, with the common purpose of reaching the best possible system-level performance. This approach is different from most previous FPGA-based CNN implementations, which typically rely on a maximally flexible accelerator to run a standard CNN from research.

\paragraph{Project Specification} The following requirements and constraints were the guidelines during the work on this project:
\par
Primary Goal: Design and Implementation of a real-time CNN demonstrator
\begin{enumerate}

	\item Implementation of best practices from prior work and recent research
	\item Optimization of a CNN for demonstration purposes
		\begin{enumerate}
			\item Image classification on ImageNet (realistic problem, impressive visuals)
        	\item Selection and Training of a suitable CNN topology (existing or custom-built)
        	\item Optimization of CNN for implementation on FPGA (resource efficiency)
        	\item Optimization of CNN for accuracy
        \end{enumerate}
	\item Elaboration of an FPGA-based architecture for the chosen CNN
		\begin{enumerate}
			\item Based on existing Zynqbox platform (Zynq XC-7Z045 + 1GB DDR3 Memory \cite{zynqbox})
        	\item Algorithm design and block-level organization (focus on energy efficiency)
        	\item Implementation using High-Level Synthesis
       		\item Optimization regarding efficiency, performance and device utilization
        \end{enumerate}
	\item Verification and evaluation of the CNN demonstrator system
\end{enumerate}

The remainder of this report details the implementation of these specifications.
\chapter[Convolutional Neural Network Analysis, Training and Optimization]%
{\scalebox{0.9}{Convolutional Neural Network}\linebreak\scalebox{0.9}{Analysis, Training and Optimization}}
\label{chap:cnn-design}

\cleanchapterquote{Perfection is achieved, not when there is nothing more to add, but when there is nothing left to take away.}{Antoine de Saint-Exupéry}{(Inspiring Writer and Pioneering Aviator)}

%
\section{Introduction}

The previous chapter introduced the two central goals of this project: The \emph{optimization of an Image Classification CNN for ImageNet}, and the \emph{design of a corresponding FPGA-based accelerator}.
This chapter is concerned with the first of the two aspects: the analysis of existing CNN Topologies, the setup of a training platform and finally the optimization of our custom CNN architecture.

First, the CNN topologies from prior work (presented in \cref{sec:cnn-topologies}) are thoroughly examined with regard to their resource efficiency. The corresponding \emph{Network Analysis Tools, Methods and Results} are presented in \cref{sec:cnn-analysis}.
The following \Cref{sec:network-training} then introduces the \emph{CNN Training Hardware and Software Setup} used for the training of more than 70 different CNN variants during this project, and shares some of the lessons learned during the countless hours of CNN training. The last \cref{sec:cnn-optimization} finally discusses the \emph{Network Optimizations} that have been applied to shape our own custom-tailored Convolutional Neural Network architecture, \emph{ZynqNet CNN}.

\section{Convolutional Neural Network Analysis}
\label{sec:cnn-analysis}

Although research in the area of Convolutional Neural Network topologies is very active and new architectures emerge almost monthly, much of the attention seems to be currently focused on accuracy improvements, and much less on resource efficiency.
During our search for an optimized CNN, the lack of tools for visualizing, analyzing and comparing CNN topologies became a serious problem.
Therefore, we decided to develop the \emph{Netscope CNN Analyzer Tool for Visualizing, Analyzing and Modifying CNN Topologies}, which is introduced in the first \cref{sec:netscope}.
In \cref{sec:resource-efficient-cnn}, we define a wish-list of desired \emph{Characteristics of a Resource Efficient CNN Architecture}.
Finally, \cref{sec:cnn-comparison} employs the Netscope tool to analyze a number of different CNN topologies, and presents the findings from this \emph{CNN Topology Efficiency Analysis}.

\subsection{Netscope CNN Analyzer}
\label{sec:netscope}

The structure of Convolutional Neural Networks is inherently multi-dimensional, which makes them difficult to grasp, analyze and optimize intuitively (see \cref{fig:fire-module-3d} in the appendix for a 3D illustration of the convolutional layers in a simple 2-layer CNN module). Because no proper tools for the analysis of CNNs were available,\footnote{
\caffe includes a python script ``draw\_net.py'', which draws the network structure but doesn't do any analysis and Excel tables tend to either explode or disintegrate after a short time.}
we decided to implement our own custom solution for the \emph{Visualization, Analysis and Modification of CNN topologies}, based on an existing tool for the visualization of CNNs~\cite{netscope-ethereon}.

The \emph{Netscope CNN Analyzer} is a web-based tool written in CoffeScript, HTML and CSS for analyzing data flows and memory requirements in CNNs, and currently has these features:

\begin{itemize}
  \item In-browser editor for CNN descriptions with syntax highlighting
  \item Load \caffe\ \texttt{.prototxt} files from GitHub Gists, from built-in presets or by copy-paste\footnote{All visualizations and analyses are calculated locally, the network description never leaves the computer.}
  \item Visualization of the layer-level CNN structure as network graph
  \item Visualization of layer type, settings and dimensions\footnote{Supported layer types (visualization + analysis): \texttt{DATA}, \texttt{CONVOLUTION}, \texttt{INNER\_PRODUCT}, \texttt{POOLING}, \texttt{BATCHNORM}, \texttt{LRN}, \texttt{CONCAT}, \texttt{RELU}, \texttt{DROPOUT}, \texttt{SOFTMAX}, \texttt{SOFTMAX\_LOSS}, \texttt{FLATTEN}, \texttt{ELTWISE}, \texttt{DECONVOLUTION}, \texttt{CROP}, \texttt{SCALE}, \texttt{IMPLICIT}.}
  \item Analysis of computational complexity and resource requirements in each layer:
  \begin{itemize}
    \item Number of operations: multiply-accumulate\footnote{One multiplication and one addition are counted as a single MACC operation.} (macc), comparison (comp), addition/subtraction (add), division (div), exponentiation (exp)
    \item Memory requirements: size of output feature maps (activation), number of weight parameters (param)
  \end{itemize}
  \item Report of analysis results in summarized and detailed table
  \item Report of layer characteristics in Excel-compatible format for further analysis
\end{itemize}

\begin{figure}[tb]
  \centering
  \setlength{\fboxrule}{0.5pt}
  \framebox{ \includegraphics[height=5.2cm]{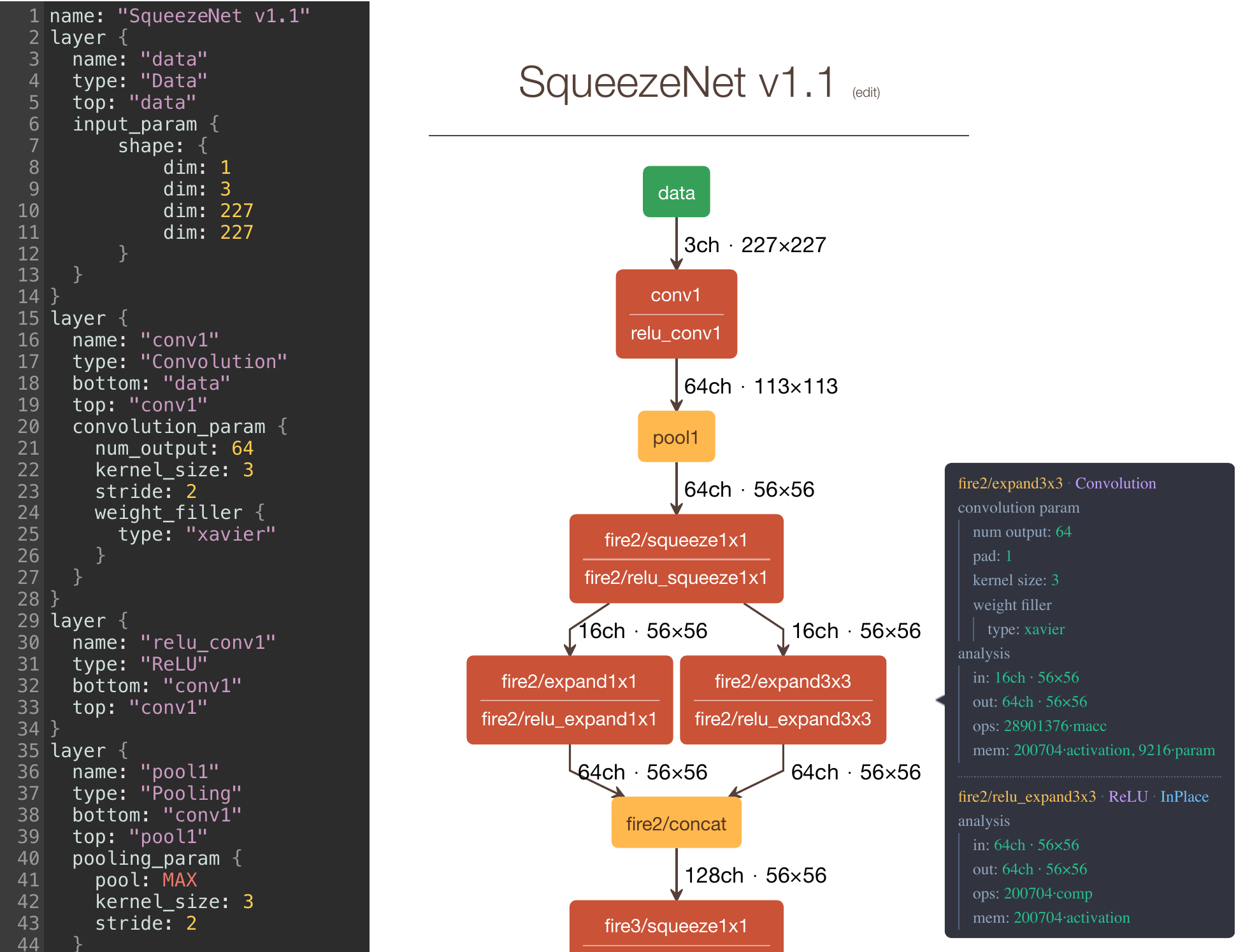} }\quad
  \framebox{ \includegraphics[height=5.2cm]{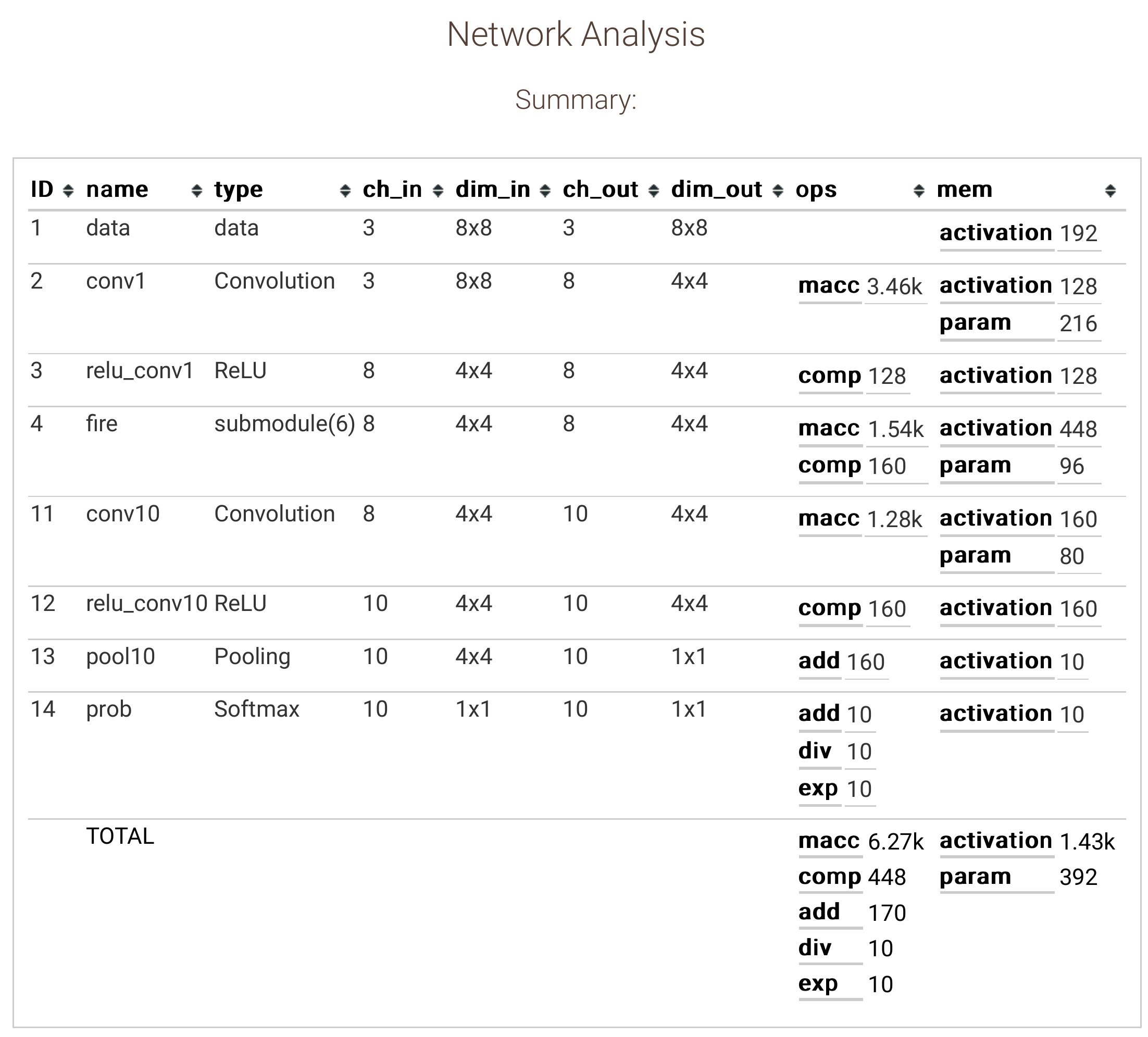} }
  \caption[Screenshots of the Netscope CNN Analyzer]{Screenshots from the Netscope CNN Analyzer: Visualization of the layer-level Network Graph (left), Analysis Summary Table (right)}
  \label{fig:netscope-screenshots}
\end{figure}

\Cref{fig:netscope-screenshots} shows two screenshots of the user interface and \cref{sec:netscope-advanced-usage} lists advanced usage tips as well as current restrictions. Netscope is accessible online~\cite{netscope} and includes presets for all the CNN topologies introduced in \cref{sec:cnn-topologies}.
The full source code for Netscope is available on Github~\cite{netscope-github}.




\subsection{Characteristics of Resource Efficient CNNs}
\label{sec:resource-efficient-cnn}
Convolutional Neural Networks are very demanding with respect to their computational complexity. In order to reach acceptable real-time performance with the resources available on the chosen embedded platform, a highly optimized Convolutional Neural Network architecture is mandatory. While a number of factors influence the resource efficiency of neural network topologies, the following characteristics are especially desirable:

\begin{description}
  \item[Low Computational Complexity] The Zynqbox platform constitutes a relatively small and low-power target (the Zynq XC-7Z045 has an upper bound of \SI{468}{GFLOP/s}~\cite{xilinx-gflops}, which is however not realistically reachable~\cite{fpga-gflops}). In addition, real-time inference is one of the project objectives, and while no required frame-rate is specified in the goals, around \SI{10}{FPS} might be a realistic target for many applications. This sets a hard upper bound of 23 billion MACCs per forward pass assuming a perfect accelerator, and makes especially small CNNs with low computational complexity attractive.

  \item[Regularity] FPGAs are very good at processing highly parallel and regular workloads. Those can be distributed and concurrently computed on many parallel yet simple processing elements, ideally in a dataflow manner. Conditional execution of operations, data-dependent decisions and complex control sequences are better suited for other platforms. Therefore, highly regular CNNs are preferred. Problematic structures include Batch Normalization and LRN layers (where different output maps influence each other), convolutional layers with many different kernel sizes (which may need to be accelerated in different ways) and overly complex network graphs.

  \item[All-Convolutional Networks] A network that consists only of convolutional layers and does not contain any fully-connected layers is called \emph{all-convolutional}. All-convolutional networks need less memory bandwidth: while weights in convolutional layers are reused multiple times, fully-connected layers need to load a new weight for every single multiply-accumulate operation. Because the memory bandwidth in the Zynq XC-7Z045 FPGA is limited,\footnote{
     The Zynqbox can access its \SI{1}{\giga\byte} of shared 32-bit DDR3-1066 memory at approximately \SI{8}{\giga\byte/\second}~\cite{zynqbox}.
     Transferring the \SI{470}{\mega\byte} of weights in the last three fully-connected VGG-16 layers then already requires more than \SI{50}{\milli\second}.
  } the higher computation-to-communication ratio found in all-convolutional CNNs is very welcome. Furthermore, all-convolutional networks can eliminate the need for pooling layers as shown by Springenberg et al.~\cite{all-convolutional}, which increases their regularity.

  \item[Accuracy] Despite all these constraints, the Image Classification CNN should still deliver top quality results, and should be optimized with respect to its classification accuracy.
\end{description}

With this wish-list of CNN characteristics in mind, the following section looks at different CNN topologies from prior work, and judges their suitability for our embedded implementation.

\subsection{Efficiency Analysis of Prior CNN Topologies}
\label{sec:cnn-comparison}

\begin{figure}[tb]
  \centering
  \includegraphics[width = \linewidth]{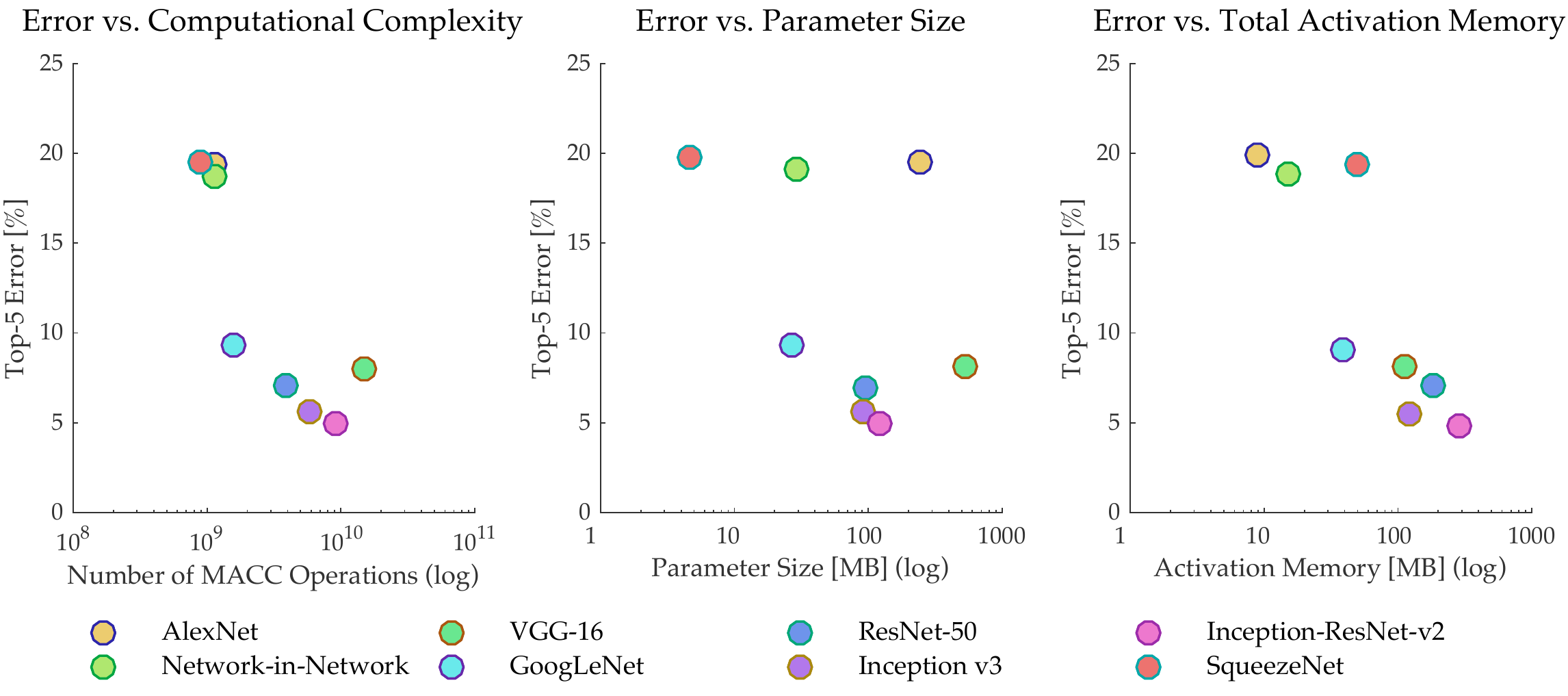}
  \caption[Design Space Exploration of CNN Topologies from Prior Work]{Design Space Exploration Charts, comparing the Top-5 Error Rate to the Number of MACCs, Parameter Size and Total Activation Memory for each of the CNN Topologies from Prior Work. Parameter and Activation Memory are calculated for 32-bit weights and activations. The x-Axes are in logarithmic scale.}
  \label{fig:topology-comparison-dse}.
\end{figure}

In search of an efficient CNN architecture, the CNN topologies from prior work introduced in \cref{sec:cnn-topologies} have been analyzed using Netscope. \Cref{fig:cnn-topologies-visualization} in the appendix shows the network graph visualizations for \emph{AlexNet}, \emph{Network-in-Network}, \emph{VGG-16}, \emph{GoogLeNet}, \emph{ResNet-50}, \emph{Inception v3}, \emph{Inception-ResNet-v2} and \emph{SqueezeNet}. A comparison of the architectures in terms of computational complexity, memory requirements and classification accuracy has been shown in \cref{tab:topology-summary} in \cref{sec:cnn-topologies}. The most relevant of these characteristics are visualized in the Design Space Exploration charts in \cref{fig:topology-comparison-dse}.

The Design Space Exploration (\cref{fig:topology-comparison-dse,tab:topology-summary}) shows that AlexNet, NiN, and Squeeze\-Net have the lowest computational complexities (approximately 1 billion MACCs), closely followed by GoogleNet (1.6 billion MACCs).
The same four CNNs also use the least amount of Activation Memory (measured as the aggregate size of all Output Feature Maps). When looking at the number of Parameters, the clear winner is SqueezeNet, which is almost 6\x\ smaller than GoogLeNet and 50\x\ smaller than AlexNet. However, GoogLeNet achieves \SI{9.2}{\%} top-5 error while SqueezeNet has \SI{19.7}{\%} top-5 error.

Most state-of-the-art CNNs reach their high performance at the price of exponentially higher computational complexity and exponentially increased memory requirements (as seen by the quasi-linear distributions in the semi-log comparison graphs). VGG-16 is always in a pareto suboptimal position. AlexNet can almost always be replaced by the smaller SqueezeNet. In the Netscope topology visualization (\cref{fig:cnn-topologies-visualization} in the appendix), GoogLeNet, ResNet and the Inception variants stand out with their architectural complexity, while AlexNet, NiN, VGG-16 and SqueezeNet look relatively compact and regular. Furthermore, NiN, VGG-16 and SqueezeNet are the only networks without Batch Normalization and LRN layers.

The final choice was made for SqueezeNet as the basis for our own CNN topology, due to its good fit for an FPGA-based implementation. The tiny parameter set could even be fit into the on-chip SRAM of a medium-sized FPGA, and optimizations are relatively easy to try thanks to the fast training cycles and the clear network structure.

\clearpage\section{Convolutional Neural Network Training}
\label{sec:network-training}

With the network topology fixed to SqueezeNet, we decided to set up a training environment for this CNN.
The following sections describe the \emph{hardware and software} used (\cref{sec:training-hwsw}), gives an introduction on \emph{how to prepare a dataset and a CNN} for a successful training run (\cref{sec:training-howto}),
and finishes with some \emph{tips and tricks} learned during more than 2200 GPU-hours of CNN training (\cref{sec:training-tips}).

\subsection{Construction of a GPU-based Training System}
\label{sec:training-hwsw}

\paragraph{First Steps} The first experiments were conducted with the \emph{Torch} framework on a workstation with an NVidia Tesla C2075 graphics card. Torch proved surprisingly tough to install, yet very flexible and powerful to use. However, only few network descriptions were readily available online. The next Deep Learning framework installed was \caffe. Using the \caffe binaries directly proved tedious, because the training progress needs to be tracked from verbose log files and each training run has to be prepared by creating scripts, editing multiple \texttt{.prototxt} files, and starting tasks manually once the GPU becomes available.

\paragraph{DIGITS Training Software} Many of these difficulties can be resolved by using NVidia's Deep Learning GPU Training System (DIGITS)~\cite{digits}, an open-source software package, which includes \caffe and Torch and can be controlled from a web-based interface. DIGITS allows the creation of datasets, the definition of CNN models, and the launch of multi-GPU training runs with extensive visual progress reports and an automatic scheduler for pending jobs. Many concurrent training runs and models can be created, compared and managed, and the weights and activations in trained CNNs can be visualized (see \cref{fig:digits_screenshots} in the appendix for screenshots of these interfaces). By accepting \texttt{.prototxt} model descriptions and internally using \caffe for the training, DIGITS retains much of the flexibility and performance while significantly simplifying the handling of multiple CNN architectures and GPUs.

\paragraph{Workstations for CNN Training} The training performance with the NVidia Tesla C2075 graphics card soon proved to be unsatisfying, mostly because of its incompatibility with NVidia's optimized CNN libraries.\footnote{
The \emph{cuDNN} library contains optimized algorithms for the calculation of Pooling, ReLU, LRN, Batch Normalization and Convolutional Layers and provides substantial speedups, but requires CUDA compute capability 3.0.}
NVidia offers a preconfigured quad-GPU Deep Learning workstation~\cite{nvidia-devbox} at a price of \SI{15000}{\$}, but we decided to build our own dedicated workstations for CNN Training. Because the system performance during CNN training is mostly determined by the number of CUDA cores and the amount of memory available in the graphics card, we decided to use a dual-GPU setup based on the NVidia GeForce GTX Titan X, the most powerful workstation graphics card at the time. In order to exploit a dual-GPU setup, a motherboard with at least two PCIe 3.0 x16 slots running both at least in x8 mode is required. The chosen Gigabyte Z170XP-SLI motherboard would support up to 4 GPUs in parallel. \caffe fully utilizes one CPU core per training process, so at least a dual-core processor is needed, even though its performance is not critical. The Titan X GPUs both have \SI{12}{\giga\byte} of graphics memory, which should at least be matched by the system's main memory.\footnote{Memory accesses during training are mostly linear, therefore SDRAM latency and CPU Cache size are not especially important for the system performance.} A fast solid-state disk is necessary to avoid a bottleneck when loading training samples, so a \SI{500}{\giga\byte} S-ATA III SSD has been chosen for the storage of models and training data. Additionally, at least a \SI{700}{\watt} power supply and a case provisioning reliable cooling is necessary~\cite{deep-learning-hw-quora,deep-learning-hw-blog}. In the appendix, \cref{tab:training-hw-list} lists the hardware components used in our setup, and \cref{fig:titan-x-hardware} shows a photograph of the assembled workstation. Two of these dual-GPU workstations have been assembled and named \emph{Rhea} and \emph{Kronos} after the Titans in Greek Mythology. The setup has proven its high performance and reliability during thousands of GPU-hours already.

\subsection{CNN Training with \caffe and DIGITS}

Successfully training CNNs requires experience and is even considered ``more art than science'' (Matthew Zeiler, winner ILSVRC 2013~\cite{cnn-training-art}).
The \caffe and DIGITS installations on the CNN Training Workstations have been used to train more than 70 Convolutional Neural Network variants during this project.

Before training, a suitable dataset with training samples has to be prepared. Optionally, this dataset can be artificially enlarged using \emph{data augmentation}, which increases the variance in the data samples and can help to improve the network performance. Next, the CNN model itself needs to be defined, and the \emph{solver}, which is responsible for the model optimization, needs to be configured. Especially this solver configuration requires the choice of several \emph{hyperparameters} that heavily influence the learning process. There are no unique valid settings, and intuition as well as experience are necessary to find a good combination.
Section \ref{sec:training-howto} in the appendix gives an overview of the training process with DIGITS, and a number of tips and tricks for the successful training of Convolutional Neural Networks.

\section{Network Optimization}
\label{sec:cnn-optimization}

Three types of optimizations have been applied while transforming the original SqueezeNet into ZynqNet CNN: \emph{efficiency-related optimizations} (detailed in \cref{sec:optim-bloats}), \emph{FPGA-related optimizations} (introduced in \cref{sec:optim-fpga}), and \emph{accuracy-related} optimizations (presented in \cref{sec:optim-accuracy}).

%
\subsection{Optimizations for Efficiency}
\label{sec:optim-bloats}

The original SqueezeNet v1.0 architecture has been published in February 2016 \cite{squeezenet} and can already be considered a highly optimized topology. Nonetheless, we have discovered some general opportunities for improvement during our experiments. Beneficial modifications have also been discovered by the authors of SqueezeNet and have led to the publication of SqueezeNet v1.1 on April 25, 2016~\cite{squeezenet-v11}. These two networks form the basis of ZynqNet.

\paragraph{Structural Analysis of SqueezeNet}
As shown in the Netscope visualization in \cref{fig:squeezenet-zynqnet-visualization}, SqueezeNet has a relatively regular structure. The topology consists of an initial convolutional layer, eight stacked \emph{fire modules}, and a last convolutional layer, with three max-pooling layers and a dropout layer interposed. All convolutional layers are complemented by ReLU nonlinearities, and the topology is finished with a global average-pooling layer. Each fire module consists of a \emph{squeeze layer} (1\x1 convolutions plus ReLU) and two parallel \emph{expand layers} (1\x1 and 3\x3 convolutions plus ReLU). The squeeze layer has relatively few output channels and is responsible for compressing the internal representation. The expand layers evaluate both 1\x1 and 3\x3 kernels on this compressed feature map, and their outputs are concatenated along the channel dimension (\cref{fig:fire-module-3d} in the appendix shows a 3D illustration of a single fire module with 16 squeeze channels and 64+64 expand channels).

\begin{figure}[tb]
  \centering
  \includegraphics[width = \linewidth]{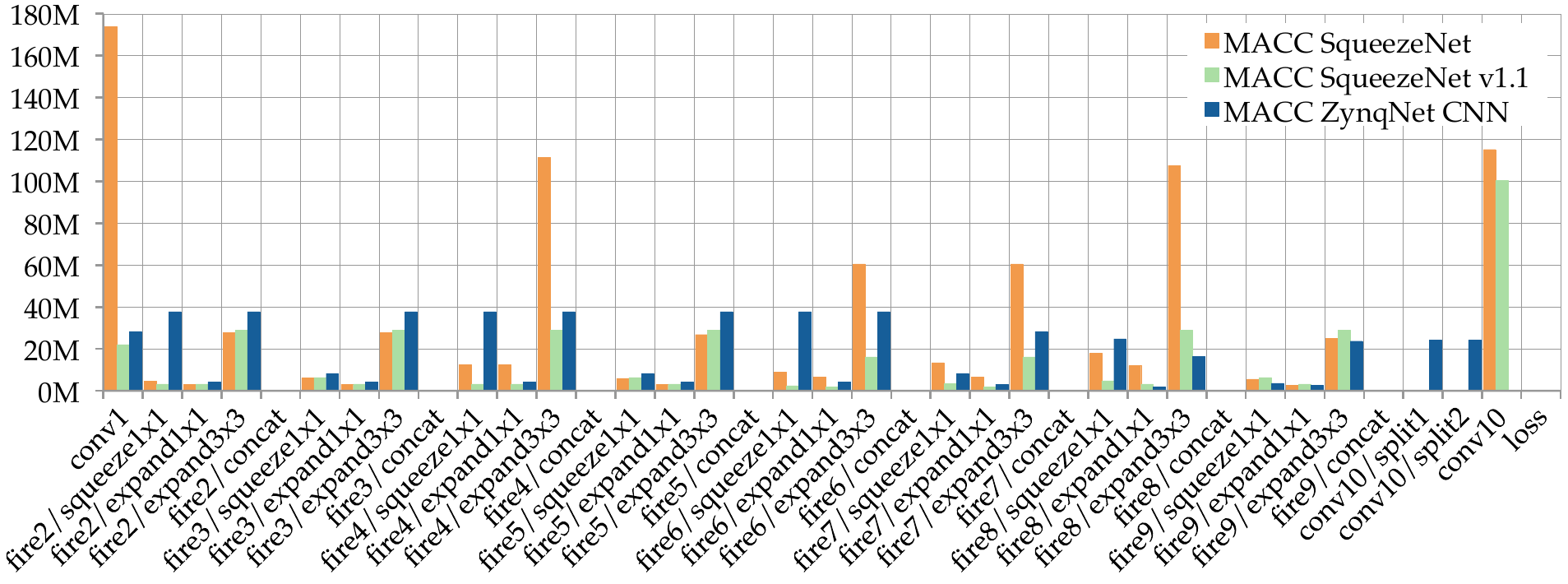}
  \caption[SqueezeNet and ZynqNet Computational Complexity Analysis]{Per-Layer Computational Complexity (Number of Multiply-Accumulate Operations) for SqueezeNet, SqueezeNet v1.1 and ZynqNet CNN.}
  \label{fig:squeezenet-complexity}
\end{figure}

\paragraph{SqueezeNet Complexity Analysis}
The computational complexity of each individual layer in SqueezeNet and ZynqNet CNN has been analyzed with Netscope, and the results are visualized in \cref{fig:squeezenet-complexity}.
The most expensive layer in SqueezeNet is \emph{conv1}, the initial 7\x7 convolutional layer (\SI{20}{\%} of all MACC operations). The final 1\x1 convolutions in \emph{conv10} and the 3\x3 convolutions in \emph{fire4} and \emph{fire8} (each approximately \SI{13}{\%}) are also disproportionately expensive.

\begin{figure}[tb]
  \centering
  \includegraphics[height = 4.75cm]{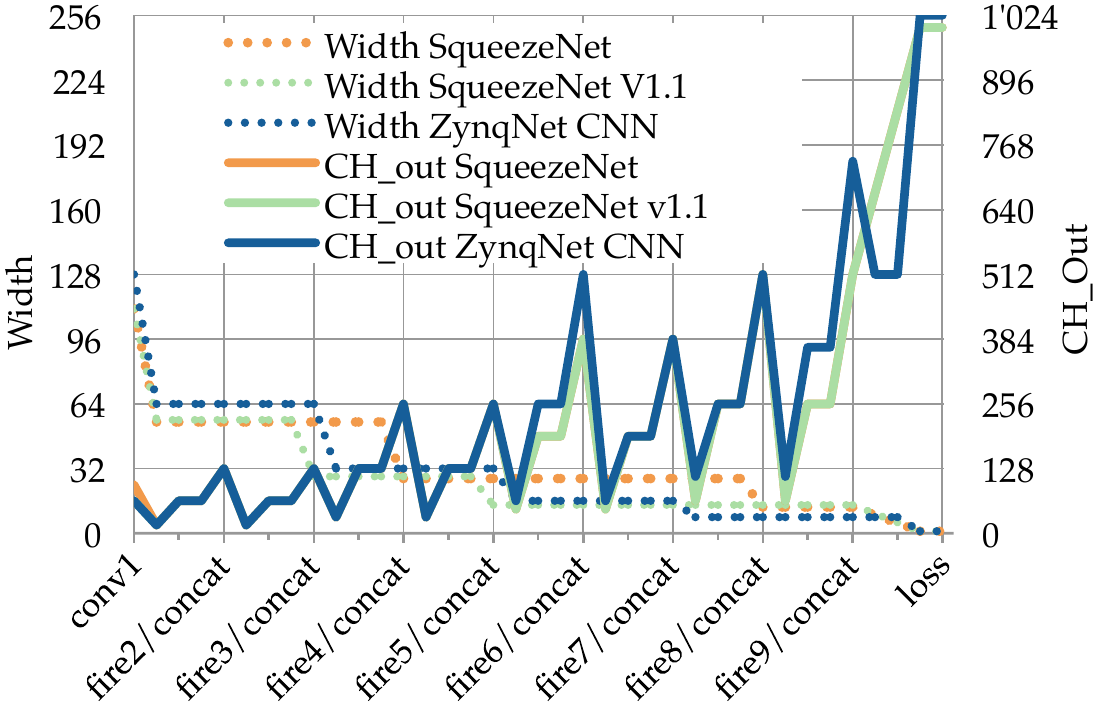}\hfill
  \includegraphics[height = 4.75cm]{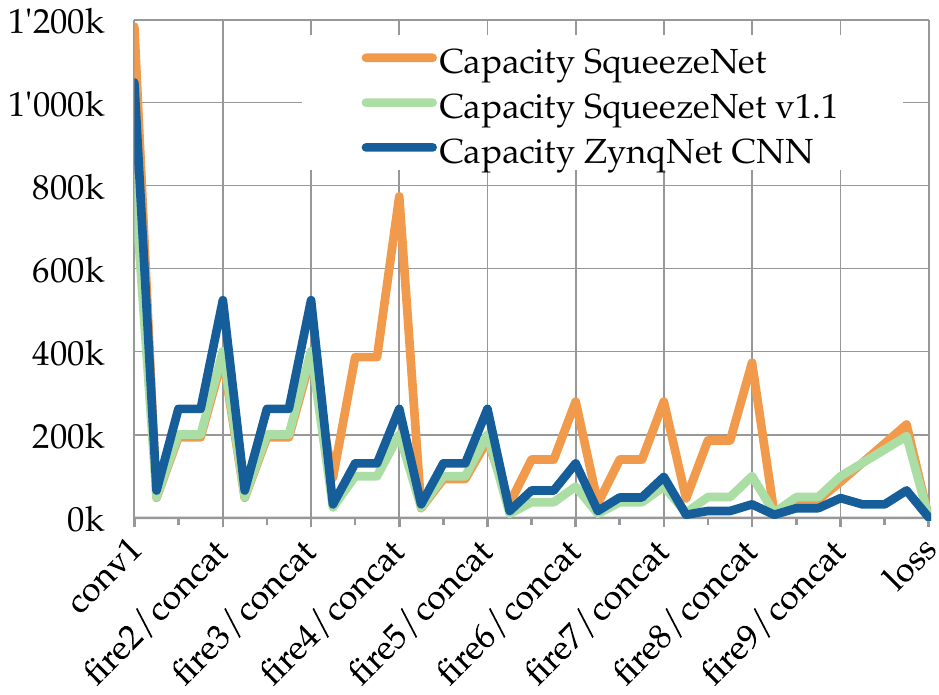}
  \caption[SqueezeNet and ZynqNet Capacity and Dimension Analysis]{
    Per-Layer Dimension Analysis of SqueezeNet, SqueezeNet v1.1 and ZynqNet CNN.
    Left: Layer Widths \wout (primary axis) and Output Channels \chout (secondary axis). Because the number of output channels in SqueezeNet and SqueezeNet v1.1 is mostly equivalent, their curves overlap.
    Right: Layer Capacities $\wout\times\hout\times\chout$. }
  \label{fig:squeezenet-dimensions}
\end{figure}

\paragraph{Out-of-Sync Dimension Adjustments} \Cref{fig:squeezenet-dimensions} takes a closer look at the layer capacities $\wout\times\hout\times\chout$, the layer widths \wout and the number of output channels \chout in each stage of the network.
Here we can see how the spatial output dimensions are periodically stepped down (using stride 2 in layers \emph{conv1}, \emph{pool1}, \emph{pool4}, \emph{pool8}, and using global pooling in \emph{pool10}).
The number of output channels is periodically increased, while the internal ratio of output channels in squeeze and expand layers is kept constant at $1:4:4$.
However, the spatial shrinking and the channel-wise expansion in the original SqueezeNet are not ideally synchronized, and \emph{fire4} as well as \emph{fire8} increase the number of output channels before decreasing the pixel count, leading to a surge in computational complexity. By decreasing the spatial dimensions earlier, both SqueezeNet v1.1 and ZynqNet CNN solve this problem. The modification saves up to \SI{40}{\%} in activation memory and reduces the computational complexity in \emph{fire4} and \emph{fire8} by a factor of 3.7 and 3.9 respectively.

\paragraph{7\x7 Convolutional Input Layer} Using a convolutional input layer with a large kernel size and a large stride is typical for many CNNs (e.g. GoogLeNet, AlexNet and ResNet) and gives the network a large receptive field in the first layer. However, the large filter dimensions are computationally expensive: A 7\x7 filter requires $5.4\times$ more MACC operations than a 3\x3 filter. As long as the learned filters are well-behaved, a 7\x7 kernel can be approximated by three stacked 3\x3 kernels, which have the same receptive field, but only need $\sfrac{27}{49}$ of the computations. At the same time, the stacked 3\x3 filters are more expressive thanks to the additional nonlinearities~\cite{cs231n-learning-schedule}. Interestingly, the accuracy in SqueezeNet dropped by less than \SI{1}{\%} when we simply replaced the 7\x7 kernels in \emph{conv1} with 3\x3 kernels. Further tests were made with 5\x5 and 11\x11 filters, as well as combinations of multiple 3\x3 conv layers such as (3\x3)/16\x(3\x3)/16\x(3\x3)/96 (which increased both accuracy and training time). See \cref{sec:appendix-training-overview} for an overview of all experiments conducted. The final decision was made for a single 3\x3 convolutional input layer with 64 output channels, similar to the one used in SqueezeNet v1.1.

\paragraph{Unnecessary Padding} The original SqueezeNet used \texttt{pad=1} in the 1\x1 conv layer \emph{conv10}. Padding makes no sense for 1\x1 kernels, and setting \texttt{pad=0} gives exactly the same results while saving \SI{33}{\%} of the MACC cycles in \emph{conv10}.

%
\subsection{Optimizations for FPGA Implementation}
\label{sec:optim-fpga}

The CNN architecture has been adapted to the FPGA requirements concurrently with the work on the FPGA-based accelerator. Most changes aim to simplify the network architecture, or make sure that all layers fit into the accelerator memory.

\paragraph{Power-of-2 Layer Dimensions} Most CNNs trained on ImageNet expect either 227\x227 or 224\x224 pixel images as input.\footnote{
  The idea is to enable repeated stride-2 downscaling to integer dimensions (e.g. $224/2/2/2/2/2 = 7$). AlexNet produces non-integer intermediate dimensions no matter if input images are 227\x227 or 224\x224 pixels, possibly due to a missing \texttt{pad=5} in the first conv layer --- reasonable padding settings help to avoid confusion.}
ZynqNet CNN however has been designed with spatial layer dimensions $w$ and $h$ which are a power of 2, such as 8, 16, 32, 64, 128 and 256. On the FPGA, multiplications and divisions by a power of 2 can be calculated with inexpensive shift operations, which enables optimizations in the addressing of image caches in the accelerator. The number of channels $ch$ was initially rounded to powers of 2 as well, but the final ZynqNet CNN uses multiples of 16 instead to make better use of the available resources. The CNN architecture with all-power-of-2 dimensions required \SI{14}{\%} more parameters and \SI{9}{\%} more MACC operations, but also reached \SI{1}{\%} higher accuracy. The adapted CNN expects 256\x256 pixel input images and is trained on a 300\x300 pixel ImageNet dataset.

\paragraph{All-Convolutional Network} In 2014, Springenberg et al. published a paper that is well-known for the introduction of the ``guided backpropagation'' method for visualizing CNN filters \cite{all-convolutional}. The same paper also introduced the idea of \emph{all-convolutional networks}, which are CNNs consisting exclusively of convolutional layers and nonlinearities.\footnote{Note the distinction between \emph{fully-connected layers} (layers where all inputs are connected to all neurons) and \emph{all-convolutional networks} (CNNs which do \emph{not} contain fully-connected, but only convolutional and nonlinearity layers).}  The authors tested networks where all max-pooling layers had been replaced by convolutional layers with stride 2, and reached state-of-the-art accuracy.
Based on this idea, we removed all max-pooling layers in our CNN, and used \emph{stride 2 in the subsequent convolutional layer}. However, all max-pooling layers in SqueezeNet are followed by 1\x1 \emph{squeeze} convolutions and stride 2 would thus waste a lot of information. We decided to increase the kernel size in these layers to 3\x3 to allow for overlapping convolutions with stride 2. The resulting all-convolutional CNN has \SI{12}{\%} more parameters and requires \SI{18}{\%} more MACC operations, but also reaches \SI{1.5}{\%} higher accuracy. In addition, the CNN architecture is strongly unified, leaving the global average pooling as the only other layer type besides 1\x1 and 3\x3 convolutional layers and their ReLU nonlinearities.\footnote{The dropout layer only needs to be considered during training.}

\paragraph{Layer Splitting} One of the most limited resources on the FPGA is on-chip memory, which is used to hold the current layer parameters. The FPGA fabric in the Zynq XC-7Z045 contains a total of \SI{2180}{\kilo\byte} Block RAM memory~\cite{zynq-datasheet}, which is enough to hold approximately \num{560000} 32-bit floating-point parameters. However, the \emph{conv10} layer in ZynqNet CNN has been designed with $\chin=736$ input channels and $\chout=1024$ output channels, and would therefore require $n=\chin\cdot\chout=\num{753664}$ kernels of size 1\x1. To make the layer fit onto the FPGA, it has been split into two parallel convolutional layers \emph{conv10/split1} and \emph{conv10/split2} with $\chout=512$, which are then concatenated along the channel dimension.\footnote{The required facilities for the concatenation are already present from the parallel \emph{expand} layers in each fire module.}

\subsection{Optimizations for Accuracy}
\label{sec:optim-accuracy}

The final type of optimizations in ZynqNet targets the classification accuracy. Multiple previous optimizations already resulted in accuracy improvements, such as replacing the max-pooling layers with 3\x3 stride 2 convolutions (+\SI{1.9}{\%}) and the power-of-2 layer dimensions (+\SI{1}{\%}). Three additional measures are introduced in this section.

\paragraph{Linear Learning Rate Policy} As already mentioned in \cref{sec:training-tips}, experiments by Mishkin et al. \cite{ducha-aiki} have shown that a linear learning rate policy works best for AlexNet. They found the same to be true for SqueezeNet, which initially used a square-root learning rate policy~\cite{squeezenet-linear-policy}. The accuracy improvement is approximately \SI{2}{\%}.

\paragraph{Equalization of Layer Capacities} Intuitively, a CNN can be understood to transform a vast amount of pixels with low individual information density into very few outputs of high abstraction level. The layer capacity $\wout\times\hout\times\chout$ can be seen as a measure for this concentration of information. As shown in \cref{fig:squeezenet-dimensions}, the layer capacities of SqueezeNet, SqueezeNet v1.1 and ZynqNet all converge from more than one million data points to just \num{1000} class probabilites. However, both SqueezeNet variants have intermediate capacity peaks which do not follow a smooth decline (besides the typical zig-zag pattern caused by the compression-expansion architecture that can be seen for all three CNNs). SqueezeNet~v1.0 has pronounced outliers in the \emph{fire4} and \emph{fire8} modules, which have already been discussed as \emph{Out-of-Sync Dimension Adjustments} in \cref{sec:optim-bloats}. Further, both SqueezeNet versions have a strong peak in \emph{conv10}. ZynqNet CNN follows a much smoother and more regular capacity reduction, which saves resources, but also increases accuracy by almost \SI{2.3}{\%}.

\paragraph{Augmented Dataset and Extended Training Runs} Data augmentation, previously mentioned in \cref{sec:training-howto}, has been patched into DIGITS and used for the final training runs of ZynqNet. The ImageNet dataset has been prepared as follows:
\begin{itemize}
  \item 300\x300 pixel images, 256\x256 pixel crops
  \item 6 copies per input image (total 7.7 million training examples)
  \item hue modulation ($\pm60^{\circ}$, $p=0.75$)
  \item contrast modulation ($0.5\times$ to $1.5\times$, $p=0.75$)
\end{itemize}
These settings add a substantial amount of variation to the images, and were chosen to approximately emulate the reduced quality of webcam images, preparing the network for actual input images during demonstrations.
In addition to the increased amount of images in the augmented dataset, the final trainings were run for 60 epochs instead of 30 epochs, effectively showing the network each image 60 times in 6 variations. This resulted in another gain of \SI{3.1}{\%} accuracy.

\paragraph{Fine-Tuning} Final experiments were conducted with fine-tuning the trained model. By re-training the finalized network for a few epochs with a very low learning rate (and possibly with data augmentation turned off), sometimes a slightly better optimum can be reached. However, our best try resulted in just \SI{0.2}{\%} accuracy gain.

\subsection{Final Results}
Overall, the top-1 validation accuracy of our ZynqNet CNN has been increased by more than \SI{7}{\%} versus the initial SqueezeNet v1.0 and by more than \SI{8}{\%} versus the SqueezeNet v1.1 architecture.\footnote{The fact that the total accuracy improvement is less than the sum of the individual improvements indicates that some optimizations were not orthogonal and had similar effects.}
The final version of ZynqNet CNN uses 2.5 million parameters, roughly twice as many as the SqueezeNet variants, but still roughly an order of magnitude less than most other CNNs. The total number of activations has been reduced by \SI{40}{\%}, and the number of MACC operations by \SI{38}{\%} with regard to the original SqueezeNet, to 530 million activations. Additionally, neither max-pooling, Batch Normalization nor LRN layers are required.
The fully-trained CNN reaches a top-1 accuracy of \SI{63.0}{\%} and a top-5 accuracy of \SI{84.6}{\%} on the ImageNet validation dataset.
\chapter{FPGA Accelerator Design and Implementation}
\label{chap:fpga-design}

\cleanchapterquote{Good [HLS coding] style not only requires an under-\\standing of the underlying hardware architecture of an algorithm, so that it
is reflected in the \cpp\ design, but also an understanding of how HLS works.}{Mike Fingeroff}{(High Level Synthesis Expert, Mentor Graphics)}

%
\section{Introduction}
\label{sec:fpga-design-intro}

This chapter introduces the \emph{ZynqNet FPGA Accelerator}, which is a purpose-built FPGA-based accelerator for the ZynqNet CNN introduced in the last chapter.
After a quick overview of the \emph{Zynqbox Platform}, a consideration of different \emph{Data Types} for the accelerator and a formulation of the \emph{Design Goals} in this section, the following \cref{sec:algorithm} introduces the \emph{Algorithm} used for the evaluation of ZynqNet CNN, and details its \emph{Parallelization} and the \emph{Caching Strategy}.
\Cref{sec:block-diagram} then presents the \emph{Hardware Architecture and Schedule}, before \cref{sec:hls-design-flow} describes the \emph{Implementation} of the FPGA accelerator and our experiences with \emph{High-Level Synthesis}.

\subsection{Zynqbox Platform Overview}

\begin{figure}[tb]
  \centering
  \includegraphics[height=6.2cm]{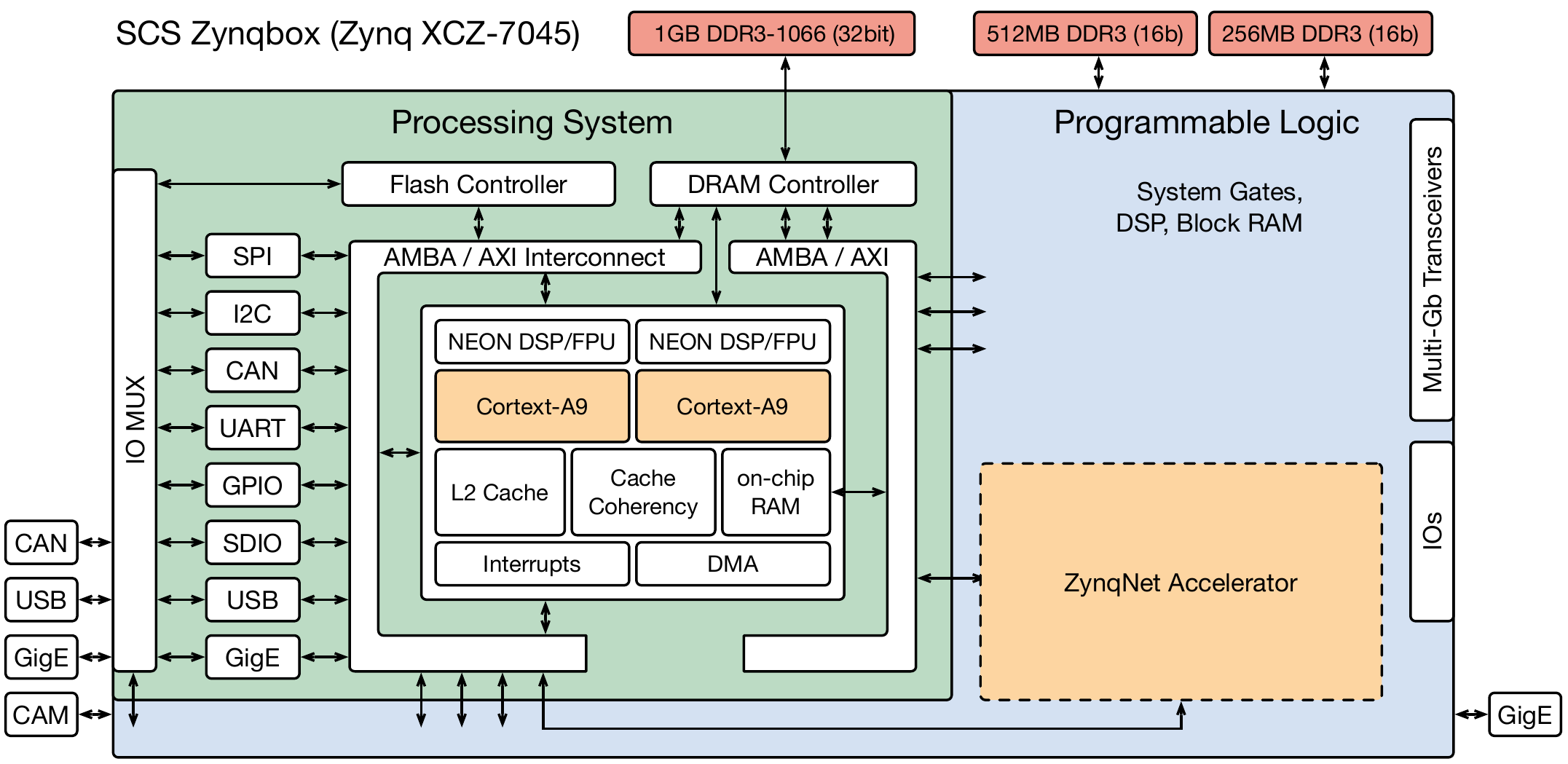}
  \caption[Schematic of the SCS Zynqbox Platform]{Schematic of the SCS Zynqbox Embedded Platform based on the Xilinx Zynq XC-7Z045.}
  \label{fig:zynqbox}
\end{figure}

The Zynqbox has been designed by Supercomputing Systems AG for the evaluation of high-performance image processing algorithms, especially in automotive settings. The embedded platform is based on the Xilinx Zynq-7000 All Programmable System-on-Chip (SoC), which combines a dual-core ARM Cortex-A9 processor with programmable FPGA fabric in a single device. The Zynqbox includes a Xilinx Zynq XC-7Z045 SoC, \SI{1}{\giga\byte} DDR3 memory for the ARM processor, \SI{768}{\mega\byte} independent DDR3 memory for the programmable logic, and plenty of connection options (Serial Camera Interfaces, USB, CAN, Gigabit Ethernet).
The Kintex-7 FPGA fabric of the SoC features 350k logic cells, 218k LUTs, \SI{2180}{\kilo\byte} Block RAM and 900 DSP slices. The CPU runs at up to \SI{1}{GHz}, boots a standard Linux operating system and is connected to the programmable logic via high-performance AXI4 ports for data exchange and control. \Cref{fig:zynqbox} shows a schematic overview of the Zynqbox platform~\cite{zynqbox}.

\subsection{Data Type Considerations}

The choice of suitable data types is an important step when implementing designs on FPGA and ASIC platforms. This section gives a quick introduction to the possible options and justifies our choice.

\paragraph{Floating-Point Data Format}
When real-valued numbers have to be represented in a computer, a floating-point data format is typically used. The single-precision floating-point format uses 32 bits, and stores numbers as a combination of \emph{sign} (1 bit), \emph{significand} (23 bit) and \emph{exponent} (8 bit). The data format can represent all numbers from approximately $-3.4\times10^{38}$ to $+3.4\times10^{38}$ with at least 6 significant decimal digits, which makes it very versatile~\cite{wiki-float}. However, the computational complexity of arithmetic operations on floating-point data is high, and specialized hardware (e.g. floating-point units) are usually needed.

\paragraph{Fixed-Point Data Format}
An alternative to the floating-point format is given by the fixed-point format. A fixed-point number format can be described by the Q-format specification $Qm.f$, where $m$ denotes the number of \emph{integer bits} and $f$ denotes the number of \emph{fractional} bits. The actual fixed-point number is stored as a normal signed integer in 2's complement format, with bit-width $1+m+f$. For example, the value $3.375$ can be stored in $Q2.5$ format within 8 bits as $011.01100_{2}$, and interpreted as $01101100_{2}/2^f = 108/32 = 3.375$. The format specification is required for the interpretation of the stored bits, however it is usually implicit and not stored with the value. The standard arithmetic units for integers can be used to calculate fast and efficiently with fixed-point numbers, and their range and precision can be adapted exactly according to the application's requirements.

\paragraph{Data Type Requirements for Convolutional Neural Networks}
Most CNN implementations use single-precision floating-point numbers for their weights and activations, arguably mostly because it is the standard data type on modern GPUs. As already shown in \cref{sec:network-compression} on network compression, CNNs are inherently very robust against the effects of limited numerical precision. Neither an enormous dynamic range nor very high precision are needed, and in the most extreme case even binary weights and activations can be sufficient to train and run a Convolutional Neural Network as shown by Courbariaux et al.~\cite{binaryconnect,binarynet}

\paragraph{Fixed-Point versus Floating-Point on the Zynqbox Platform}
Today's FPGAs typically do not contain specialized floating-point hardware.\footnote{A notable exception are the higher-end Arria-10 and Stratix-10 FPGA series by Altera which include hardened floating-point support in each DSP block~\cite{altera-fp}.} In the Zynq's programmable logic, each floating-point multiplication \emph{or} addition occupies two to three DSP slices as well as hundreds of look-up tables and flipflops, and limits the clock speed to a maximum of \SI{460}{\MHz}~\cite{xilinx-fp-resource-utilization}. On the other hand, a fixed-point multiplication \emph{and} addition (MACC) with 18-bit or smaller operands can be carried out by a single DSP slice at up to \SI{750}{\MHz}~\cite{xilinx-dsp48,zynq-datasheet}. As a result, the Zynq XC-7Z045 can reach up to \SI{1500}{GMACC/s} in fixed-point, but only \SI{468}{GFLOP/s} in floating-point~\cite{xilinx-gflops}.
An additional advantage of fixed-point numbers is their reduced memory requirement. Using 16-bit values doubles the number of weights and activations which can be stored on-chip, and even 8-bit values should be precise enough for many CNNs.

\paragraph{Fixed-Point Quantization with Ristretto}
Together with their paper from April 2016, Gysel et al. published a CNN approximation tool called \emph{Ristretto} \cite{ristretto}. The application converts the weights and activations in \caffe-based Neural Networks to fixed-point format, and automatically determines the number of integer and fractional bits which are necessary to avoid serious degradation of the resulting classification accuracy (the maximum allowed accuracy loss can be configured). The authors are able to quantize SqueezeNet to 8-bit weights and activations with an accuracy drop well below \SI{1}{\percent}.

\paragraph{Choice of Data Type}
Even though the fixed-point numbers have many advantages for an FPGA-based accelerator and can improve the quality of results significantly, we decided to use \emph{single-precision floating-point numbers} in this project.
The deciding factor was the wish to retain compatibility with the GPU-based \caffe version of ZynqNet CNN.
Incorporating the fixed-point quantization would have resulted in a higher project risk, more potential points of failure and increased debugging complexity.
Unfortunately, \emph{Ristretto} had not yet been published by the time of this decision, as it might well have changed this choice by heavily reducing the risk involved in a fixed-point implementation.\footnote{%
A conversion of the current ZynqNet FPGA Accelerator implementation from floating-point format to fixed-point would not be trivial because the memory and computational resource requirements have influenced architectural decisions. Nonetheless, a conversion should be feasible and remains a very important optimization of the current architecture. Approximate results for a potential 16-bit fixed-point version are also reported in \cref{chap:results}.
}


\subsection{Design Goals}
\label{sec:fpga-design-goals}

This project focuses on the proof-of-concept implementation of an FPGA-accelerated embedded CNN. First and foremost, the challenge in this chapter consists of
\begin{quote}
\itshape
fitting a complete CNN for image classification on ImageNet\\
onto the low-power Zynq~XC-7Z045 with decent performance.
\end{quote}
Performance refers to \emph{throughput} in this context, measured as the number of images classified per second ($\textrm{FPS}$).
In order to maximize the throughput, the CNN needs to be computed as fast as possible, which implies the following design goals for the algorithm and accelerator:
\begin{itemize}
  \item minimum number of operations and clock cycles
  \item minimum number of data relocations per classified image
  \item maximum possible clock rate
  \item maximum amount of parallelization and resource utilization (especially DSP Slices)
\end{itemize}
In addition to throughput, \emph{power efficiency} is a key characteristic, because both heat dissipation and input power are typically limited in an embedded system. Power efficiency can be measured as the number of images classified per energy consumed ($\textrm{Images}/\textrm{J} = \textrm{FPS}/\textrm{W}$).


\section{Algorithm Design}
\label{sec:algorithm}


\subsection{Requirements Analysis}

\begin{figure}[tb]
  \centering
  \includegraphics[angle=90,width = \linewidth]{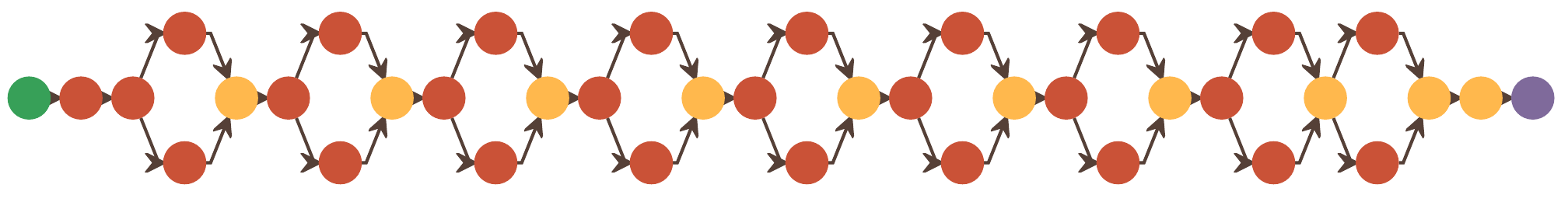}
  \caption[Topology Visualization of the ZynqNet CNN]{High-Level Visualization of the ZynqNet Topology. Red dots symbolize Convolutional Layers with ReLU Nonlinearities, yellow dots Concatenations or the final Average Pooling.}
  \label{fig:zynqnet-tiny}
\end{figure}

ZynqNet CNN, visualized in \cref{fig:zynqnet-tiny} and \cref{fig:squeezenet-zynqnet-visualization}, and detailed in \cref{tab:zynqnet-description} in the appendix, is a stripped-down version of SqueezeNet and consists exclusively of convolutional layers, ReLU nonlinearities and a global average pooling.
The network is highly regular, with most layers arranged in \emph{fire modules}.
Each fire module combines three convolutional layers: a \emph{squeeze} layer, followed by two parallel \emph{expand} layers.
The output channels of both \emph{expand} layers are concatenated to form a single feature map with twice as many output channels.
This ability to concatenate two layers is reused in convolutional layer \emph{conv10}, which is calculated in two separate splits \emph{conv10/split1} and \emph{conv10/split2} to reduce the memory requirements.
The dropout layer \emph{drop9} is only relevant during training, and can be completely ignored during inference.
\emph{Pool10} reduces the spatial dimensions from 8\x8 pixels to 1\x1 pixel by computing the mean, while leaving the channel dimension intact. Finally, a \emph{softmax} classifier is used to calculate the individual class probabilities.

The computational complexity in ZynqNet comes almost entirely from the 1\x1 and 3\x3 convolutions, which add up to 530 million MACC operations. The ReLU nonlinearities amount to 3 million comparisons. The average pooling requires \num{66000} additions and one division, and the final softmax executes 1024 exponentiations, additions and divisions.

Because the exponentiations and divisions are rare, the softmax layer can be readily handled by the ARM processor. This leaves the FPGA-based accelerator with the following layer types:
\begin{itemize}
  \item convolutional layers
  \begin{itemize}
    \item kernel size 1\x1, padding 0
    \item kernel size 3\x3, padding 1
    \item stride 1 or stride 2
  \end{itemize}
  \item ReLU nonlinearities
  \item concatenation
  \item global average pooling
\end{itemize}
These layer types need to be efficiently accelerated in order to successfully run the ZynqNet Embedded CNN on the FPGA.


\subsection{Algorithmic Options}

\subsubsection{The Mathematics behind Convolutional Layers}
\label{sec:math-formulation-convolution}

The central operation to be accelerated is the 2D convolution of multiple input feature maps with a number of small filter kernels.
The two-dimensional convolution of an input image and a filter can be intuitively understood as the result from sliding the filter over the input image, and taking the dot product between the filter and the pixels underneath at each possible filter position.
For a filter of size $k \times k$, each dot product $\mathbf{\langle A,B \rangle} = \sum_{i=0}^{n-1} A_i\cdot B_i = A_0\cdot B_0 + A_1\cdot B_1 + \cdots + A_{n-1}\cdot B_{n-1}$ requires $k^2$ multiplications and additions.
The 2D convolution between $k\times k$ filter $\mathbf{F}$ and $H\times W$ input image $\mathbf{I}$ yields output image $\mathbf{O}$ with
\begin{align}
\label{eq:conv2d-formula}
  \mathbf{O}_{(y,x)} =
  \sum_{j=-{\lfloor\sfrac{k}{2}\rfloor}}^{\lfloor\sfrac{k}{2}\rfloor} {
    \sum_{i=-{\lfloor\sfrac{k}{2}\rfloor}}^{\lfloor\sfrac{k}{2}\rfloor} {
      \mathbf{I}_{(y-j, x-i)} \cdot \mathbf{F}_{(j,i)}
  } }
\end{align}
under the assumptions that $k$ is an odd integer and the input image is appropriately zero-padded, i.e. $\mathbf{I}_{(y,x)} = 0$ for all pixels outside of the valid image area $W\times H$.
In convolutional layers there is not a single input image, but a three-dimensional stack of \chin input images called \emph{input feature maps} $\mathbf{I}^{(ci)}_{(y,x)}$. The convolutions then produce a stack of \chout output images, called the \emph{output feature maps} $\mathbf{O}^{(co)}_{(y,x)}$ by applying a bank of filters $\mathbf{F}^{(ci,co)}$. Under the above assumptions, a convolutional layer computes
\begin{align}
\label{eq:convlayer-formula}
  \mathbf{O}^{(co)}_{(y,x)} =
  \sum_{ci=0}^{\chin-1} {\left(
    \sum_{j=-{\lfloor\sfrac{k}{2}\rfloor}}^{\lfloor\sfrac{k}{2}\rfloor} {
      \sum_{i=-{\lfloor\sfrac{k}{2}\rfloor}}^{\lfloor\sfrac{k}{2}\rfloor} {
        \mathbf{I}^{(ci)}_{(y-j, x-i)} \cdot \mathbf{F}^{(ci,co)}_{(j,i)}
    } }
  \right)}
  = \sum_{ci=0}^{\chin-1} {
      \mathbf{\langle}
        \mathbf{I}^{(ci)}_{
          \binom
            {y+\lfloor\sfrac{k}{2}\rfloor\, \ldots\, y-\lfloor\sfrac{k}{2}\rfloor}
            {x+\lfloor\sfrac{k}{2}\rfloor\, \ldots\, x-\lfloor\sfrac{k}{2}\rfloor}
        },
        \mathbf{F}^{(ci,co)}
      \mathbf{\rangle}
  }
\end{align}
for every output pixel $(y,x)$ and every output channel $co$, which amounts to a total of $n_{\mathrm{MACC}}=H\times W\times \chin \times \chout \times k^2$ multiplications and accumulations. Despite requiring a high computational effort, the mathematical operations behind convolutional layers are not complex at all, and offer a lot of opportunities for data reuse and parallelization, which will be explored in the next section.\footnote{Padding and strides larger than 1 add some complications, but the overall operation stays the same.}

\subsubsection{Different Approaches to Calculating 2D Convolutions}
When it comes to the calculation of the convolutional layers, there are two other approaches besides the direct ``sliding-filter'' method described above.

\paragraph{Matrix Multiplication} The first approach transforms the 2D convolution into one large \emph{matrix multiplication}.
For this, each local input region (the image region underneath each possible filter location) is stretched out into a column vector, and all the column vectors are concatenated to form a matrix $\mathbf{C}$.
Since the filter's receptive fields usually overlap, every image pixel is replicated into multiple columns of $\mathbf{C}$.
The filter weights are similarly unrolled into rows, forming the matrix $\mathbf{R}$.
The 2D convolution is then equivalent to a matrix product $\mathbf{R}\mathbf{C}$, which can be calculated very efficiently using highly optimized linear algebra (BLAS) routines which are available for CPUs, GPUs and DSPs.
The disadvantage of this approach is the exploding memory consumption of the column matrix~\cite{cs231n-convnets}. For a small 3\x3 filter, matrix $\mathbf{C}$ is already blown up by a factor of 9 compared to the original input image. This makes it necessary to split the problem into a number of overlapping tiles, and later stitch the results back together, which artificially increases the complexity of the problem.
On FPGAs and in ASICs, matrix multiplications can be efficiently implemented with a \emph{systolic architecture}.
A suitable systolic array consists of a regular grid of simple, locally-connected processing units. Each of them performs one multiplication and one addition, before pushing the operands on to their neighbors.
Thanks to the locality of computation, communication and memory, these architectures are very hardware-friendly~\cite{systolic-mmult}.

\paragraph{Fast Fourier Transformation}
The second approach to 2D convolutions makes use of the fact that a convolution in the Spatial Domain corresponds to a simple element-wise multiplication in the Fourier Domain. This approach can be implemented using the \emph{Fast Fourier Transformation} (FFT) and is especially suited for large kernels and large batch sizes, where it can provide speedups of more than to 20\x compared to the matrix multiplication method~\cite{fbfft}.

\paragraph{Advantages of the Sliding-Filter 2D Convolution Approach}
Both the Matrix Multiplication and the FFT approach are well suited for general-purpose architectures such as GPUs.
They are especially efficient for large problem sizes and batched computation.
However, their additional memory consumption and the resulting need for tiling and re-stitching introduce artificial memory and computation requirements, which reduce the resource efficiency of the architecture.
Our focus on the regular, well-optimized ZynqNet CNN further eliminates the need to support all kinds of different parameter combinations.
Therefore we believe the direct 2D convolution approach as formulated in \cref{eq:convlayer-formula} to be the most efficient way to implement an FPGA-based accelerator, regarding both memory and computational requirements.
The nested summations clearly expose parallelism and translate well into nested loops in a high-level programming language, which makes the approach a good fit for High-Level Synthesis.

\subsubsection{Algorithm Description}
Based on the above considerations, a straightforward, nested-loop based formulation of 2D convolution was chosen as the foundation for this CNN accelerator. The loops are arranged in the order \emph{layer > height > width > input channels > output channels > kernel elements}.
For each layer, the outermost loops traverse all pixels left-to-right, top-to-bottom. At each pixel position, one input channel after the other is focused, and all corresponding output channels are calculated and accumulated.\footnote{Note that the algorithm actually calculates the 2D \emph{cross-correlation} between the filter and the input image, which corresponds to a 2D convolution with the filter mirrored at the origin. The \caffe implementation also uses this variation \cite{caffe-correlation}.}
\Cref{alg:cnn-algorithm} gives an algorithmic formulation for the complete ZynqNet CNN in pseudo-code, including \emph{stride} and \emph{concatenation} facilities.

\begin{algorithm}[tbp]
  \caption[Computation Algorithm for the ZynqNet CNN]{Nested-Loop based Computation of the All-Convolutional ZynqNet CNN.}
  \label{alg:cnn-algorithm}

  \begin{algorithmic}[1]

  \algnewcommand\algorithmicto{\textbf{to}}
  \algrenewtext{For}[3]{\algorithmicfor\ $#1\ \gets\ #2\ \algorithmicto\ #3$\ \algorithmicdo}
  \renewcommand{\algorithmicrequire}{\textbf{Input:}}
  \renewcommand{\algorithmicensure}{\textbf{Output:}}
  \renewcommand{\Statex}{HELLO}

  \algblockx[Block]{Begin}{End}[1]{\textbf{block} \emph{#1}\textbf{}}{\textbf{end block}}

  \Procedure {Zynqnet}{input image $\mathbf{I}$, trained weights $\mathbf{W}$, layer config}

  \State $in[0, \ldots] \gets \mathbf{I}$\;
  \For {L}{0}{layers-1}
    \Comment{loop over all layers}

    \Begin{} \Comment{per-layer setup: configuration and initialization}
      \State load layer config $\wout, \hout$ \Comment output width and height
      \State load layer config $\chin, \chout$ \Comment input and output channels
      \State load layer config $k, s$ \Comment kernel size ($k\times k$) and stride length
      \State load layer config $is\_1st\_split$ \Comment flag for \emph{expand1x1} and \emph{split1} layers
      \State load layer config $is\_2nd\_split$ \Comment flag for \emph{expand3x3} and \emph{split2} layers
      \If {not $is\_2nd\_split$}
        \State $out[L,\ldots] \gets 0$
        \Comment {initialize output feature maps $out[L]$}
      \EndIf
    \End

     \For {y}{0}{\hout-1}         \Comment{loop over y dimension}
      \For {x}{0}{\wout-1}        \Comment{loop over x dimension}
        \For {ci}{0}{\chin-1}     \Comment{loop over input channels}
          \For {co}{0}{\chout-1}  \Comment{loop over output channels}

            \Begin{} \Comment{dot-product at pos. $(y,x,ci)$ for output channel $(co)$}
              \State $dotprod \gets 0$
              \For {j}{-\lfloor\sfrac{k}{2}\rfloor}{\lfloor\sfrac{k}{2}\rfloor}
                \For {i}{-\lfloor\sfrac{k}{2}\rfloor}{\lfloor\sfrac{k}{2}\rfloor}
                  \State $image\_pixel = in[L, s \cdot y + j, s \cdot x + i, ci]$ \label{alg:line:imgcache}
                  \State $filter\_pixel = \mathbf{W}[L, ci, co, j, i]$ \label{alg:line:weightscache}
                  \State $dotprod = dotprod + image\_pixel \cdot filter\_pixel$\;
                \EndFor
              \EndFor
            \End

            \Begin{} \Comment{accumulate contributions from different input channels}
            \If {is\_2nd\_split}  \Comment{concatenate to existing output channels}
              \State $out[L,y,x,co + \chout] \gets out[L,y,x,co + \chout] + dotprod$ \label{alg:line:ocache1}
            \Else
              \State $out[L,y,x,{co}] \gets out[L,y,x,{co}] + dotprod$ \label{alg:line:ocache2}
            \EndIf
            \End

          \EndFor
        \EndFor  \Comment{one pixel done}

        \For {co}{0}{\chout-1} \Comment {apply bias and ReLU to pixel $(y,x,co)$}
          \State $out[L,y,x,co] \gets \mathrm{ReLU}(out[L,y,x,co] + W[L,bias,co])$
        \EndFor

      \EndFor
    \EndFor \Comment{one layer done}

    \If {$is\_1st\_split$} \Comment{second split layer will have same $in$ and $out$}
      \State $in[L+1,\ldots] = in[L,\ldots]$
      \State $out[L+1,\ldots] = out[L,\ldots]$
    \Else
      \State $in[L+1,\ldots] = out[L,\ldots]$
    \EndIf

  \EndFor \Comment{all layers done}

  \For {co}{0}{\chout-1}    \Comment{global average pooling}
    \State $out[layers,0,0,co] = \sum_{y,x}{in[layers,y,x,co]} \cdot \sfrac{1}{(\hout\cdot\wout)}$ \label{alg:line:gpoolcache}
  \EndFor

  \State $\mathbf{P} = \mathrm{softmax}(out[layers,\ldots])$ \Comment{final softmax classifier}
  \EndProcedure

  \end{algorithmic}
\end{algorithm}


\subsection{Parallelization}
\label{sec:algo-parallelization}

To reach a good throughput, the ZynqNet FPGA Accelerator needs to use all the computational resources available on the FPGA platform, especially the DSP slices.
The computation algorithm introduced in the last section therefore needs to be parallelized.

\paragraph{Parallelization Opportunities}
The nested loops can be a source of \emph{loop-level parallelism}: independent loop iterations can be partially or fully unrolled and executed in parallel on different processing elements.
The following sources of loop-level parallelism can be exploited in the ZynqNet CNN:

\begin{itemize}
  \item independence of \emph{layers} when applied to different image frames
  \item independence of dot-products at different \emph{pixel positions} $(y,x)$
  \item independence of \emph{input channels} $ci$
  \item independence of \emph{output channels} $co$
  \item independence of \emph{intra-kernel multiplications}
\end{itemize}

The inter-layer parallelism is interesting for batched, pipelined or dataflow implementations, but not for low-latency real-time inference where we would like to finish processing the current frame before starting with the next.
A second source of parallelism lies in the spatial domain. In principle, filter applications at different locations $(y,x)$ can be computed concurrently without interdependencies.\footnote{Note, however, that the memory access patterns for loading local receptive fields are unfavorable, unless these are stretched out beforehand, such as in the Matrix Multiplication method.}
When parallelizing over the input channels, multiple results are generated which stem from different input channels $ci$, but target the same output channel $co$. They must therefore be summed up, according to the summation over $ci$ in \cref{eq:convlayer-formula}.
This is very similar for parallelization over the output channels, where multiple filters $\mathbf{F}^{(ci,co)}$ can be applied to the same input channel $ci$ to generate results for different output channels $co$. The results for the individual output channels need to be accumulated as well, however with the difference that this accumulation concerns each output channel separately and can happen in a distributed way.
The final and most straight-forward opportunity for concurrency lies in the dot-product operation, where all multiplications can be executed in parallel. In the example of a 3\x3 kernel, a speedup factor of 9 can be reached.

\paragraph{Parallelization Strategy}
We exploit intra-kernel parallelism by fully unrolling all 3\x3 kernels into 9 parallel multiplications combined with an adder tree.
Additionally, we make use of the independence of the output channels and partially unroll the $co$ loop by a parametrizable factor of $\npe$.
Although many more opportunities exist, these transformations are enough to fully utilize the available computational capacity of the given FPGA platform.
Furthermore, no unnecessary multiplications are executed, which makes the chosen algorithm and parallelization strategy ideal with respect to the design goals.

\subsection{Data Reuse}
\label{sec:algo-data-reuse}

\paragraph{Need for on-chip Caching}
Looking at the pseudo-code in \cref{alg:cnn-algorithm}, it can be seen that multiple memory locations are read and written more than once.
Accesses into main memory are expensive, both in terms of latency and energy. They cannot be completely avoided because the on-chip memory is not big enough to hold all CNN parameters as well as the intermediate feature maps.
However, the goal is to minimize the number of reads and writes to the external memory by maximizing on-chip data reuse.
Furthermore, all unavoidable memory operations should be linear in order to facilitate \emph{burst mode transfers}.
\emph{Caches} allow both the linearization of memory accesses, as well as the temporary storage of values that will be reused shortly after. The arrays which can profit from caching in \cref{alg:cnn-algorithm} are:

\begin{itemize}
  \item the \emph{input feature maps} $in[L,y,x,ci]$ (\cref{alg:line:imgcache})
  \item the \emph{output feature maps} $out[L,y,x,co]$ (\cref{alg:line:ocache1,alg:line:ocache2})
  \item the \emph{weights memory} $\mathbf{W}[L,ci,co]$ (\cref{alg:line:weightscache})
\end{itemize}


\begin{figure}[tb]
  \centering
  \includegraphics[width = 0.7\linewidth]{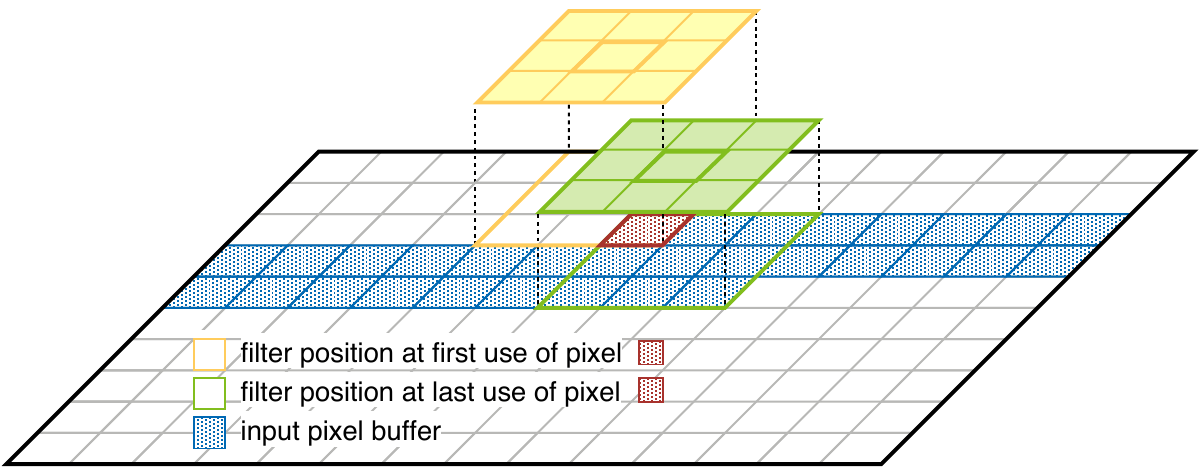}
  \caption[Illustration of the Input Line Buffer in the ZynqNet FPGA Accelerator]{Illustration of the Input Line Buffer, which needs to span slightly more than 2 full image lines of each input feature map to cover all reuse opportunities.}
  \label{fig:input-linebuffer}
\end{figure}

The ZynqNet CNN has a maximum filter size of 3\x3, therefore each input pixel $(y,x,ci)$ is part of up to 9 local receptive fields and is used in up to 9 dot-products. Additionally, there are $co$ output filters which are applied to the same input feature map which further increases the reuse factor by $co$.
In the formulation of \cref{alg:cnn-algorithm}, each input pixel $(y,x,ci)$ must be cached for a little more than 2 full image lines $y$ to cover all reuse opportunities, as illustrated in \cref{fig:input-linebuffer}.
The output feature maps are \emph{read-modify-write} accessed when the output channels are accumulated in \cref{alg:line:ocache1,alg:line:ocache2}. This is a particularly inefficient memory access pattern, because latency is involved in both the read and the write transfer. However, a buffer that holds all output channels of the current pixel is enough to keep these memory accesses on-chip.
A similar opportunity for caching exists during global average pooling, where the system needs to hold the accumulated mean values calculated in \cref{alg:line:gpoolcache}.
Finally, for each pixel $(y,x)$, all of the current layer's weights $\mathbf{W}[L,ci,co]$ are required. These $ci\times co \times (k\times k)$ parameters should be kept local and not be fetched from main memory for each single pixel.

\paragraph{Caching Strategy}
To optimize the memory accesses, we introduce four on-chip caches.
\begin{description}

\item[Image Cache (ICache)] is a line buffer which holds a few lines of each input feature map.
The largest input image has a width of 256 pixels, and the deepest input feature maps count 736 channels. However, the ZynqNet CNN trades image width against channel depth in each layer, with the result that an image line never contains more than \num{8192} pixels. A little more than 2 lines need to be cached, but for simplicity and speed, a capacity of 4 lines is used in the current accelerator.\footnote{The necessary division and modulo by 4 in the address calculation are essentially free in hardware.} The ICache therefore holds \num{32768} elements.

\item[Output Cache (OCache)] is a small cache used to buffer the output channels of a single pixel. Different input channels generate contributions to the same output channel, which are accumulated on the \num{512}-element OCache.
This buffer can be written back in a burst transfer when all input and output channels of a pixel have been calculated.

\item[Global Pooling Cache (GPoolCache)] is similar to the OCache, and holds the intermediate accumulation results during the global average pooling.

\item[Weights Cache (WCache)] is the final and biggest cache. It holds all the $ci \times co$ filters of the current layer. The accelerator benefits massively from the low parameter count in ZynqNet CNN, which allows all weights to be kept on-chip. The maximum number of $384\times112\times(3\times3) = \num{387072}$ weights plus 112 bias values is required in layer \emph{fire8/squeeze3x3}. Layer \emph{conv10} requires a comparable number of $736\times512\times(1\times1)=\num{376832}$ parameters plus 512 bias values. Due to implementation details, the cache is implemented with a capacity of $16\times3\times1024\times9 = \num{442368}$ elements.
\end{description}

These caches are sufficient to completely avoid any unnecessary accesses to the main memory. Each input pixel $in[L,y,x,ci]$ is loaded exactly once, each output pixel $out[L,y,x,co]$ is written exactly once and each weight $\mathbf{W}[L,co,ci]$ is fetched only once. In terms of main memory accesses, the chosen algorithm and caching strategy are therefore ideal.




\section{Hardware Architecture and Schedule}
\label{sec:block-diagram}

\paragraph{Introduction}
The nested-loop algorithm presented in the last \cref{sec:algorithm} provides the basis for the computation of the ZynqNet CNN in the FPGA-based accelerator. The parallelization strategy introduced in \cref{sec:algo-parallelization} defines which operations can be efficiently executed in parallel. And finally, the caching strategy from \cref{sec:algo-data-reuse} describes the necessary buffers to avoid unnecessary main memory accesses.
With all this groundwork laid, the architecture is basically ready to be implemented in software and compiled with High-Level Synthesis.
However, HLS expert Mike Fingeroff warns very early on in the introduction of his \emph{High Level Synthesis Blue Book}, that good quality of results "requires an understanding of the underlying hardware architecture of an algorithm, so that it is reflected in the \cpp design''~\cite{hls-bluebook}. We learned the essentiality of this statement the hard way, during multiple complete redesigns of the \cpp software description.
This section introduces both a hardware block diagram and a detailed schedule, which try to capture the previously described algorithm and the proposed optimizations. Although time-consuming and not strictly necessary, it is highly recommended to prepare such documents before starting with the HLS implementation to avoid even more expensive redesigns.

\begin{figure}[tbp]
  \centering
  \includegraphics[width = 0.95\linewidth]{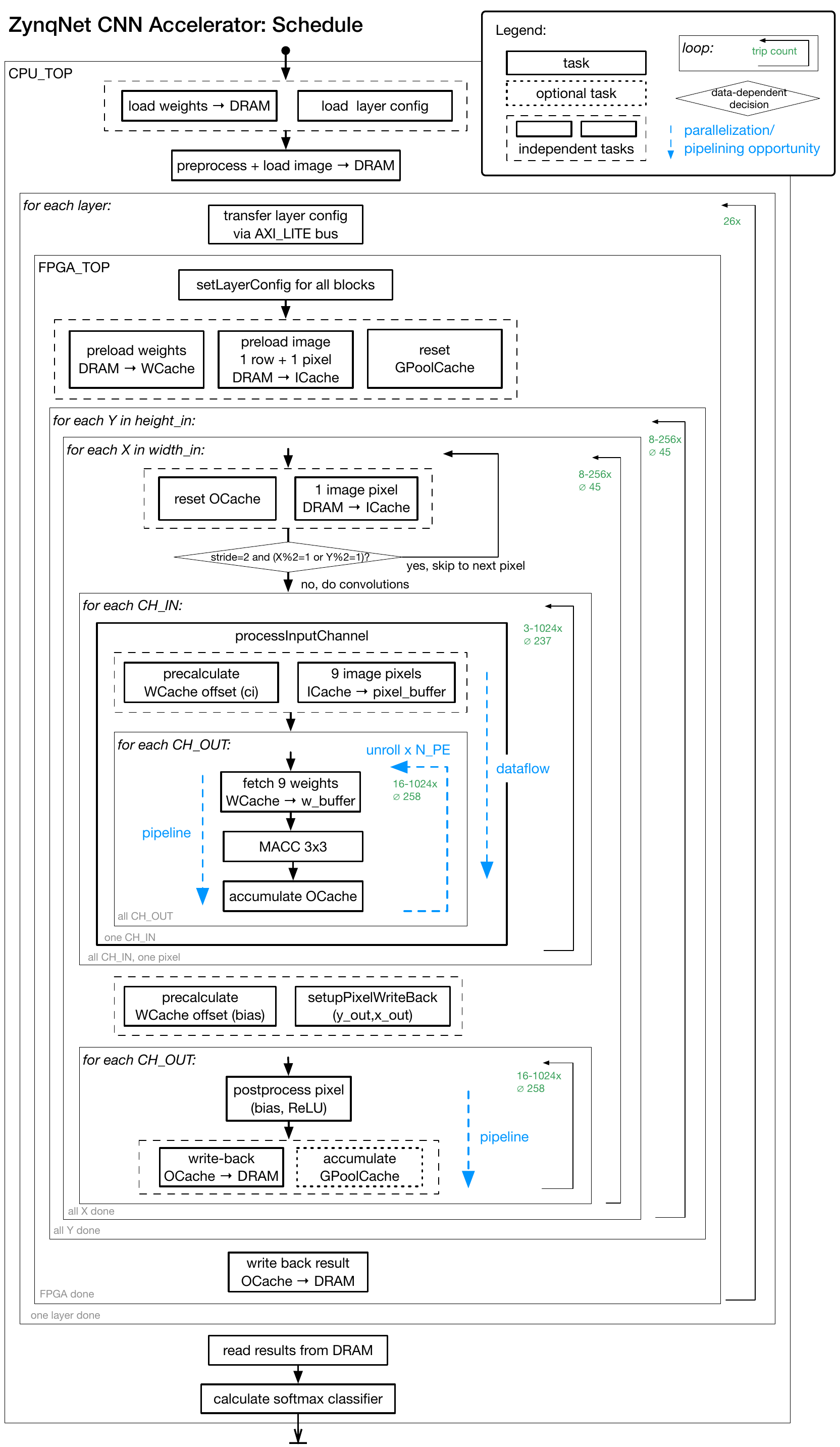}
  \caption[Algorithmic Schedule of the ZynqNet FPGA Accelerator]{Detailed Algorithmic Schedule for the ZynqNet FPGA Accelerator, highlighting the most important Parallelization and Pipelining Opportunities.}
  \label{fig:schedule}
\end{figure}

\paragraph{Algorithmic Schedule} \Cref{fig:schedule} captures the nested-loop algorithm defined in the previous sections, and turns it into a detailed schedule, which also highlights the most important opportunities for \emph{task-level parallelism} (blocks drawn in parallel and sections marked with ``dataflow''), \emph{loop-level parallelism} (loops marked with ``unroll'') and \emph{pipelining} (sections marked with ``pipelining'').

\paragraph{High-Level Block Diagram} Even though not strictly necessary, drawing a block diagram can help essentially to

\begin{itemize}
  \item optimize the HLS code for the given hardware platform
  \item get early estimates on resource utilization and memory requirements
  \item get a feeling for possible bottlenecks in the design, such as high-fanout nets, large muxes or possible routing congestions
  \item efficiently and appropriately structure the software representation
\end{itemize}

\begin{figure}[tbp]
  \centering
  \includegraphics[width = \linewidth]{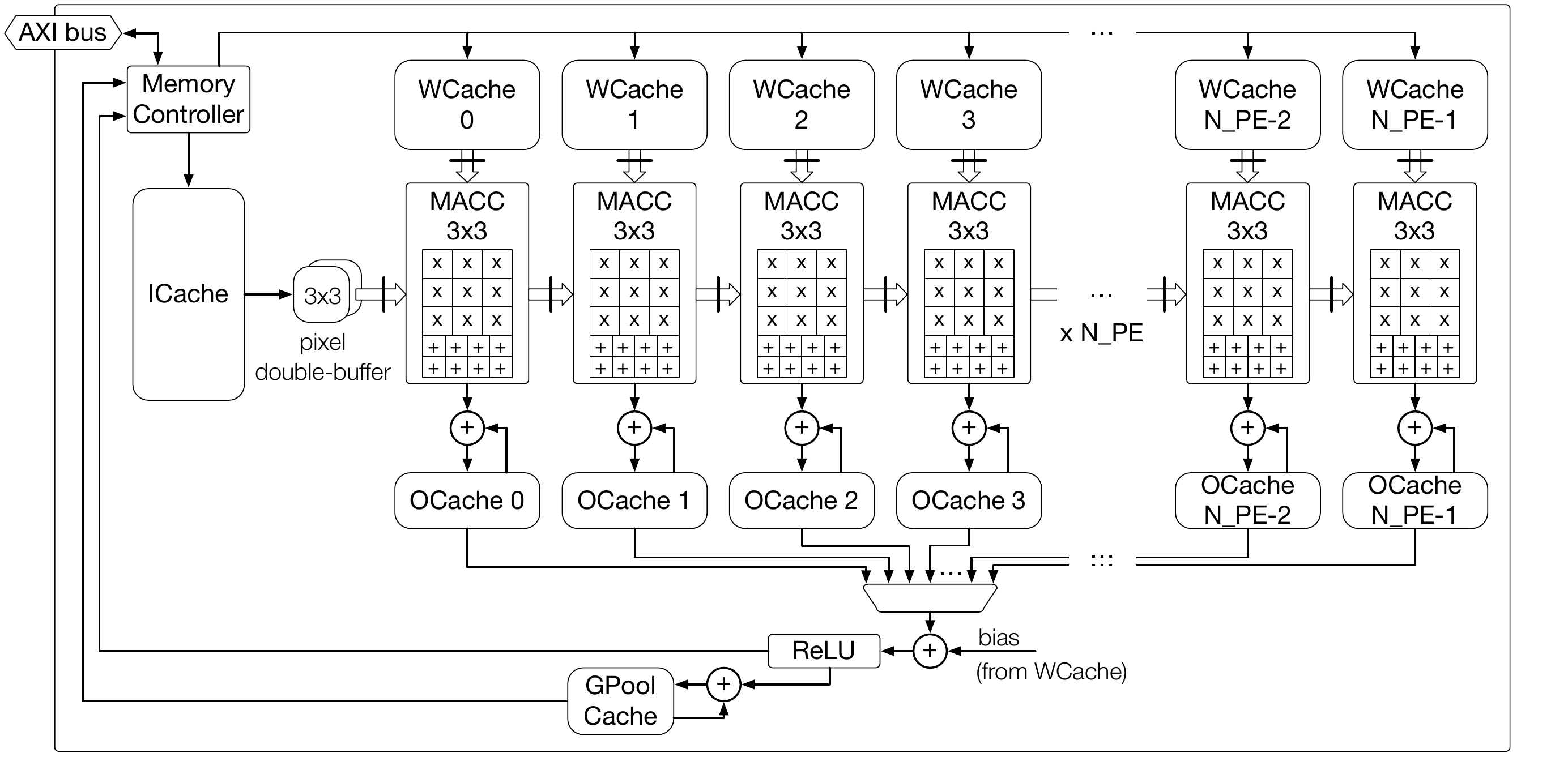}
  \caption[High-Level Block Diagram of the ZynqNet FPGA Accelerator]{High-Level Block Diagram of the FPGA-based Accelerator for the ZynqNet CNN.}
  \label{fig:highlevel-blockdiagram}
\end{figure}

\Cref{fig:highlevel-blockdiagram} gives an overview of the hardware organization in the final version of the FPGA-based accelerator. Another block diagram which includes the actual cache sizes and references to the \cpp software implementation can be found in \cref{fig:detail-blockdiagram} in the appendix. Note, however, that both diagrams were manually created, and are therefore not necessarily representative for the hardware generated by Vivado HLS.

\section{High-Level Synthesis Design Flow}
\label{sec:hls-design-flow}

ASICs and FPGA-based systems are typically planned and designed down to the finest details before implementation. Debugging those designs is very inconvenient, requires long simulation runs, and hidden bugs cannot easily be fixed later. Designers therefore start with a high-level block diagram, and then carefully refine the architecture down to the last register and state-machine before writing the first line of RTL code.
High-Level Synthesis promises an alternative approach. By abstracting away many implementation details (such as the design of finite state machines, insertion of pipeline registers, definition of handshake interfaces, setup of test-benches, etc.) and handling them in the HLS compiler instead, designers can start experimenting and optimizing immediately after they have a working software implementation in C, \cpp or SystemC.
High-Level Synthesis is supported for the Zynq FPGAs by \emph{Vivado HLS}, and Xilinx talks of ``4\x\ speed-up in development time'' for designs using their High-Level Productivity Design Methodology~\cite{xilinx-high-level-design-methodology}. Faster development time has been especially attractive considering the short time frame of this master thesis. Therefore, even though neither the author nor any of the co-workers at Supercomputing Systems AG had previous experience with \emph{Vivado HLS}, we decided to give this promising design methodology a try.

The following sections introduce some of the experiences gained through working with Vivado HLS 2016.2, ranging from different \cpp coding styles (\cref{sec:hls-codingstyle}) to different ways of constraining and shaping the synthesis (\cref{sec:hls-pragmas}). We also report some of the difficulties and limitations encountered in the current generation of VHLS in \cref{sec:hls-limitataions}.

\subsection{Introduction to Vivado HLS}
Vivado HLS, abbreviated VHLS and formerly known as AutoESL, is a High-Level Synthesis tool for Xilinx FPGAs. It is a standalone application based on the Eclipse development environment. VHLS can be used to write and debug C, \cpp and SystemC specifications of FPGA designs. Its most important component is the \emph{HLS compiler}, which analyzes and translates the high-level code into an intermediate low-level representation of all the necessary instructions to run the program.
It then optimizes and parallelizes these instructions into a \emph{schedule} and a \emph{resource allocation} scheme, and generates suitable RTL code in either Verilog or VHDL.

A very important capability of HLS software is the automated \emph{verification} of the generated RTL code, which enables designers to use the original high-level software specification as a \emph{test-bench}.
If the software model includes enough test cases, and the automated verification passes, the generated RTL code can be assumed to be correct. This feature works by means of \emph{co-simulation} in Vivado HLS: In a first step, the software model is executed and all input and output data consumed and generated by the function-to-be-synthesized are recorded. Then the RTL code is simulated with the recorded data as input stimuli, and the output from both the software model and the RTL simulation are compared. All data values consumed and emitted must match in order for the co-simulation to be successful.

A unique characteristic of high-level synthesis with C and \cpp is the complete absence of the concepts of timing and clock cycles in the software specification (which can be both a curse and a blessing, as further explained in \cref{sec:hls-limitataions} on the limitations of the HLS approach).
The HLS design is constrained, shaped and optimized using a number of \emph{compiler directives} (either as in-code \emph{pragmas} or using a separate TCL-based script). The directives can be used to specify the implementation and partitioning of memories, the unrolling of loops, function-level pipelining, etc. More details on compiler directives follow in \cref{sec:hls-pragmas}.

Vivado HLS also has an astonishingly good support for object-oriented \cpp. There is full support for \cpp classes, private and public member variables, and even (compile-time resolvable) inheritance.\footnote{Note, however, that the top-level function has to be a plain global function.}
Pointers and even double-pointers are also supported, albeit with some limitations: Pointers can only be casted between native C types, arrays of pointers may only point to scalars or arrays of scalars, and all functions which use a double-pointer are inlined together~\cite{xilinx-ug902}.

After each synthesis run, Vivado HLS estimates the device utilization and the maximum achievable clock frequency of the design. The tool also provides a number of different analysis views that visualize the resources allocated for each code section as well as the exact schedules for each loop and function.

Handing the tedious process of writing register transfer level code off to the compiler can heavily speed up the development of FPGA designs. Xilinx talks about average 4\x~speed gains in the development of new components, and speed gains of up to 10\x~when adapting previous designs, while reaching between \SI{70}{\percent} to \SI{120}{\percent} of the quality of results with respect to hand-coded RTL~\cite{xilinx-high-level-design-methodology}. This speedup, combined with a more agile development style and increased flexibility are especially important with regard to the ever-growing design complexities and the increasing capacities of newer FPGA generations.

\subsection{Coding Style}
\label{sec:hls-codingstyle}

The \emph{Vivado Design Suite User Guide for High-Level Synthesis} (UG902, \cite{xilinx-ug902}) is the most important document when working with Vivado HLS. It states that HLS enables designers to ``work at a level that is abstract from the implementation details, which consume development time'' and ``create specific high-performance hardware implementations'' by ``controlling the C synthesis process through optimization directives'', which sounds almost too good to be true.

The \emph{High-Level Synthesis Blue Book} \cite{hls-bluebook} by Mike Fingeroff, HLS expert at Mentor Graphics, is another major resource for guidelines regarding the design with High-Level Synthesis.
While also highlighting the benefits of the higher level of abstraction and the dramatic improvements made in the last years, the author also notes that ``there is still the potential for ending up with poor quality RTL when the \cpp is not well written'' and that ``good style not only requires an understanding of the underlying hardware architecture of an algorithm, so that it is reflected in the \cpp design, but also an understanding of how HLS works.'' In his book, he advocates an astonishingly ``low-level'' design style, which tries to directly mimick individual registers, muxes and arithmetic operations down to bit level in the C code -- something which is arguably not very ``abstract from the implementation details'' of the algorithm.

Of course, HLS compilers get better at understanding different algorithms and coding styles with every new version, and their coming can be compared to the rise of logic synthesizers which required very specific description styles initially, and are capable of generating relatively efficient logic from almost any specification today. One key aspect is to accept a design that may not be a \emph{perfect} solution, but which does the job \emph{well enough}~\cite{hls-bluebook}.
With RTL designs, the goal was often to optimize an architecture down to the last bit. This may no longer be feasible with HLS -- but maybe also no longer necessary, thanks to the abundance of resources available in modern FPGAs.

Initially, we found it very difficult to find a satisfactory coding style, given the short code examples and the different styles used in the two HLS guidelines. This section describes the experiments made and reports their successes and failures.\footnote{More code examples can also be found in the Vivado High-Level Synthesis Tutorials (UG871)~\cite{xilinx-ug871}.}


\subsubsection{Unstructured, Monolithic}

The first software implementation of the ZynqNet FPGA Accelerator algorithm was designed to be as simple as possible:
\begin{itemize}
  \item straightforward implementation of \cref{alg:cnn-algorithm}
  \item 7 nested loops (layers, height, width, input channels, output channels, kernel y and x)
  \item all arrays and most variables declared as global
\end{itemize}
This seemed like an adequate representation of the FPGA hardware (where all resources necessarily exist ``globally'') with minimum complexity of the control flow (no function calls, no handshakes, just a number of interconnected counters for the loop indices).
The \cpp software model eventually compiled and worked well. However, the HLS synthesis got stuck in the middle of the \emph{design analysis} phase without any error or indication of what was currently being analyzed. Many variants and changes to the code were tried without success, and overall this coding style proved to be:
\begin{itemize}
  \item hard to constrain using pragmas (scopes of application are not well defined)
  \item hard to read and maintain due to the unstructured ``spaghetti code'' nature
  \item very hard to debug due to the lack of indicative error messages
\end{itemize}


\subsubsection{Object-Oriented}
Having learned from the previous approach, the software model was rewritten from scratch. The new version included extensive testing, logging, debugging and verification facilities. Additionally, a conversion tool for \texttt{.prototxt} network description files and \texttt{.caffemodel} weight files as well as support for strided convolution and on-the-fly padding were added. This enables a very easy adaptation and reconfiguration of the architecture for different network topologies.
This time, the accelerator core was written in an object-oriented manner:
\begin{itemize}
  \item hardware blocks modeled as class instances (MemoryController, ImageCache, Weights\-Cache, OutputCache, ProcessingElement)
  \item arrays and variables encapsulated as private class members
  \item data movement via high-level member functions (\cppcode{data_t ICache::getPixel(y,x,ci)}, \cppcode{void PE::macc2D(data_t pixels[9], data_t weights[9], data_t& result)}\dots)
  \item control flow still via nested loops in top-level function (layer, height, width, input channels) and inside \cppcode{class ProcessingElement} (output channel, kernel y and x)
\end{itemize}
This coding style gave better results and was more pleasant to work with.
By splitting the code into separate functional units, problems during synthesis became easier to trace and isolate. The usage of compiler directives was simplified, because the lexical scopes to which the directives apply now coincide with functional blocks (e.g. pipelining can be explicitly applied to the postprocessing function and to the pixel writeback loop).

However, it was still not possible to complete synthesis. This time we experienced multiple fatal crashes of the HLS compiler process during the \emph{RTL generation} phase. Closer inspection suggested that the compiler automatically inlined multiple hierarchical levels of function calls in order to avoid double or even triple pointers, and tripped somewhere in that process.
 Double and triple pointers are very easily created in object-oriented code. For example, assume a class \cppcode{ProcessingElement}, which includes a reference to an instance of another class \cppcode{WeightsCache *ProcessingElement::WCache}, which \emph{itself} contains an array \cppcode{data_t WeightsCache::BRAM[]} (the variable \cppcode{BRAM} may be hidden behind an interface \cppcode{data_t WeightsCache::read(int addr)}, but due to various reasons, this type of functions tend to be inlined by Vivado HLS). \cppcode{BRAM} is then accessed as \cppcode{this->WCache->BRAM} from inside class \texttt{ProcessingElement} (double pointer), and is itself a pointer to elements of type \cppcode{data_t} (triple pointer). The HLS compiler tries to avoid these double and triple pointers, and may for example try to inline the whole instance \cppcode{WCache} into \cppcode{ProcessingElement}, but this quickly gets messy (imagine for example that a member function of another class \emph{also} accesses the \texttt{BRAM} array).
Therefore, the object-oriented coding style had mixed success:
\begin{itemize}
  \item much easier to read, modify and debug
  \item much easier to apply compiler directives
  \item still no successful synthesis (triple pointers due to class instances)
\end{itemize}


\subsubsection{Block-Structured}
The third approach was finally designed to be a compromise between the flat spaghetti code approach and the fully hierarchical object-oriented approach. This coding style uses \emph{namespaces} to structure the code conceptually, while avoiding the need for references and pointers.
With namespaces, modular and object-centric code can be written (such as \cppcode{data_t px = ImageCache::getPixel(y,x,ci)} or \cppcode{OutputCache::reset()}), but the actual hierarchy stays flat (when \cppcode{OutputCache} is simply a namespace and not an object, no references or pointers to it are needed to access \cppcode{data_t OutputCache::BRAM[]}).
The software model for the ZynqNet FPGA Accelerator has been partially rewritten to fit this \emph{namespace-based} or \emph{block-structured} coding style:
\begin{itemize}
  \item use namespaces to structure code into modules
  \item arrays and variables are encapsulated in namespace-scopes
  \item data movement is done via high-level namespace-scoped functions
  \item control flow still via nested loops in top-level function (layer, height, width, input channel) and inside \cppcode{namespace ProcessingElement} (output channel, kernel y and x)
\end{itemize}

This approach worked very well, except for one flaw: Global pointers are not supported for synthesis in Vivado HLS, and variables defined within namespaces are also considered global. Therefore, the declaration of a pointer into main memory via \cppcode{data_t* MemoryController::} \cppcode{SHARED_DRAM} is not synthesizable, and accesses into main memory can not be properly hidden behind an interface (such as \cppcode{data_t MemoryController::loadNextWeight()}).
Instead the pointer into main memory (which comes in as an argument to the top-level function) has to be dragged through all affected functions as an additional argument (such as
\cppcode{data_t MemoryController::loadNextWeight(data_t* SHARED_DRAM)}, and therefore also \cppcode{WeightsCache::loadFromDRAM(data_t *SHARED_DRAM)}). While this solution is not very elegant, it works and this last coding style finally resulted in a synthesizeable design. The \emph{namespace-based} coding style combines the advantages of both previous attempts, and we would describe our next HLS design in this coding style again:
\begin{itemize}
  \item straightforward, close to hardware description
  \item easy to read, modify and debug code
  \item easy to apply compiler directives
\end{itemize}

\newcommand{\pragma}{\textcolor[rgb]{0.00,0.44,0.13}{\textit{\texttt{\#pragma}}}\xspace}
\newcommand{\pragmaHLS}[1]{\textcolor[rgb]{0.00,0.44,0.13}{\textit{\texttt{\#pragma HLS\xspace#1}}}}


\subsection{Compiler Directives}
\label{sec:hls-pragmas}
The high-level languages C and \cpp by themselves do not allow the designer to specify con\-cur\-rency in the code. Frameworks which enable the explicit parallelization of C and \cpp programs typically use either the concept of \emph{kernels} or \emph{threads} which are launched in parallel (e.g. CUDA, OpenCL or Pthreads), or they allow designers to annotate the source code with \emph{compiler directives} that specify the desired type of parallelization (e.g. OpenMP).

As already indicated earlier, Vivado HLS uses this second approach and supports the annotation of the high-level source code using \pragma directives.\footnote{Alternatively, TCL-based scripts can be used, which allows a separation of the optimization directives and the code. The scripts support the same compiler directives, but have not been used in this project.}
The compiler directives affect all code in the lexical scope in which they have been placed (such as a function, loop, or the branch of an if-clause), and can influence e.g. the synthesis of FPGA memories from arrays, the derivation of control and data flows, and the parallelization and pipelining of individual code sections.
However, in comparison to directly writing RTL code where the structure and timing of the design can be exactly controlled, shaping an architecture using compiler directives can feel more like trying to thread a needle while wearing fireproof gloves.

In this section we introduce the most important \pragmaHLS{} compiler directives which have been used for the ZynqNet FPGA Accelerator.

\subsubsection{Interfaces}
Vivado HLS usually synthesizes C/\cpp functions into different functional entities. All blocks automatically receive clock and reset ports (\texttt{ap\_clk}, \texttt{ap\_rst}). The function arguments are turned into RTL ports of different types.
The compiler directive \pragmaHLS{INTERFACE <mode> [register] [depth=<D>] port=<P>} allows the specification of the \emph{function-level interface protocol} and the \emph{port-level interface protocol} for each argument.

\paragraph{Function-Level Interface Protocols}
The function-level interface protocol is set by applying the \pragma to the \texttt{return} port. The choices are \texttt{ap\_none}, \texttt{ap\_ctrl\_hs} and \texttt{ap\_ctrl\_chain}, where the \emph{handshake} protocol \texttt{ap\_ctrl\_hs} is the default and creates \texttt{ap\_start}, \texttt{ap\_done}, \texttt{ap\_ready} and \texttt{ap\_idle} signals which let the blocks negotiate data transfers.

\paragraph{Port-Level Interface Protocols}
When the \pragma is applied to individual arguments to set the port-level interface protocol, there are many modes available depending on the type of argument, and both inputs and outputs can be automatically registered. \texttt{ap\_none} is the default mode for scalar pass-by-value and pass-by-reference inputs and corresponds to a simple wire. The \texttt{ap\_vld}, \texttt{ap\_ack}, \texttt{ap\_ovld} and \texttt{ap\_hs} modes add increasingly complex handshake protocols to the individual signals, with the \emph{output-valid} protocol \texttt{ap\_ovld} being standard for pass-by-reference outputs. Arrays on the function interface are normally synthesized into \texttt{ap\_memory} ports, which creates \emph{data, address} and \emph{RAM control} ports. Alternatively, the port can be turned into an \texttt{ap\_fifo} interface if the access patterns correspond to a first-in-first-out buffer behavior.

\paragraph{AXI4 Interfaces}
On the top-level, Vivado HLS also supports the \texttt{axis} (AXI4-Stream), \texttt{m\_axi} (AXI4-Master) and \texttt{s\_axilite} (AXI4-Lite) interfaces which strongly simplify the connection of the design into a larger system architecture. This project uses the AXI4-Master interface to connect to the main memory via the AXI4 bus in the Zynq XC-7Z045. An AXI4-Lite interface is used for configuring, starting and stopping the accelerator. Vivado automatically generates C/\cpp driver files for accessing the AXI4-Lite ports from software running either on the Zynq's ARM cores or on Soft Processor Cores in the FPGA fabric.

\paragraph{AXI4 Depth Settings}
When proceeding to the co-simulation, it is crucial to set the \texttt{depth} of the AXI4-Master ports correctly (i.e. to the exact number of elements in the array connected to this port on the test-bench side). Setting the depth too small results in the simulation getting stuck. Setting the depth too large results in ambiguous error messages or even segmentation faults in Vivado HLS 2016.2. The depth can also be passed to the \pragma using a \cppcode{const int} variable in \cpp.


\subsubsection{Data and Control Flow}
\label{sec:data-control-flow}

There are a number of \pragma directives that affect the control flow in hardware, and are therefore very important for parallelizing algorithms.

\paragraph{Loop Unrolling}
\pragmaHLS{UNROLL [factor=<N>]} instructs the compiler to unroll the loop in which the \pragma is placed, either completely or partially by a factor of \texttt{N}. Because Vivado HLS by default schedules all operations as soon as they are ready for execution, these unrolled iterations are then executed in parallel. Of course, unrolling only works if there are no dependencies between the loop iterations, and complete unrolling requires known loop bounds at compile time. Besides the opportunity for parallel execution of the loop body, unrolling also removes the loop entry and exit overhead, which otherwise adds two clock cycles to every iteration.

\paragraph{Dependencies}
\pragmaHLS{DEPENDENCE variable=<var> <intra|inter> [false]} allows to override the automatic (and relatively conservative) dependency analysis. This directive needs to be applied when loops cannot be unrolled because a (false) inter-iteration dependency is detected by the compiler. For example, a loop which executes a read-modify-write operation on every individual element of an array cannot be unrolled by default, because the compiler sees read-after-write operations on the same array variable. However, the designer knows that the operations target different elements in the array in every loop cycle, and can therefore assert a \emph{false dependency} to re-enable loop unrolling.

\paragraph{Loop and Function Pipelining}
\pragmaHLS{PIPELINE [II=<N>]} is a very important optimization directive for loops as well as for functions. This \pragma enables pipelining for the context in which it is placed, and for all entities in the hierarchy below. Vivado tries to build a pipelined design with an initiation interval \texttt{II=<N>} (default: \texttt{N=1}), which means that a new data element is accepted into the pipeline every \texttt{N} clock cycles. The necessary depth of the pipeline (and the corresponding latency) are automatically determined by the compiler. An important caveat is the fact that pipelining \emph{forces all loops in the hierarchy below to be fully unrolled}. Full unrolling requires fixed loop bounds, and therefore this requirement can often prevent the pipelining of higher-level loops and functions, even if the lower-level loops are themselves pipelined and would be fully compatible with e.g. \texttt{II=1}.

\paragraph{Resource Specification and Pipelining of Arithmetic Operations}
\pragmaHLS{RESOURCE variable=<var> core=<string> [latency=<N>]} specifies the resource (core) that should be used to implement variable \texttt{var} in the RTL. This can be useful to select a certain type of memory for an array (e.g. dual-ported block RAM \texttt{RAM\_2P\_BRAM} or single-ported distributed ROM \texttt{ROM\_1P\_LUTRAM}), but it is also very useful to pipeline arithmetic operations:
\begin{minted}{cpp}
  int sum = a + b;
  int product = a * b;
  #pragma HLS RESOURCE variable=sum core=AddSubnS latency=2
  #pragma HLS RESOURCE variable=product core=MulnS latency=4
\end{minted}
This code instructs Vivado HLS to use a pipelined \texttt{AddSub} block with 2 register stages for the addition, and a multiplier with 4 pipeline stages for the multiplication. Pipelining arithmetic operations like this can be very useful to resolve problems with slow paths, and complements \pragmaHLS{PIPELINE}.

\paragraph{Function Inlining}
\pragmaHLS{INLINE} forces a function to be inlined into all its callers, which effectively creates copies and additional hardware, and thereby avoids the overhead of the function-level handshake (which is typically around 2 to 3 clock cycles). Vivado HLS often inlines functions automatically, e.g. to increase throughput. This can be prohibited by specifying \pragmaHLS{INLINE off}.

\paragraph{Function Instantiation}
\pragmaHLS{FUNCTION\_INSTANTIATE variable=<arg>} also creates multiple copies of the function in which it is placed, one for each value that the function argument \texttt{<arg>} takes on during execution. In contrast to \emph{inlining}, this \pragma keeps the function hierarchy intact. It allows the specialization of each function instance for a fixed value of \texttt{arg}. This is an important \pragma in combination with the parallelization of array accesses. Consider the following code example:
\begin{minted}{cpp}
int readArray(int block, int idx) {
  return array[block][idx];
}
for (int i = 0; i < N; i++) {
  readArray(i%4, i/4);        // -> sequential array access
}                             //    [0][0],[1][0],[2][0],[3][0],[0][1],[1][1],...
\end{minted}
Assuming that \texttt{array} has enough read ports, the loop could be partially unrolled to allow parallel read accesses. However, this is prevented because the function \texttt{readArray} can only be called sequentially. Adding the function instantiation directive creates four copies of \texttt{readArray}, and unrolling by a factor of 4 becomes possible:
\begin{minted}[samepage]{cpp}
int readArray(int block, int idx) { // instances: readArray_{0,1,2,3}
  #pragma HLS FUNCTION_INSTANTIATE variable=block
  return array[block][idx];
}
for (int i = 0; i < N; i++) {
  #pragma HLS UNROLL factor=4
  readArray(i%4, i/4);        // -> parallel array access
}                             //    [0,1,2,3][0],[0,1,2,3][1], ...
\end{minted}

\paragraph{Dataflow}
\pragmaHLS{DATAFLOW} activates the dataflow optimization used for task-level parallelism in Vivado HLS. By default, the compiler always tries to minimize latency and improve concurrency by scheduling operations as soon as possible. Data dependencies limit this type of parallelism: By default, a process A must finish all write accesses to an array before it is considered finished and a second process B can start consuming the data.

By adding the \emph{dataflow} directive, Vivado HLS analyzes which data elements are produced and consumed in the individual processes within the directive's scope, and tries to create channels (double-buffer/pingpong RAMs or FIFOs) between producer and consumer loops or functions. These allow data elements to be exchanged as soon as they are ready. However, there are multiple restrictions: Only single-producer single-consumer schemes are allowed, blocks cannot be bypassed or conditionally executed, and feedback between tasks is not supported. Further, \emph{dataflows} cannot be created within loops with variable bounds or with multiple exit conditions.

The dataflow \pragma is used in this project to allow simultaneous prefetching of a new image patch, while the filters are applied to the previously fetched image patch. Example code for this scenario can be seen in \cref{lst:dataflow-example}.
All dependencies between the two tasks should be made explicit via function arguments (i.e. exchanging data between the two blocks via class member variables does not work reliably).

\begin{listing}[tbp]
  \caption[Dataflow Code Example]{Example Code using the \emph{Dataflow} Compiler Directive for Task-Level Parallelism.}
  \label{lst:dataflow-example}
  \centering
    \begin{minipage}{0.9\textwidth}
      \begin{minted}[autogobble,fontsize=\small,frame=none]{cpp}
        void ProcessingElement::processInputChannel(const coordinate_t y,
                                                    const coordinate_t x,
                                                    const channel_t ci,
                                                    const channel_t num_ch_out) {
        #pragma HLS INLINE off

          // Dataflow Channels:
          weightaddr_t ci_offset;  // precalculated offset into WCache
          data_t px_buffer[9];     // double-buffer for preloaded pixels (reg. file)
        #pragma HLS ARRAY_PARTITION variable=pixel_buffer complete dim=0

        #pragma HLS DATAFLOW

          // Task 1: Preload Image Pixel Buffer (fetch pixels around (y,x,ci))
          //         and precalculate ci-dependent offset into Weights Cache
          preloadPixelsAndPrecalcCIoffset(y, x, ci, num_ch_out, ci_offset, px_buffer);

          // Task 2: MACC All Output Channels on Preloaded Pixels
          processAllCHout(num_ch_out, ci, ci_offset, px_buffer);
        }
      \end{minted}
    \end{minipage}
\end{listing}

\paragraph{Latency}
\pragmaHLS{LATENCY [min=<int>] [max=<int>]} specifies a minimum and/or maximum latency for a certain code segment (such as a function, loop iteration, etc.) Specifying a high minimum latency for uncritical code sections can reduce the resource consumption and increase sharing. Specifying a low maximum latency causes Vivado HLS to increase its scheduling effort to achieve the target. This directive can be especially useful to relax the latency constraints in short blocks and increase the scheduling effort in larger tasks of a dataflow pipeline.



\subsubsection{Array Synthesis}

\paragraph{Memory Type and Style}
Memories in the FPGA hardware are described as \texttt{array}s in the high-level source code. Only statically declared arrays with a fixed size are supported for synthesis. The mapping between the C/\cpp arrays and the underlying hardware can be influenced with a number of compiler directives. The memory type (RAM, ROM, FIFO) and implementation style (Block RAM, Distributed RAM, Shift Register) can be chosen by using the previously introduced \pragmaHLS{RESOURCE} directive.

\begin{figure}[tbp]
  \centering
  \includegraphics[width = 0.75\linewidth]{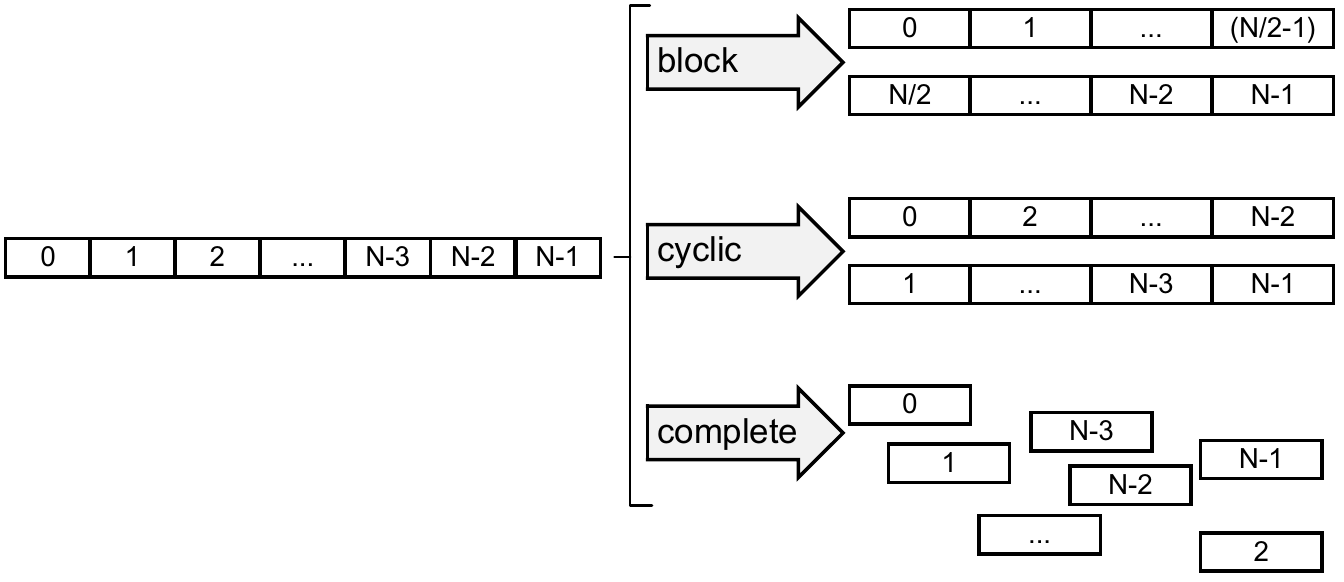}
  \caption[Example of Different Array Partitioning Modes in Vivado HLS]{Example of different Array Partitioning Modes in Vivado HLS (Illustration from \cite{xilinx-ug902}).}
  \label{fig:array-partition}
\end{figure}

\paragraph{Array Partitioning}
\pragmaHLS{ARRAY\_PARTITION variable=<var> <block, cyclic,\\ complete> [factor=<int>] [dim=<int>]} is the most important compiler directive with regard to memory synthesis.
By default, arrays are mapped linearly to Block or Distributed RAM (depending on their size) and receive either one or two access ports (two if it helps to reduce latency).
The \emph{array partitioning} directive allows an array \texttt{<var>} to be split into sub-arrays along different dimensions, which results in additional read and write ports.
These enable concurrent access to array elements of the different sub-arrays, and thereby increase the overall parallelizability of the design.
\pragmaHLS{ARRAY\_PARTITION} is especially important when \emph{load} or \emph{store conflicts} prevent the optimization of the design.
\Cref{fig:array-partition} illustrates the partitioning of a one-dimensional array by \texttt{factor=2} in the \emph{block} and \emph{cyclic} modes, as well as \emph{complete partitioning}. For multi-dimensional arrays, \texttt{dim} specifies the dimension to be partitioned. If an array is partitioned with \texttt{dim=0} and mode \texttt{complete}, it is fully disassembled in all dimensions and a register field is inferred.
Unfortunately, the \pragma cannot be applied repeatedly to the same dimension to create more complex array partitions. It is therefore recommended to already structure the arrays in C/\cpp code along multiple dimensions, and then split the required dimensions into separate memories using \emph{complete partitioning}. The following code example partitions the weights cache memory \texttt{WBRAM} in multiple dimensions:
\begin{minted}{cpp}
// Weights BRAM Config for ZynqNet CNN:
//          [ 16 ][       3        ][    1024  ][9] x 32bit
data_t WBRAM[N_PE][NUM_BRAMS_PER_PE][BLOCK_SIZE][9];

// Array Partitioning (dimensions indexed from 1)
#pragma HLS ARRAY_PARTITION variable=WBRAM complete dim=1 // PE ID
#pragma HLS ARRAY_PARTITION variable=WBRAM complete dim=2 // block ID
#pragma HLS ARRAY_PARTITION variable=WBRAM complete dim=4 // weight ID
#pragma HLS RESOURCE variable=WBRAM core=RAM_S2P_BRAM latency=3
\end{minted}

%

\subsection{Limitations and Problems}
\label{sec:hls-limitataions}
Considering the enormous complexity of the transformation of high-level sequential code into optimized FPGA hardware, Vivado HLS does an impressively good job. However, the road is still full of bumps -- designers should not expect a smooth ride, especially for larger and more complicated designs.
This section highlights a few of the most relevant problems and limitations that we encountered while working with High-Level Synthesis.

\begin{description}
  \item[Global Pointer and Multi-Pointer Support] As explained earlier, the fact that global pointers are not supported in Vivado HLS prevented the abstraction of the main memory interface. While the problem can be circumvented by including a pointer to the shared memory in all of the concerned function's arguments, the solution is cumbersome and inelegant. Very similar problems occur when the pointer to the shared memory is stored in a class, this time due to multi-pointers which result when accessing these classes.

  \item[Unsupported Pointer Casting to Custom Data Types] Because pointer casting is only allowed between native data types (\cppcode{char}, \cppcode{int}, \cppcode{float}, \dots) and not from and to custom data types (including \cppcode{typedef} aliases and \cppcode{struct}), loading such variables and especially structures over a memory bus is very complicated.
  Interpreting a structure \cppcode{struct { char c; float f; } S;} as an array of bytes or integers is straightforward in C and \cpp: \cppcode{char *B = (char*) &S; uint32_t *U = (uint32_t*) &S;}. In RTL code, it is just as simple to reconstruct the original data by bit-slicing and reassembling the incoming words. This combination would make it very easy to transfer arbitrary structures over a 32-bit AXI4-Master bus.
  Unfortunately, the necessary pointer reinterpretation (\cppcode{(uint32_t*) &S}) is unsupported in the current version of VHLS, and tedious, error-prone manual transformations between the custom types or structures and the bus data type are necessary, for example using \emph{unions}:%
  \footnote{This strategy has been used to transfer custom integer data types via a \texttt{float} AXI4-Master bus. Note, however, that this only works for custom types which are smaller or equal to the width of the bus data type.}
\begin{minted}{cpp}
  union { custom_t custom; uint32_t bus; } U;
  U.custom = ...; bus_transfer(U.bus);      // custom-to-bus
  U.bus = bus_receive(); ... = U.custom;    // bus-to-custom
\end{minted}

  \item[Imperfect Code Analysis] While the HLS compiler mostly does a good job at interpreting the high-level source code, it sometimes misses very important code optimizations. Consider the following example which describes the wrapping logic of a counter:
  \begin{center}
    \begin{minipage}{4cm}
      \begin{minted}{cpp}
        a++;
        if (a == MAX) a = 0;
      \end{minted}
    \end{minipage}
    \qquad\qquad\qquad
    \begin{minipage}{4cm}
      \begin{minted}{cpp}
        if (a==MAX-1) a = 0;
        else          a = a + 1;
      \end{minted}
    \end{minipage}
  \end{center}
  The left version is an intuitive description in \cpp. The right version is a more verbose description of the same logic. While both pieces of code have the exact same functionality,\footnote{Assuming that \texttt{a} is not declared \texttt{volatile}.} the left version takes two clock cycles to execute when synthesized, while the right version takes only one clock cycle. Manual low-level optimizations can therefore still be necessary in unexpected places.

  \item[Difficulties with Compiler Directives] Constraining the design with compiler directives alone can be difficult. The \pragma directives are not binding, and are sometimes interpreted by the HLS compiler rather as suggestions than imperatives. Furthermore, when multiple directives are combined, the result depends on the order of the \pragma commands as well as the order in which the directives are encountered by the compiler.

  \item[Pipelining Requires Loop Unrolling] As already mentioned above in \cref{sec:data-control-flow}, the compiler directive \pragmaHLS{PIPELINE} requires all loops in the function hierarchy below to be fully unrolled. The rationale is probably that VHLS can only build one giant pipeline with a single initiation interval in the current version. However, it would often be convenient to build pipelines while retaining the loop hierarchy: For example, two outer loops could be responsible for iterating over pixels, precalculating some values, and then feeding an additional \emph{inner loop} which does the heavyweight calculations. It is currently possible to pipeline this inner loop. However, the outer loops and therewith the whole outer loop entry and exit logic, the precalculations, etc. cannot be pipelined without fully unrolling the inner loop. In a more ideal scenario, it would be possible to create an outer pipeline which feeds the inner pipeline, where both have independent initiation intervals and both can be specified with a simple \pragma directive.

  \item[Pipeline Flushing Issue for Pipelines Nested in Dataflow]
  \label{sec:dataflow-and-pipeline-problem}
  Similar to the previous issue, Vivado HLS currently has a serious limitation when it comes to the combination of \texttt{dataflow} and \texttt{pipeline} compiler directives: An inner pipeline (\texttt{L\_INNER}) that is part of a \texttt{dataflow} scheme, which itself is placed inside an outer loop (\texttt{L\_OUTER}), is unnecessarily \emph{flushed} in every iteration of \texttt{L\_OUTER}.
  \Cref{lst:dataflow-pipeline-issue} illustrates this configuration using a minimal example.
  This issue is present in the current implementation of the
  elerator and heavily degrades its performance. Xilinx has acknowledged the problem and currently cannot offer a workaround or a solution. The only recommended workaround is to avoid High-Level Synthesis altogether and rewrite the architecture as RTL code~\cite{ralph-wittig-email}.

  \item[Slow Simulation Runs due to Unnecessary Warnings]
  The current version Vivado HLS 2016.2 seems to introduce a bug into the RTL code for floating-point multipliers and adders, which causes the \texttt{OPMODE} input port oft the DSP slices involved to contain undefined values. While the functionality of the simulation model is not impaired, the undefined values cause the simulator to issue hundreds of warnings per clock cycle. This slows the co-simulation so much that even the smallest designs take hours to simulate. The full ZynqNet FPGA Accelerator architecture, simulated with a small five-layer CNN, has been run for four days without reaching an end. Suppressing the warning message is not possible and thus we had to live without the assurance of a working co-simulation.


\end{description}

 \begin{listing}[tbp]
  \caption[Dataflow and Pipelining Example]{Example Code combining Dataflow and Pipelining Compiler Directives. Loop \texttt{L\_OUTER} cannot be pipelined, because loop \texttt{L\_INNER} in the hierarchy below cannot be unrolled due to its variable bounds. Even worse, the Pipeline in loop \texttt{L\_INNER} is unnecessarily flushed in every iteration of loop \texttt{L\_OUTER}.}
  \label{lst:dataflow-pipeline-issue}
  \centering
    \begin{minipage}{0.9\textwidth}
      \begin{minted}[autogobble,fontsize=\small,frame=none]{cpp}
        void task1_precalculate(int &channel) {
        #pragma HLS INLINE off                    // needed to enable dataflow
          channel = precalculate();
        }

        void task2_hardwork(int o, int &channel) {
        #pragma HLS INLINE off                    // needed to enable dataflow
          L_INNER: for (int i = 0; i < o; i++) {  // (variable loop bounds)
        #pragma HLS PIPELINE II=1
            work_with(channel, i, o);             // do calculations (inlined)
          }
        }

        void f_dataflow(int o) {
        #pragma HLS INLINE off                    // needed to enable dataflow
        #pragma HLS DATAFLOW
          int channel;
          task1_precalculate(channel);
          task2_hardwork(o, channel);
        }

        L_OUTER: for (int o = 0; o < o_max; o++) {
        #pragma HLS PIPELINE      /* DOES NOT WORK
                                       because L_INNER can't be unrolled */
          f_dataflow(o);          /* FLUSHES the innermost pipeline
        }                              on every call */
      \end{minted}
    \end{minipage}
\end{listing}


%
\section{Post-HLS Design Flow}

Despite the significant performance loss due to the Pipeline Flushing Issue explained above, we decided to finish the design and try to estimate its performance.
After a successful synthesis run, Vivado HLS creates a register transfer level description of the design in either VHDL or Verilog format. This RTL code can be exported in the so-called ``IP Catalog'' format, which is directly compatible with the Post-HLS \emph{Vivado Design Suite} design flow.

\subsection{Vivado Design Suite}

\begin{figure}[tbp]
  \centering
  \includegraphics[width = \linewidth]{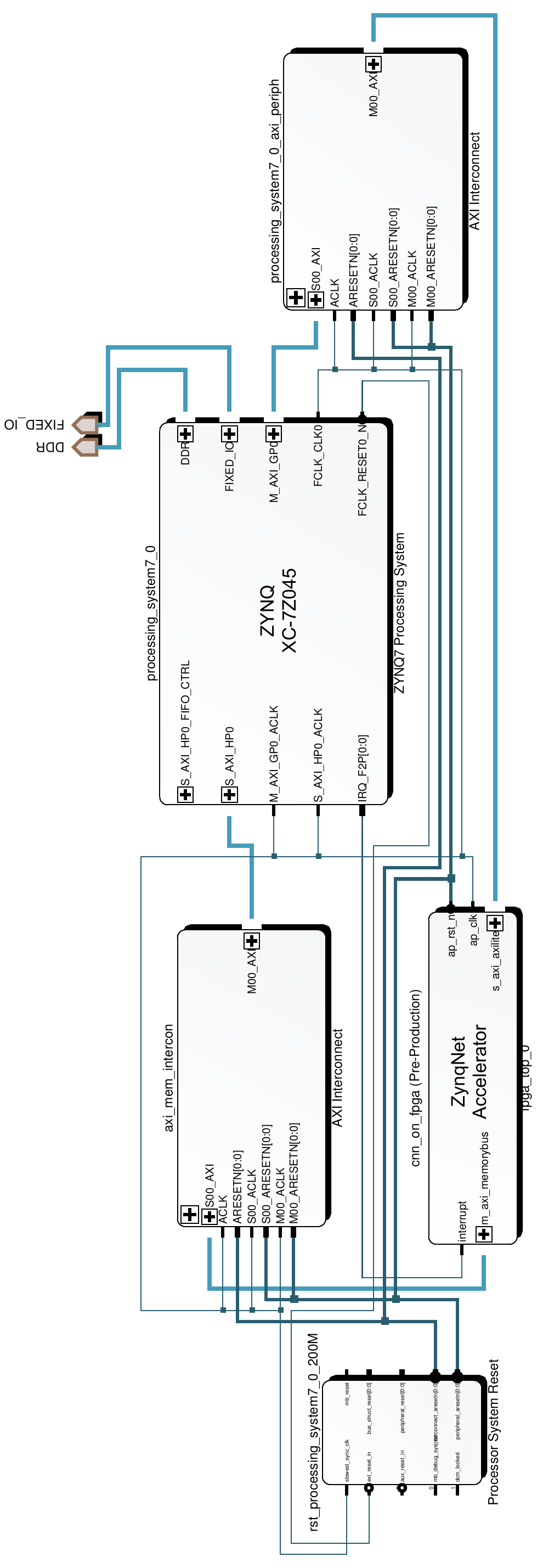}
  \caption[Vivado Design Suite Block Design of the ZynqNet FPGA Accelerator]{Block Diagram of the ZynqNet FPGA Accelerator and Zynq Processing System as generated by the Vivado Design Suite Block Design tool.}
  \label{fig:vivado-blockdesign}
\end{figure}

\paragraph{Importing the VHLS Design}
The RTL code generated by Vivado HLS still hast to be compiled into a bit-stream for the FPGA. Luckily, this has been made very easy with the Vivado Design Suite. All required steps are described and illustrated in the High-Level Synthesis Tutorial (UG871) \cite{xilinx-ug871}.
After importing the previously exported ``IP Catalog'' file in the Vivado Design Suite IP Catalog, a new \emph{Block Design} can be created, and the VHLS design can be added as a new component. Further, the ``Zynq7 Processing System'' block has to be added and configured, before the components can be automatically connected using the \emph{Run Block Automation} tool. \Cref{fig:vivado-blockdesign} shows a diagram of the fully connected ZynqNet FPGA Accelerator and Zynq XC-7Z045 blocks in the Vivado Design Suite Block Design tool.
The Vivado Design Suite then automatically generates the necessary VHDL or Verilog wrapper files and instantiates the VHLS design. At this point, the design is ready for synthesis.

\paragraph{Synthesis}
With the schedule and resource allocation already fixed and all timing constraints properly set by Vivado HLS, there is not much left to be configured in Vivado Design Suite itself. The synthesis and implementation can be influenced slightly by choosing from a number of different preset strategies, and should run through smoothly.
However, the timing results reported by the Vivado Design Suite can be quite different from the estimates reported by Vivado HLS.\footnote{In our (very limited) experience, small designs resulted in faster implementations than estimated by Vivado HLS, while larger designs sometimes ran into routing problems and resulted in significantly slower implementations.}
The Design Suite results respect the load and fanout of each signal, and include all actual wire delays. The ZynqNet FPGA Accelerator synthesized to a slightly slower design than estimated, due to highly congested areas where more than \SI{85}{\percent} of the routing capacity was utilized. The synthesis reports also highlight the longest paths, which gives vital hints for the optimization of the VHLS design. The routing delays can be significant, and therefore early synthesis runs, even with incomplete designs, are highly recommended.
When the RTL synthesis is finished, the \emph{bitstream} containing the binary FPGA configuration can be exported as a \texttt{.bit} file and loaded onto the Zynqbox.


\subsection{Zynq Driver Development}
With all these steps done, the FPGA side of the CNN accelerator is complete. The CPU-side software is also ready and tested as part of the test suite in Vivado HLS. However, there is still a missing key component: The low-level driver which lets the CPU software communicate with the FPGA block.

\paragraph{Xilinx Software Development Kit}
The Vivado Design Suite exports a \texttt{.hdf} \emph{Hardware Design File} which contains a description of the Zynq setup configured in the \emph{Block Design} step. Additionally, C-based driver files for the AXI4-Lite port are created. The hardware design file is then normally opened in the \emph{Xilinx Software Development Kit} (SDK) application. The Xilinx SDK supports the creation of both bare-metal applications which do not rely on an operating system, and Linux-based applications.
It includes all the tools needed to create a completely new, custom-tailored Linux environment including a custom \emph{First Stage Boot Loader} (FSBL) for the chosen Zynq configuration and a custom \emph{device tree} which has to be passed to the Linux kernel at boot time. However, the Zynqbox already runs a fully working and properly tested Linux installation and includes tools to load new bitstreams into the programmable logic. Due to lack of time, we strongly favored reusing this existing installation.

\paragraph{Memory-Mapped Input and Output}
The C-based driver files for the AXI4-Lite port, which can be exported from Vivado Design Suite, include functions for:
\begin{itemize}
  \item starting and stopping the top-level FPGA entity
  \item checking the status of the accelerator (idle, ready, done)
  \item setting and getting every top-level function argument bundled into the AXI4-Lite interface
\end{itemize}
The files also contain all the relative address offsets of the corresponding memory-mapped registers. However, the driver relies on the \emph{Userspace I/O} (UIO) kernel module, which in turn relies on the correct {device tree} being loaded into the kernel at boot time. Neiter of these requirements is fulfilled in the default SCS Zynqbox installation, and the advanced project time did not allow to fix this.
Therefore, we had to patch the low-level driver functions to directly access the Zynq's memory bus to talk to the FPGA-based block, instead of using the elegant UIO module.

In Linux, the root user can directly access the physical memory bus without going through the virtual-to-physical address translation by reading and writing the \texttt{/dev/mem} character file.
The physical address range which is assigned to the accelerator's AXI4-Lite interface can be found in the Address Editor in Vivado Design Suite's Block Design tool. The corresponding section of the \texttt{/dev/mem} file can then be \emph{memory-mapped} into the application's own memory space using \kern-1ex\mintinline[breaklines]{cpp}{int fd = open("/dev/mem", O_RDWR); volatile uint32_t* axilite = (uint32_t*)mmap(NULL, AXILITE_LENGTH, PROT_READ|PROT_WRITE, MAP_SHARED, fd,}\\\cppcode{AXILITE_BASEADDR);} All subsequent reads and writes of \cppcode{*(axilite + byte_offset)} are mapped into the \texttt{/dev/mem} file, and from there directly onto the memory bus.
This method has been successfully implemented, and the communication between the FPGA accelerator and the CPU-based software is fully functional. The only drawback is the requirement for root privileges when running the ZynqNet Embedded CNN.

\paragraph{Remarks on the current ZynqNet Driver}
\begin{itemize}
  \item The current First Stage Boot Loader (FSBL) in the Zynqbox configures the \texttt{FCLK\_CLK0} clock source for the programmable logic to \SI{100}{\MHz}. This setting cannot easily be changed, and therefore the ZynqNet FPGA Accelerator is currently only running at one half of the full \SI{200}{\MHz} clock speed which it was synthesized for.
  \item Before launching the driver, the \emph{High Performance AXI Slave} port \texttt{S\_AXI HP0} needs to be configured for 32\,bit bus width. This can be done by calling \texttt{axi\_hp\_config 0 32} on the Zynqbox.
  \item The bitstream for the ZynqNet FPGA Accelerator can then be loaded by calling \texttt{loadbit zynqnet\_200MHz.bit} in the firmware directory.
\end{itemize}


%
\chapter{Evaluation and Results}
\label{chap:results}

\cleanchapterquote{You can't always get what you want.\\ But if you try, sometimes\\ You just might find\\ You get what you need.}{The Rolling Stones}

The last two chapters have given a detailed introduction the both the \emph{ZynqNet CNN} and the \emph{ZynqNet FPGA Accelerator}.
Both of these components have been completed successfully, and together constitute the fully operable \emph{ZynqNet Embedded CNN}.
This chapter is concerned with an in-depth evaluation of this system regarding different aspects.
First, we assess the performance of the \emph{ZynqNet CNN} and compare it to prior work (\cref{sec:cnn-performance}).
Next, the performance of the \emph{ZynqNet FPGA Accelerator} is estimated, both in its current version and with a number of potential improvements applied (\cref{sec:fpga-performance}).
The final section brings both components together and investigates the overall system performance of the \emph{ZynqNet Embedded CNN} (\cref{sec:system-performance}).


%

\section{ZynqNet CNN Performance}
\label{sec:cnn-performance}

\begin{figure}[tbp]
  \centering
  \includegraphics[width=0.9\linewidth]{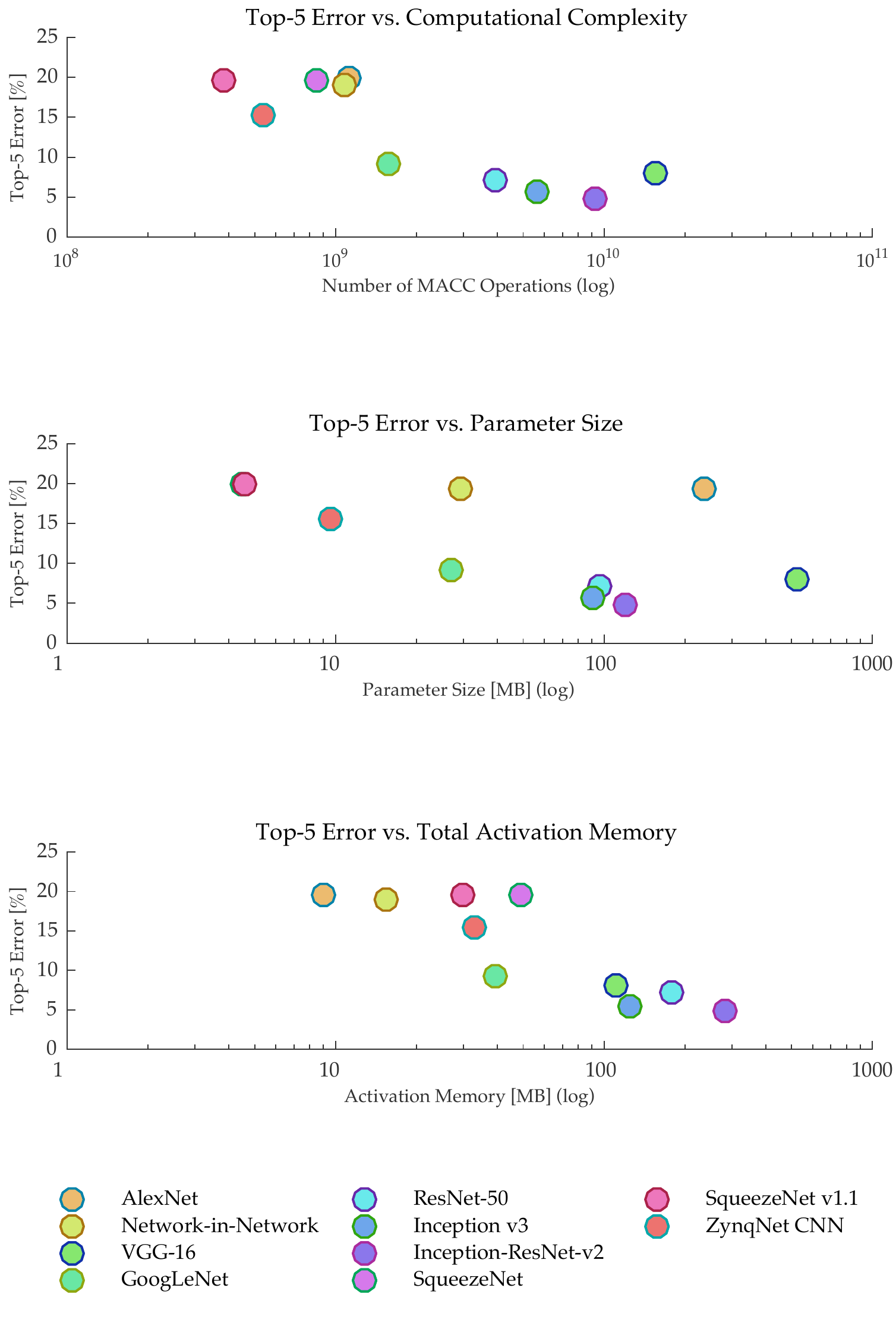}\vspace{2mm}
  \caption[Design Space Exploration of CNN Topologies including ZynqNet CNN]{Comparison of the ZynqNet CNN to CNN Architectures from Prior Work. Note the Logarithmic Scale on the x-Axes.}
  \label{fig:cnn-topologies-comparison}
\end{figure}

In \cref{sec:cnn-optimization} on the optimization of ZynqNet CNN, many aspects of the convolutional neural network's performance have already been discussed. Therefore, we confine ourselves to a summary of the most important characteristics in this section.
To start with, \cref{tab:topology-summary-zynqnet} repeats the comparison of the different CNN topologies, and this time includes the ZynqNet CNN and its key parameters. \Cref{fig:cnn-topologies-comparison} shows the updated design space exploration charts including the ZynqNet CNN.\footnote{The table as well as the design space exploration charts also report the parameters for SqueezeNet~v1.1, which has been published during the development of ZynqNet CNN.}

\clearpage
\begin{table}[tp]
\centering
\caption[Comparison of CNN Topologies for ImageNet Classification including ZynqNet]{Comparison of ZynqNet CNN to CNN Architectures from Prior Work.\protect\footnotemark{}}
\label{tab:topology-summary-zynqnet}
\begin{tabular}{@{}lrrrrr@{}}
\toprule
      & \begin{tabular}[c]{@{}r@{}}\#conv.\\ layers\end{tabular}
      & \begin{tabular}[c]{@{}r@{}}\#MACCs\\{[}millions{]}\end{tabular}
      & \begin{tabular}[c]{@{}r@{}}\#params\\{[}millions{]}\end{tabular}
      & \begin{tabular}[c]{@{}r@{}}\#activations\\{[}millions{]}\end{tabular}
      & \begin{tabular}[c]{@{}r@{}}ImageNet\\top-5 error\end{tabular} \\
\midrule 
\bfseries ZynqNet CNN& \bfseries\num{18}   & \bfseries\num[group-minimum-digits = 4]{530}   & \bfseries\num{2.5}   & \bfseries\num{8.8}   & \bfseries\num{15.4}{\%} \\  
AlexNet             & \num{5}    & \num[group-minimum-digits = 4]{1140}  & \num{62.4}  & \num{2.4}   & \num{19.7}{\%} \\  
Network-in-Network  & \num{12}   & \num[group-minimum-digits = 4]{1100}  & \num{7.6}   & \num{4.0}   & \textasciitilde\num{19.0}{\%} \\  
VGG-16              & \num{16}   & \num[group-minimum-digits = 4]{15470} & \num{138.3} & \num{29.0}  & \num{8.1}{\%}  \\  
GoogLeNet           & \num{22}   & \num[group-minimum-digits = 4]{1600}  & \num{7.0}   & \num{10.4}  & \num{9.2}{\%}  \\  
ResNet-50           & \num{50}   & \num[group-minimum-digits = 4]{3870}  & \num{25.6}  & \num{46.9}  & \num{7.0}{\%}  \\  
Inception v3        & \num{48}   & \num[group-minimum-digits = 4]{5710}  & \num{23.8}  & \num{32.6}  & \num{5.6}{\%}  \\  
Inception-ResNet-v2 & \num{96}   & \num[group-minimum-digits = 4]{9210}  & \num{31.6}  & \num{74.5}  & \num{4.9}{\%}  \\  
SqueezeNet          & \num{18}   & \num[group-minimum-digits = 4]{860}   & \num{1.2}   & \num{12.7}  & \num{19.7}{\%} \\  
SqueezeNet v1.1     & \num{18}   & \num[group-minimum-digits = 4]{390}   & \num{1.2}   & \num{7.8}   & \num{19.7}{\%} \\  
\bottomrule
\end{tabular}
\end{table}

\footnotetext{The ImageNet top-5 error rate is reported for single-net single-crop evaluation. \#MACCs refers to the number of multiply-accumulate operations in one forward pass. \#activations is the total pixel count in all output feature maps of all layers.}

\subsection{Accuracy}
The DSE charts in \cref{fig:cnn-topologies-comparison} as well as the parameter summary in \cref{tab:topology-summary-zynqnet} both show the increased accuracy of ZynqNet CNN with respect to both SqueezeNet and AlexNet. The top-5 error rate has been improved by more than \SI{20}{\percent} relative to the starting point, or by a total of 4.3 percentage points.
ZynqNet labels \SI{84.6}{\percent} of all ImageNet examples correctly in the top-5 validation test.
The top-1 accuracy has been improved from \SI{55.9}{\%} to \SI{63.0}{\%}.
These results place ZynqNet approximately in the midfield of the tested CNN topologies. Note, however, that most of the other topologies have been optimized for maximum possible accuracy at the cost of heavily increased computational and memory requirements.

\subsection{Computational Complexity}
The computational complexity of ZynqNet has been lowered by almost \SI{40}{\percent} in comparison to the original SqueezeNet and by more than \SI{50}{\percent} compared to AlexNet. The network requires only 530 million multiplications and accumulations for one forward pass, making it one of the least expensive CNNs for image classification on ImageNet.
The lightness of ZynqNet CNN is complemented by its highly regular architecture. The CNN consists only of convolutional layers, ReLU nonlinearities and one global average pooling layer.

\subsection{Memory Requirements}
In terms of its memory requirements, ZynqNet CNN differs from its ancestor SqueezeNet. SqueezeNet is mostly concerned with the minimization of the number of weight parameters. On the other hand, ZynqNet CNN tries to strike a balance between the number of parameters, the computational complexity, the size of each intermediate feature map and the overall accuracy of the convolutional neural network. Therefore, ZynqNet CNN uses 2.5 million weight parameters, which is twice as many as SqueezeNet, yet still roughly one order of magnitude less than most other CNNs for image classification on ImageNet. The total number of activations as well as the size of the largest output feature maps are approximately equal in the two networks.

\subsection{Resource Efficiency}
\paragraph{MACC Operations}
From the \emph{Top-5 Error vs. Computational Complexity} graph in \cref{fig:cnn-topologies-comparison}, it can be seen that ZynqNet CNN belongs to the Pareto-optimal designs, and in particular outperforms AlexNet, NiN and the original SqueezeNet in both accuracy and computational effort.
Reaching a higher accuracy generally seems to be very expensive: In order to improve the top-5 accuracy by 6 percent-points compared to ZynqNet, GoogLeNet requires 3\x\ more MACC operations. The state-of-the-art Inception-ResNet-v2 requires more than 17\x\ more operations to improve the top-5 error by 11 percent-points. Of course, these last few percent-points towards \SI{100}{\percent} accuracy contain the hardest images in ImageNet and thus require an overproportional effort. However, this implies that every actual application should precisely assess whether these last few percents of accuracy are actually required, or if orders of magnitude of computational effort can be saved in compromise.

\paragraph{Parameter Memory}
Another measure of resource efficiency can be seen in the \emph{Top-5 Error vs. Parameter Size} graph in \cref{fig:cnn-topologies-comparison}. Here, a seemingly log-linear relationship between the number of parameters and the top-5 error of each model shows up: reducing the top-5 error by 5 percent-points requires approximately twice the number of weight parameters in all of the Pareto-optimal designs.
ZynqNet CNN is again one of the Pareto-optimal designs, which highlights its good efficiency with regard to the number of weight parameters used.

%
\section{ZynqNet FPGA Accelerator Performance}
\label{sec:fpga-performance}

First and foremost, the ZynqNet FPGA Accelerator is meant to be a proof-of-concept for the implementation of CNNs on the basis of an FPGA. The secondary goal targets a maximum throughput on the given small and low-power platform, and in consequence a good power efficiency. The design and implementation details have been thoroughly discussed in the previous \cref{chap:fpga-design}, and the chosen architecture was found to be optimal with regard to the number of arithmetic and memory operations required.
This section evaluates the finished design with regard to the factors \emph{resource utilization}, \emph{achieved clock frequency} and \emph{operation schedule}, which determine the throughput of the accelerator. Finally, a number of potential architectural optimizations are highlighted.

\subsection{Resource Utilization}
The final ZynqNet FPGA Accelerator contains $\npe = 16$ processing units, which concurrently operate on the calculation of different output feature maps. Each processing unit contains a fully pipelined 3\x3 multiply-accumulate unit with 9 separate floating-point multipliers and a subsequent adder tree for the summation of their products. This results in a total of 144 floating-point multipliers and 128 floating-point adders, which constitute the computational core of the accelerator.
The processing units are fed from on-chip caches. In total, up to \SI{1.7}{\mega\byte} parameters (\num{442000} single-precision floating-point weights) and \SI{133}{\kilo\byte} image data are buffered in the on-chip Block RAM.
When synthesized for the Zynq XC-7Z045 FPGA, this configuration results in the resource requirements and device utilization figures shown in \cref{tab:fpga-utilization}.
The fact that more than \SI{90}{\percent} of all Block RAM resources and more than \SI{80}{\percent} of the DSP slices are utilized highlights the good fit of the architecture to the given FPGA and is a result from the co-optimization of both the  FPGA architecture and the ZynqNet CNN.

\begin{table}[tbp]
\centering
\caption[FPGA Utilization Report]{Resource Requirements and FPGA Utilization of the ZynqNet FPGA Accelerator when synthesized for the Zynq XC-7Z045.}
\label{tab:fpga-utilization}
\begin{tabular}{@{}lrrrr@{}}
\toprule
resource    & Block RAM         & DSP Slices        & FF                & LUT               \\ \midrule
used        & 996               & 739               & 137\,k              & 154\,k          \\
available   & 1090              & 900               & 437\,k              & 218\,k          \\ \midrule
utilization & \SI{91}{\percent} & \SI{82}{\percent} & \SI{31}{\percent} & \SI{70}{\percent} \\ \bottomrule
\end{tabular}
\end{table}

\subsection{Maximum Clock Frequency}
Despite the high resource utilization and the resulting long paths in the interconnect, the ZynqNet FPGA Accelerator can still be synthesized for an adequate clock frequency of $f_\mathrm{max} = \SI{200}{\MHz}$. This is possible because the architecture fully distributes the computation as well as all the required data onto the different computational units. There are no dependencies between the individual computational units, even their results are accumulated separately. This leads to mostly local routing and few global interconnections, all of which can be sufficiently pipelined.

\subsection{Operation Schedule}
The last factor that determines the ZynqNet FPGA Accelerator's throughput is the efficiency of the operation schedule. The nested loops that form the system's algorithmic basis principally allow a fully pipelined operation, where new inputs are fetched and processed in every clock cycle. There are no data dependencies or feedback loops in the architecture that could prevent pipelining within a single convolutional layer.

\paragraph{Pipeline Flushing Issue in Vivado HLS 2016.2} An ideal processing pipeline also requires correspondingly efficient control logic and scheduling.
When using High-Level Synthesis, the state machine that determines the operation schedule is automatically derived from the software model during synthesis.
Unfortunately, as described in \cref{sec:dataflow-and-pipeline-problem}, Vivado HLS 2016.2 has an issue with the derivation of an efficient operation schedule when pipelined regions are nested within dataflow sections, which are themselves part of an outer loop.
In such situations, the scheduler flushes the complete inner pipeline in each iteration of the outer loop --- something which is diametrically opposed to the idea of a pipelined core.
This HLS-related deficiency strikes a weak spot in the ZynqNet FPGA Accelerator architecture, and results in a total slow-down of a factor of 6.2\x\ across all ZynqNet CNN layers (see \cref{tab:slow-down-calculation-flushing-issue} in the appendix for the calculation).
Therefore, the FPGA currently spends more than \SI{80}{\%} of its time flushing the innermost pipeline rather than performing any useful operations. The situation is worst for layers with a small number of output channels, where all channels can be calculated in one or two clock cycles using the 16 parallel processing units. The computation-to-flushing ratio is then as bad as $1:63$ or $2:64$.
If the pipelining would function correctly, the computation of these layers would be limited by the time it takes to prefetch a new image patch (currently 9 clock cycles, with room for optimizations).

\pagebreak

\subsection{Potential Improvements}
\label{sec:fpga-accelerator-improvements}
\begin{enumerate}
  \item The \emph{Pipeline Flushing Issue} is by far the most pressing problem. Most other optimizations only make sense when the pipelining functions correctly. Besides waiting for a fix in a future version of Vivado HLS, the only workaround is the implementation of the architecture in RTL code. Correct pipelining should improve the FPGA accelerator performance by a factor of 6.2.
  \item The incorporation of \emph{fixed-point arithmetic} is the second most important issue. A quick test synthesis with Vivado HLS indicates the potential to {save \SI{50}{\percent} of the Block RAMs and \SI{80}{\percent} of the DSP slices by using a 16-bit fixed-point data format}. One DSP slice suffices to calculate a multiply-accumulate operation in 16-bit fixed-point format, which allows 5\x\ more processing units on the same FPGA fabric. However, the potential for parallelization in the output channels is mostly used up and moderate architectural changes would be necessary to tap the potential for parallelization in the input channels.\footnote{Bit-widths smaller than 16 bits might be feasible from the CNN side, but would not allow further parallelization due to the lack of further DSP resources in the FPGA fabric. The DSP48E1 slices in the Zynq-7000 family do not support single-instruction-multiple-data (SIMD) for the multiplication of smaller data types.}
  \item An architectural bottleneck can be seen in the \emph{prefetching of image pixels from the image cache}. Although this task is executed in parallel to the actual output channel calculation to hide the prefetch latency, the current delay of 9 clock cycles is relatively long. The architecture of the image cache might need to be improved to allow for more parallel read accesses, or a register-field might be used to cache the active image patch. This should be viable as the image cache occupies less than \SI{8}{\percent} of the Block RAMs, and a total of 300\,k flipflops are still unused. An ideal image cache would have a latency of less than 5 clock cycles, which would result in a speedup factor of 1.4.
  \item A further architectural optimization concerns the \emph{removal of the Global Pooling Cache}. As the latest CNN training experiments have shown, the ReLU nonlinearity in the last convolutional layers does not influence the overall classification accuracy (see \cref{tab:training-overview} in the appendix). It is therefore possible to use the existing Output Cache for the pixel-wise accumulation during global average pooling. The Global Pooling Cache can be omitted, freeing approximately 16 Block RAMs and 5 DSP slices.
  \item Finally, \emph{1\x1 convolutions} are currently not implemented efficiently: a full 3\x3 MACC unit is used for the single necessary multiplication. The potential overall speedup from utilizing all 9 multipliers in the MACC units for individual 1\x1 convolutions is approximately 1.2 with the current prefetch latency of 9 clock cycles. With an ideal Image Cache, a speedup factor of nearly 1.5 could be achieved.
\end{enumerate}
In the ideal case, the incorporation of all these improvements would increase the ZynqNet FPGA Accelerator throughput by a factor of almost 64. 




\clearpage\section{System Performance}
\label{sec:system-performance}

The ZynqNet Embedded CNN has been completely assembled and successfully taken into operation on a SCS Zynqbox. The full test system consists of
\begin{itemize}
  \item SCS Zynqbox (Zynq XC-7Z045 with \SI{1}{\giga\byte} DDR3 memory), running under Linux\footnote{A custom Debian-based distribution with Linux Kernel version 3.12.0.}
  \item ZynqNet CNN network description and trained weights, copied to the Zynqbox
  \item ZynqNet FPGA Accelerator bitstream, loaded into the FPGA fabric
  \item ZynqNet \texttt{/dev/mem} driver, connected to the AXI4-Lite configuration bus and the shared main memory
  \item ZynqNet CPU-side application, feeding the input images, launching the FPGA accelerator, measuring the timing and checking the classification results.
\end{itemize}
Using the above system configuration, the ZynqNet Embedded CNN has been evaluated in a realistic embedded scenario. The following final sections take a look at the overall system performance in terms of throughput and power efficiency.

\subsection{Throughput}
The embedded CNN's throughput is measured in terms of \emph{images per second}. In a typical scenario, the CNN accelerator is configured with the network description and the trained weights beforehand, and is then utilized to classify an incoming stream of images. Therefore, the run-time per frame is measured from the moment when the FPGA accelerator is started, to the moment when the calculation of the Softmax Classification layer is finished. The ARM CPUs take $t_{CPU} = \SI{45}{\s}$ to calculate the ZynqNet CNN using the software model, with all optimizations and hardware floating-point support enabled.
The current version of the ZynqNet FPGA Accelerator requires $t_F = \SI{1955}{\milli\second}$ per frame, which corresponds to a frame rate of $r_F=\SI{0.51}{FPS}$.
There are two important limiting factors at play in this result:
\begin{enumerate}
  \item the FPGA clock rate \texttt{FCLK\_CLK0} has been configured to \SI{100}{\MHz} instead of \SI{200}{\MHz}
  \item the Pipeline Flushing Issue slows the design down by a factor of $s\approx6.2$.
\end{enumerate}
It is safe to assume that with these two issues corrected, the design would reach $t_F' \approx \SI{158}{\milli\second}$ per frame, and a reasonable real-time frame rate of $r_F'=\SI{6.3}{FPS}$. Additionally, switching to a 16-bit fixed-point data format could potentially boost the frame rate to \SI{30}{FPS}. With all improvements from \cref{sec:fpga-accelerator-improvements} implemented, the frame rate could ideally reach \SI{65}{FPS}.

\subsection{Power Efficiency}
The energy consumption of the complete Zynqbox platform running the ZynqNet Embedded CNN has been evaluated using a Fluke 177 Multimeter and a TTi EX1810R laboratory power supply. The power measurements include all conversion losses, peripheral devices, as well as the system fan and can be found in \cref{tab:fpga-power}. The system has not been optimized for low-power operation due to the advanced project time, and a significant amount of energy is already consumed in the idle state.

All measurements regarding the ZynqNet FPGA Accelerator's power dissipation have to be considered with caution due to the presence of the Pipeline Flushing Issue.
The issue might substantially reduce the amount of switching activity in the FPGA fabric by causing zeros to be flushed through the computation pipeline, and might thereby distort the energy consumption. It is therefore currently not possible to make any precise statements regarding the system's power efficiency.

It is however relatively safe to assume a power consumption well below \SI{20}{\watt} even under maximum load.\footnote{For example, consider that the Texas Instruments PMP8251 Power Management Reference Design for the Zynq-7000 Platform is dimensioned for a maximum power consumption of \SI{23}{\watt} \cite{ti-pmp8251}. The ZynqNet FPGA Accelerator utilizes neither transceivers, I/Os nor the additional memory interfaces, and almost no peripherals, which would all cost considerable amounts of energy.}
With a moderate assumption of $P = \SI{12}{\watt}$ system power\footnote{Based on estimations using the Xilinx Power Estimator (XCE) tool \cite{zynq-power-estimator}.}
 and $r_F'=\SI{6.3}{FPS}$, the ZynqNet FPGA Accelerator's power efficiency would be at
\begin{equation}
  \eta_\text{\,ZynqNet} = \frac{r_F'}{P} = \frac{\SI{6.3}{frames}}{\SI{12}{\watt\second}} \approx \SI{0.53}{images/\joule}
\end{equation}
The NVidia Jetson TX1 Whitepaper \cite{nvidia-jetson-whitepaper} provides some context for this number: AlexNet computed on a Intel Core i7 CPU reaches an efficiency $\eta_\text{\,Corei7}=\SI{1.3}{images/\joule}$, the same CNN on a NVidia Titan X $\eta_\text{\,TitanX} = \SI{2.5}{images/\joule}$ and on a NVidia Tegra X1 $\eta_\text{\,TegraX1}=\SI{8.6}{images/\joule}$. Although these figures probably do not consider conversion losses and the total system power, they show that further improvements of the FPGA accelerator, such as those presented in \cref{sec:fpga-accelerator-improvements}, are unavoidable if the embedded system requires best-in-class power efficiency. With all known improvements (corrected pipeline flushing, 16-bit fixed-point arithmetic, improved image caching and ideal 1x1 convolutions) applied, the power efficiency could possibly be boosted to a respectable $\eta_\text{\,improved} = \sfrac{\SI{65}{frames}}{\SI{12}{\watt\second}} \approx \SI{5.4}{images/\joule}$.

\begin{table}[tbp]
\centering
\caption[System Power Measurement Results]{Power Measurement Results for the Zynqbox Platform running ZynqNet Embedded CNN. The total System Power includes all Conversion Losses, Peripherals and the System Fan.}
\label{tab:fpga-power}
\begin{tabular}{@{}lrr@{}}
\toprule
system state              &   current draw @\SI{12}{\volt}    &     power dissipation  \\
\midrule
system idle               &   \SI{486}{\milli\ampere}         &     \SI{5.83}{\watt}    \\
CPU cores under full load &   \SI{502}{\milli\ampere}         &     \SI{6.02}{\watt}    \\
FPGA accelerator idle     &   \SI{622}{\milli\ampere}         &     \SI{7.46}{\watt}    \\
FPGA accelerator running  &   \SI{650}{\milli\ampere}         &     \SI{7.80}{\watt}    \\
\bottomrule
\end{tabular}
\end{table}
\chapter{Conclusion}
\label{chap:conclusion}



In this master thesis, I designed and implemented a proof-of-concept FPGA-accelerated embedded Convolutional Neural Network. The \emph{ZynqNet Embedded CNN} is designed for image classification on the ImageNet dataset and consists of two main components: \emph{ZynqNet CNN}, a highly optimized and customized CNN topology, and the \emph{ZynqNet FPGA Accelerator}, a FPGA-based architecture for the evaluation of ZynqNet CNN.

\begin{enumerate}
	\item
\emph{ZynqNet CNN} is derived from the small and efficient SqueezeNet CNN topology.
A detailed network analysis and optimization using the custom-designed \emph{Netscope CNN Analyzer} tool has enabled a reduction of the classification error by \SI{20}{\%} relative to SqueezeNet. At the same time, the ZynqNet CNN requires \SI{38}{\%} less multiply-accumulate operations. Its topology is highly regular and consists of just three layer types: convolutional layers, ReLU nonlinearities and a global average pooling layer.
Further, all layer dimensions have been converted to powers of two, which enables optimizations in the cache and memory addressing on the FPGA. Finally, the individual layers have been shaped to fit ideally onto the on-chip caches in the FPGA architecture.


\item
The \emph{ZynqNet FPGA Accelerator} is an FPGA-based architecture which allows the efficient evaluation of ZynqNet CNN and similar networks. It accelerates the convolutional layers, which encompass \SI{99.3}{\%} of all required operations, as well as the ReLU nonlinearities and the global average pooling.
The FPGA architecture benefits from the optimized CNN topology and is conceptually simple. Nevertheless, it supports a nested-loop algorithm which minimizes the number of arithmetic operations and memory accesses necessary for the evaluation of ZynqNet CNN, and can therefore be considered ideal.
The FPGA accelerator has been synthesized using Vivado High-Level Synthesis for the Xilinx Zynq XC-7Z045 System-on-Chip, and reaches a clock frequency of \SI{200}{\MHz} with a device utilization of \SI{80}{\%} to \SI{90}{\%}.\footnote{The application of High-Level Synthesis has been an interesting and instructive, yet also adventurous journey, which has been extensively detailed in this report for the benefit of later users.}
\end{enumerate}

The ZynqNet Embedded CNN has been assembled into a fully working proof-of-concept system on the Xilinx Zynq-7000 All Programmable platform.
This project clearly demonstrates the feasibility of FPGA-based embedded CNN implementations.
The current solution already exhibits a reasonable performance, and a number of opportunities for further gains in throughput and power efficiency have been pointed out.


The tough requirements of embedded CNNs regarding the size, efficiency and computational power of the underlying computing platform are very hard to meet with the systems available today. A number of different platforms can be considered for future implementations, and by now it is not clear which one will conquer this market.
Even though the presented ZynqNet Embedded CNN does not yet provide the massive amounts of computational power required for future applications of embedded image understanding, it may still serve as a stepping stone and a guide for further explorations of the FPGA as a platform for embedded CNNs.
The biggest advantage of these FPGA-based systems can be seen in their scalability. Using a larger device, much higher performance can be attained at comparable efficiency figures, while most other platforms are inherently limited by the amount of computational power available on a given chip.
FPGAs therefore provide a promising path towards the vision of powerful embedded CNNs and the abundance of fascinating applications which could profit from on-board image understanding --- and the ZynqNet Embedded CNN may be a first small step on this path.

I am looking forward to the exciting times ahead in this fast-paced field of research.

\cleardoublepage


\begin{appendices}
\let\cleardoublepage\clearpage 
\crefalias{section}{appsec} 
\crefalias{chapter}{appsec}

\chapter{Declaration of Originality}
\label{sec:declaration-of-originality}

\begin{figure}[H]
  \centering
  \includegraphics[width=0.9\linewidth]{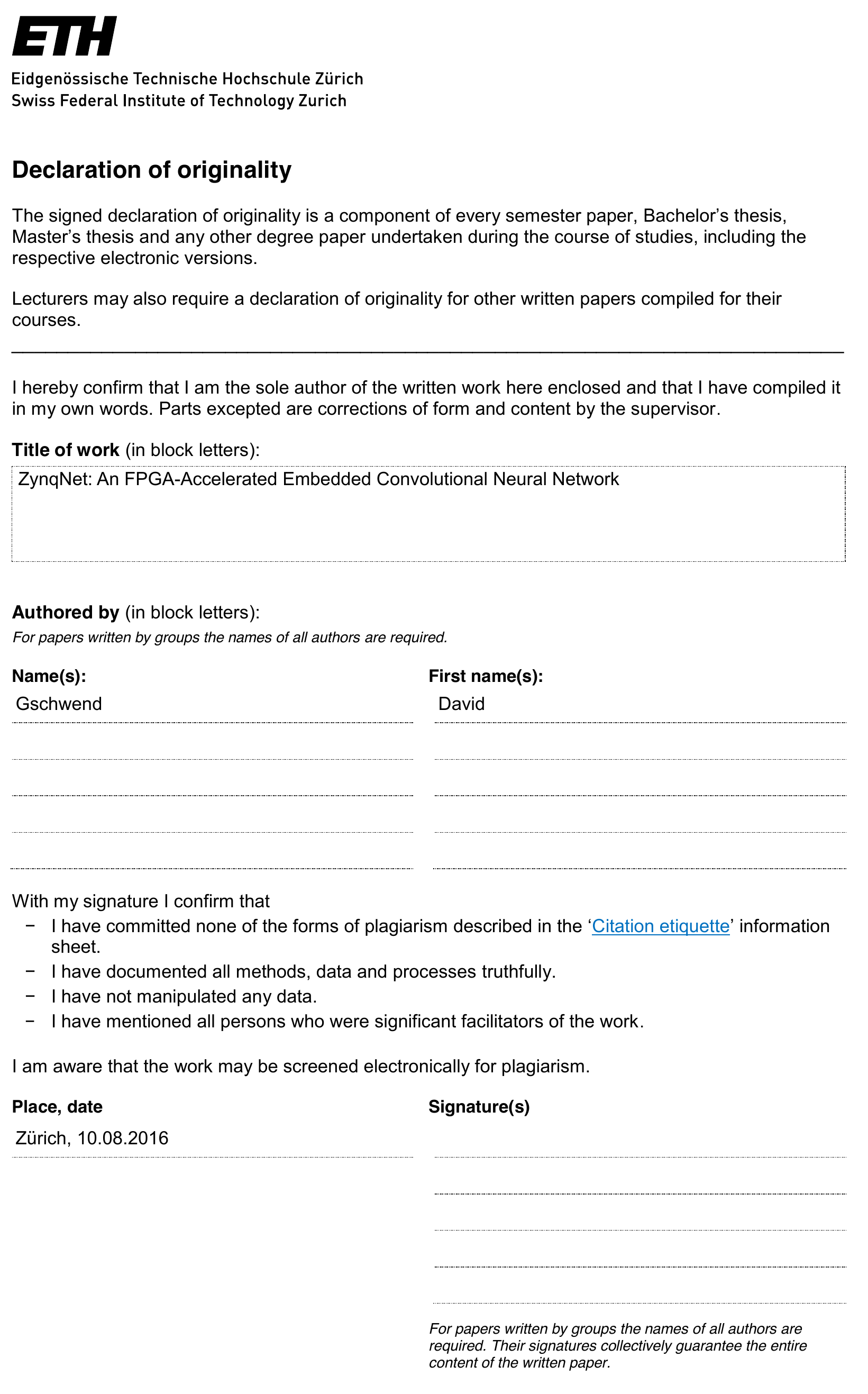}
\end{figure}

\chapter{Task Description}
\label{sec:task description}

\vspace*{-1.2cm}
\begin{figure}[H]
  \centering
  \includegraphics[width=0.96\linewidth]{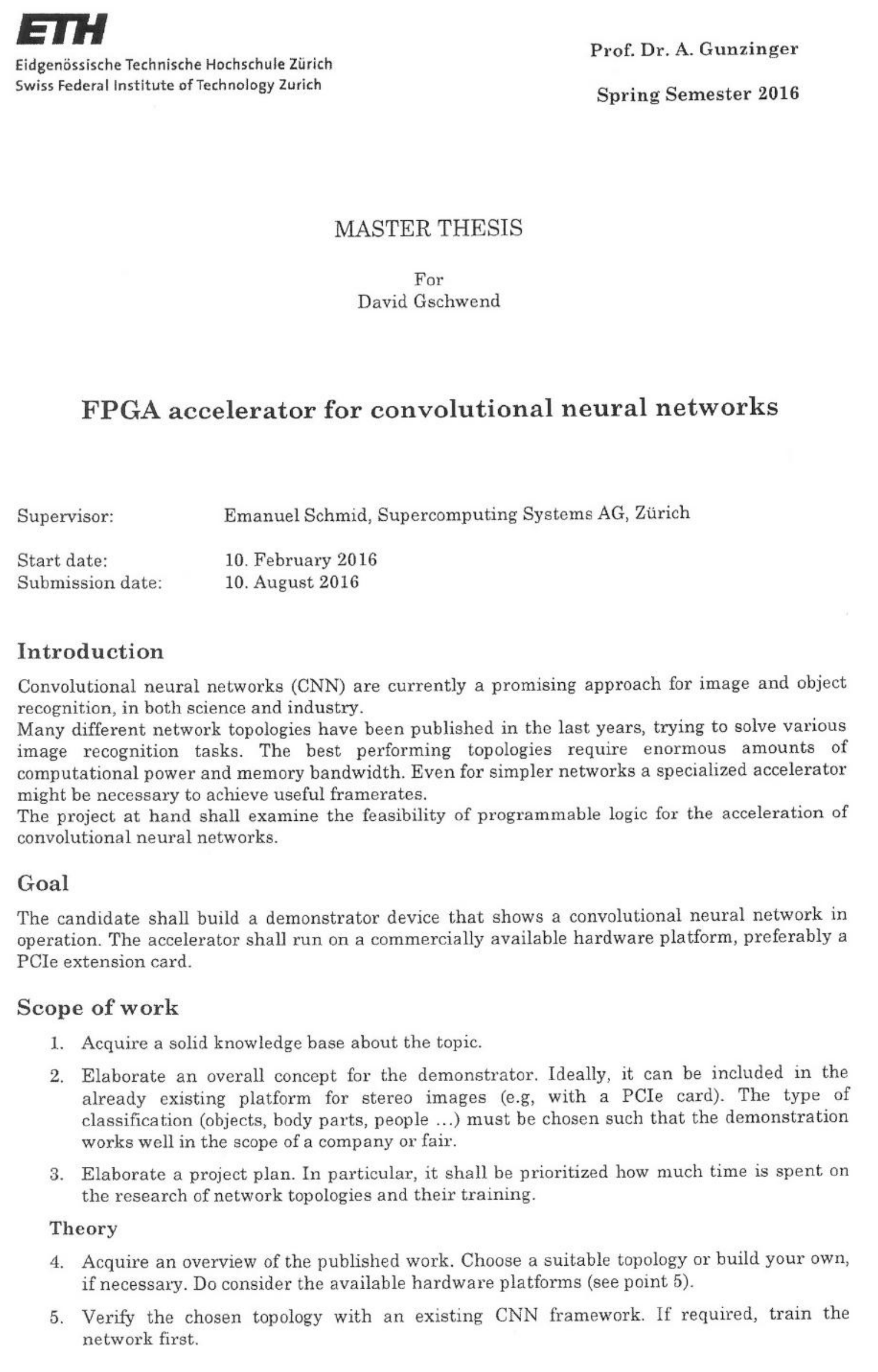}
\end{figure}
\pagebreak
\vspace*{2.5cm}
\begin{figure}[H]
  \centering
  \includegraphics[width=\linewidth]{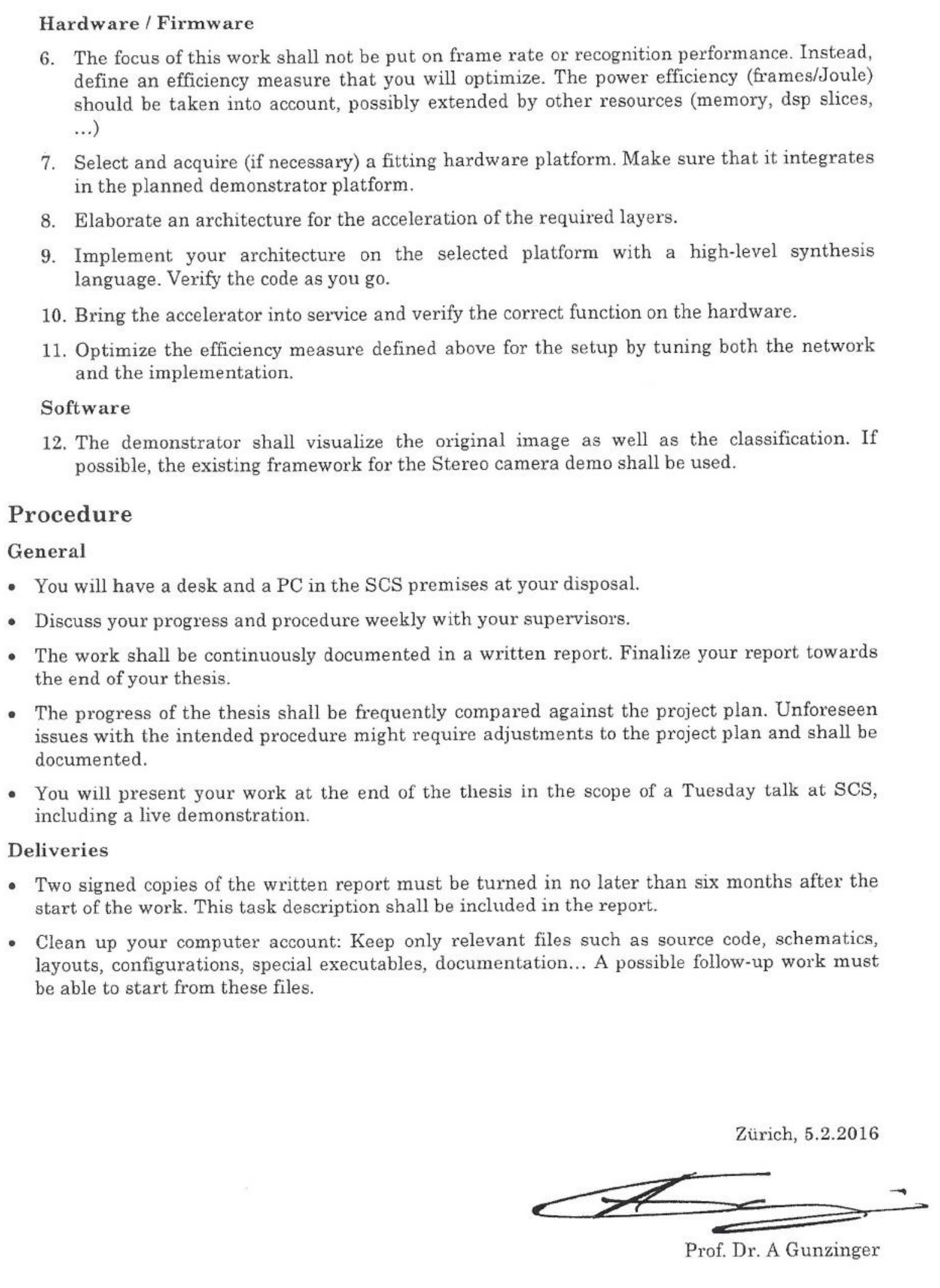}
\end{figure}

\chapter{Convolutional Neural Network Visualizations}

%
\section{3D Illustration of Convolutional Layers}
\label{sec:appendix-illustration-fire-module}

\begin{figure}[H]
  \centering
  \includegraphics[width=0.9\linewidth]{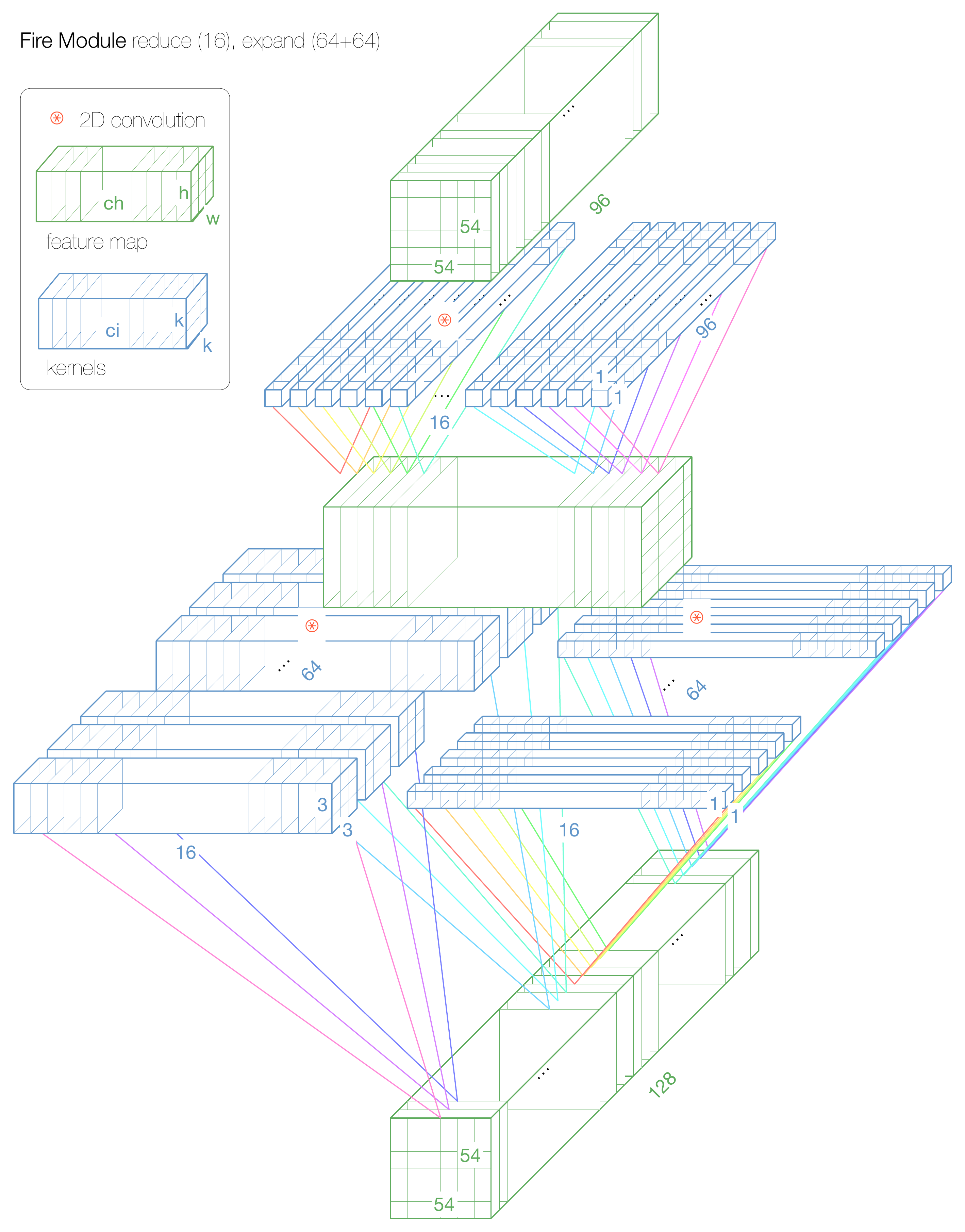}
    \vspace{0.5cm}

  \caption[3D Illustration of Convolutional Layers in a ZynqNet Fire Module]{3D Illustration of the Convolutional Layers in a SqueezeNet or ZynqNet Fire Module. Convolutional Layers can be seen as Transformations on 3D Volumes.}
  \label{fig:fire-module-3d}
\end{figure}

\section{Netscope Visualizations of Different CNN Topologies}
\label{sec:appendix-topology-visualization}

\begin{figure}[H]
  \centering
  \scalebox{0.93} {
      \begin{minipage}[c]{\textwidth}

      \includegraphics[width=0.5cm]{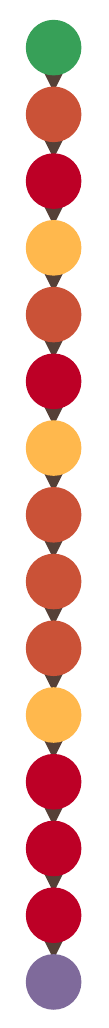} \hfill
      \includegraphics[width=0.5cm]{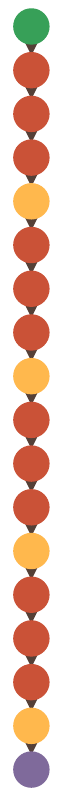} \hfill
      \includegraphics[width=0.5cm]{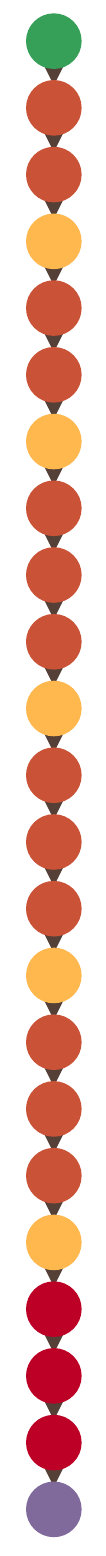} \hfill
      \includegraphics[width=2.5cm]{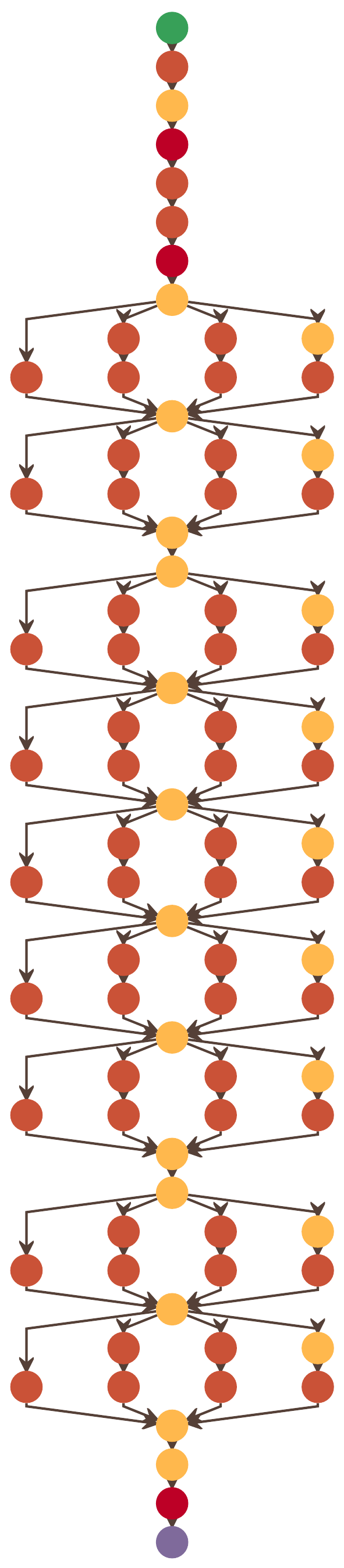} \hfill
      \includegraphics[width=1.13cm]{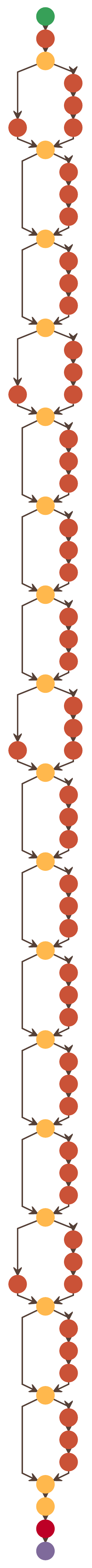} \hfill
      \includegraphics[width=2.35cm]{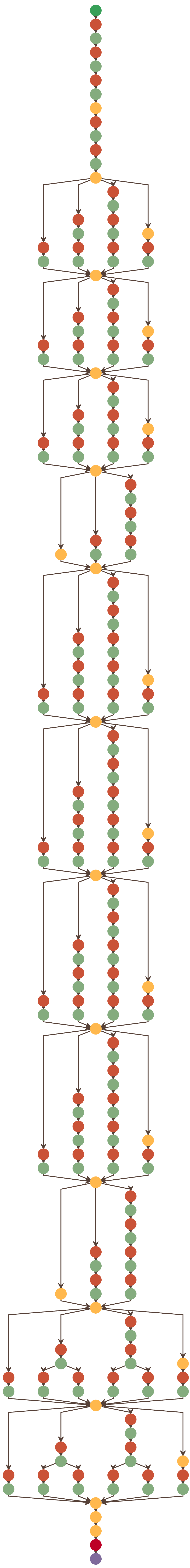} \hfill
      \includegraphics[width=1.84cm]{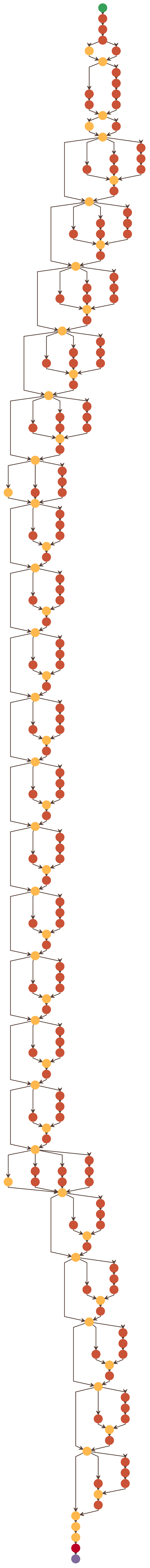} \hfill
      \includegraphics[width=1.15cm]{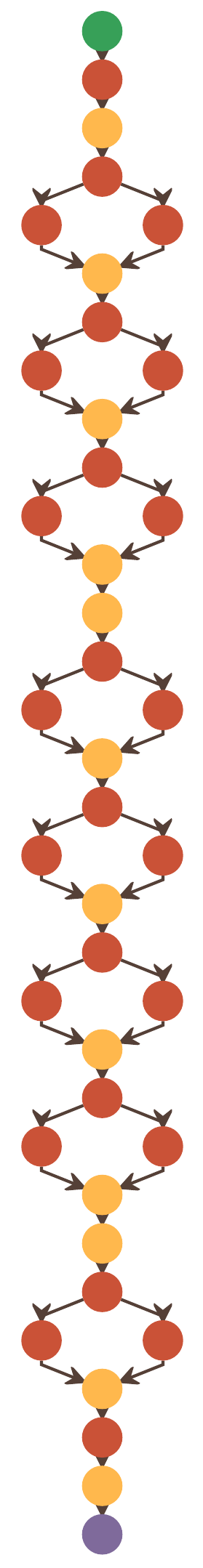} 


	  \end{minipage}
  }
  \caption[Netscope Visualizations of CNN Topologies from Prior Work]{Netscope Visualizations of CNN Topologies from Prior Work. Left to right: AlexNet, Network-in-Network, VGG-16, GoogLeNet, ResNet-50, Inception v3, Inception-ResNet-v2, SqueezeNet.}
  \label{fig:cnn-topologies-visualization}
\end{figure}

\begin{figure}[H]
  \centering
      \includegraphics[height=\textheight]{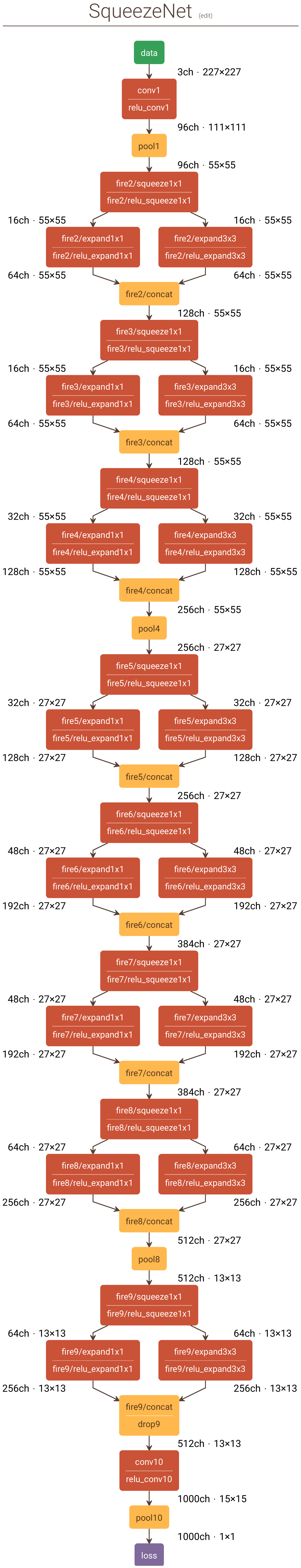} \qquad\qquad
      \includegraphics[height=\textheight]{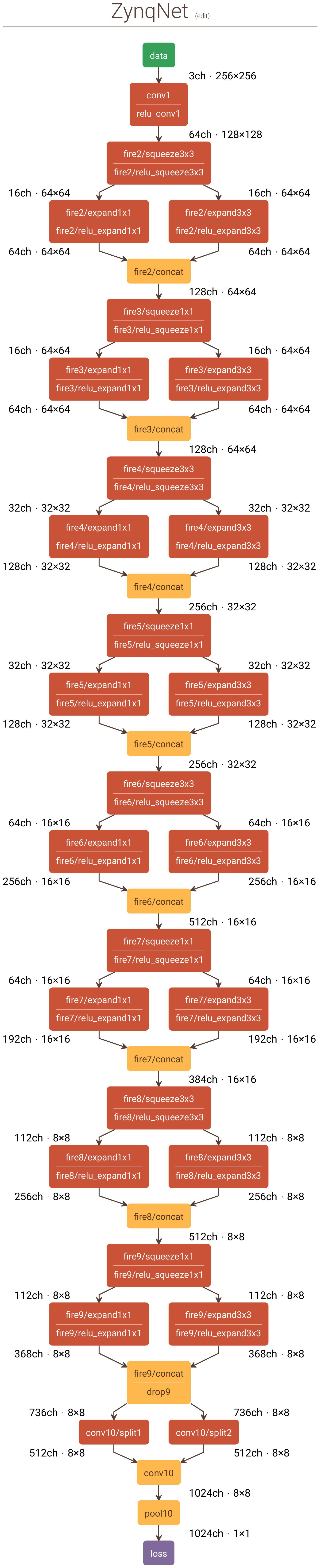}

  \caption[Detailed Netscope Visualizations of SqueezeNet and ZynqNet CNN]{Detailed Netscope Visualizations of the SqueezeNet and the ZynqNet CNN Topologies.}
  \label{fig:squeezenet-zynqnet-visualization}
\end{figure}

\clearpage
\section{Advanced Usage Tips and Restrictions for Netscope}
\label{sec:netscope-advanced-usage}

\paragraph{Advanced Usage Tips}
\begin{itemize}
  \item Clicking a layer in the network graph directly scrolls to its entry in the summary table and vice-versa.
  \item The \emph{edit} link next to the network title opens the \texttt{.prototxt} source code for the current CNN for editing.
  \item Shift-Enter in the Editor updates the graph and all tables.
  \item Naming layers according to the scheme "module/layer" groups these layers as one module in the summary table.
  \item Clicking ``Excel-Compatible Results'' at the very bottom opens a list with the most relevant layer characteristics, suited for further analysis in e.g. Excel or Matlab.
\end{itemize}

\paragraph{Current Restrictions}
\begin{itemize}
  \item In each layer, the field \texttt{top} needs to match the field \texttt{name}, except for \texttt{InPlace} layers where \texttt{top} matches \texttt{bottom}.
  \item \texttt{Data} and \texttt{Input} Layers are not accepted in all possible \texttt{.prototxt} syntaxes, refer to the built-in presets for valid examples.
\end{itemize}

\chapter{CNN Training Details and Results}

\section{Hardware Components of the CNN Training Workstations}
\label{sec:appendix-training-hw}

\begin{table}[H]
\centering
\caption[Hardware Components of the CNN Training Workstations]{Hardware Components used in the GPU-based CNN Training Workstations.}
\label{tab:training-hw-list}
\begin{tabular}{@{}rllr@{}}
\toprule
Count     & Component        & Type Name                                    & Price (CHF) \\ \midrule
2$\times$ & Graphics Card    & Gigabyte GTX Titan X XTREME (12GB)           & 2300.00        \\
1$\times$ & ATX Motherboard  & Gigabyte Z170XP-SLI                          & 150.00         \\
1$\times$ & Processor        & Intel Core i5 6400 Quad Core (2.70 GHz)      & 200.00         \\
4$\times$ & DRAM Memory      & Corsair Vengeance LPX 8GB DDR4-2400          & 150.00         \\
1$\times$ & Solid-State Disk & Kingston SSDNow V300 (120GB, System)         & 50.00         \\
1$\times$ & Solid-State Disk & Samsung 850 EVO Basic (500GB, Data)          & 160.00         \\
1$\times$ & Hard Disk Drive  & Western Digital Caviar Black (1TB, Archive)  & 70.00         \\
1$\times$ & Power Supply     & Cougar GX 800 V3 80 Plus Gold (800W)         & 120.00         \\
1$\times$ & PC Case          & Corsair Carbide 100R (Midi Tower)            & 60.00          \\ \midrule
          & Total            &                                              & 3070.00
\end{tabular}
\end{table}

\begin{figure}[H]
  \centering
  \includegraphics[width = 0.48\linewidth]{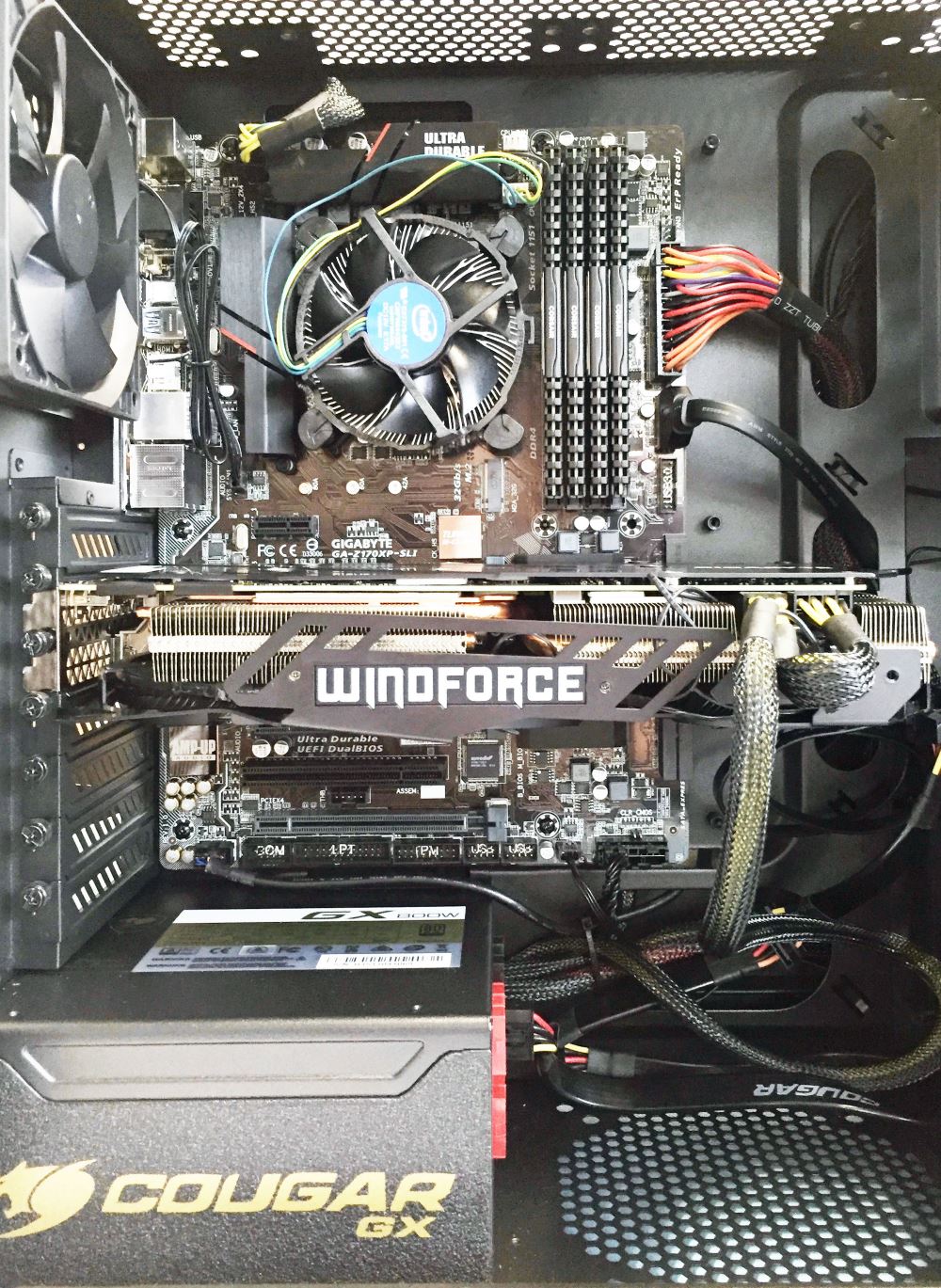}\quad
  \includegraphics[width = 0.48\linewidth]{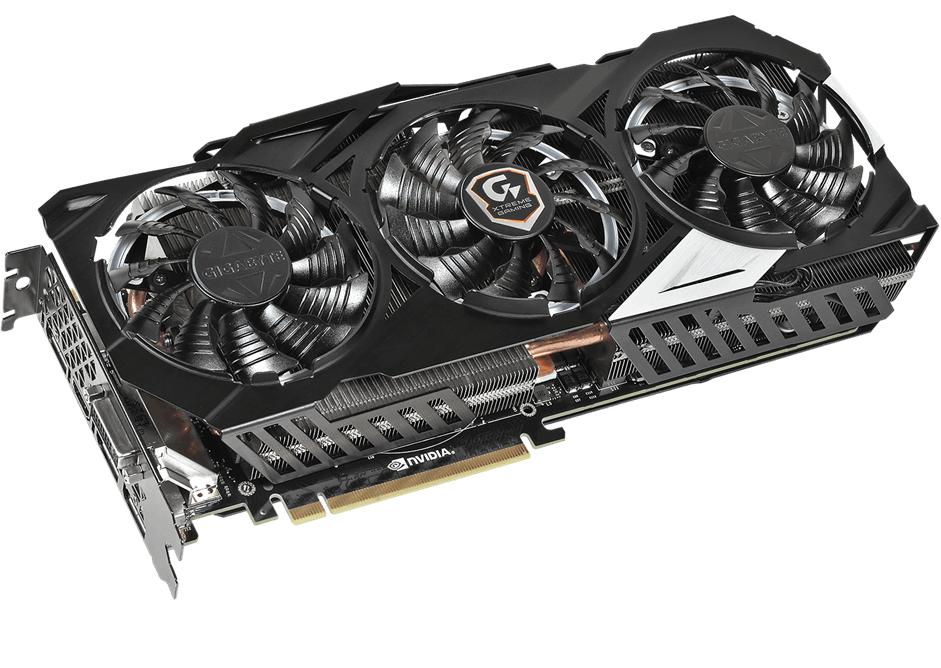}
  \caption[Photos of the GPU-based CNN Training Workstations]{Photograph of the custom-built CNN Training Workstation (with one of two NVidia GeForce GTX Titan X installed) and of a Titan X Graphics Card (GPU photo from \cite{titan-x-gpu-image}).}
  \label{fig:titan-x-hardware}
\end{figure}

%
\section{Screenshots from the DIGITS CNN Training Software}
\label{sec:appendix-digits-screenshots}

\begin{figure}[H]
  \makebox[\textwidth][c]{ 
  \scalebox{1.1}{\begin{minipage}{\linewidth}
  \framebox{\includegraphics[height = 5.25cm]{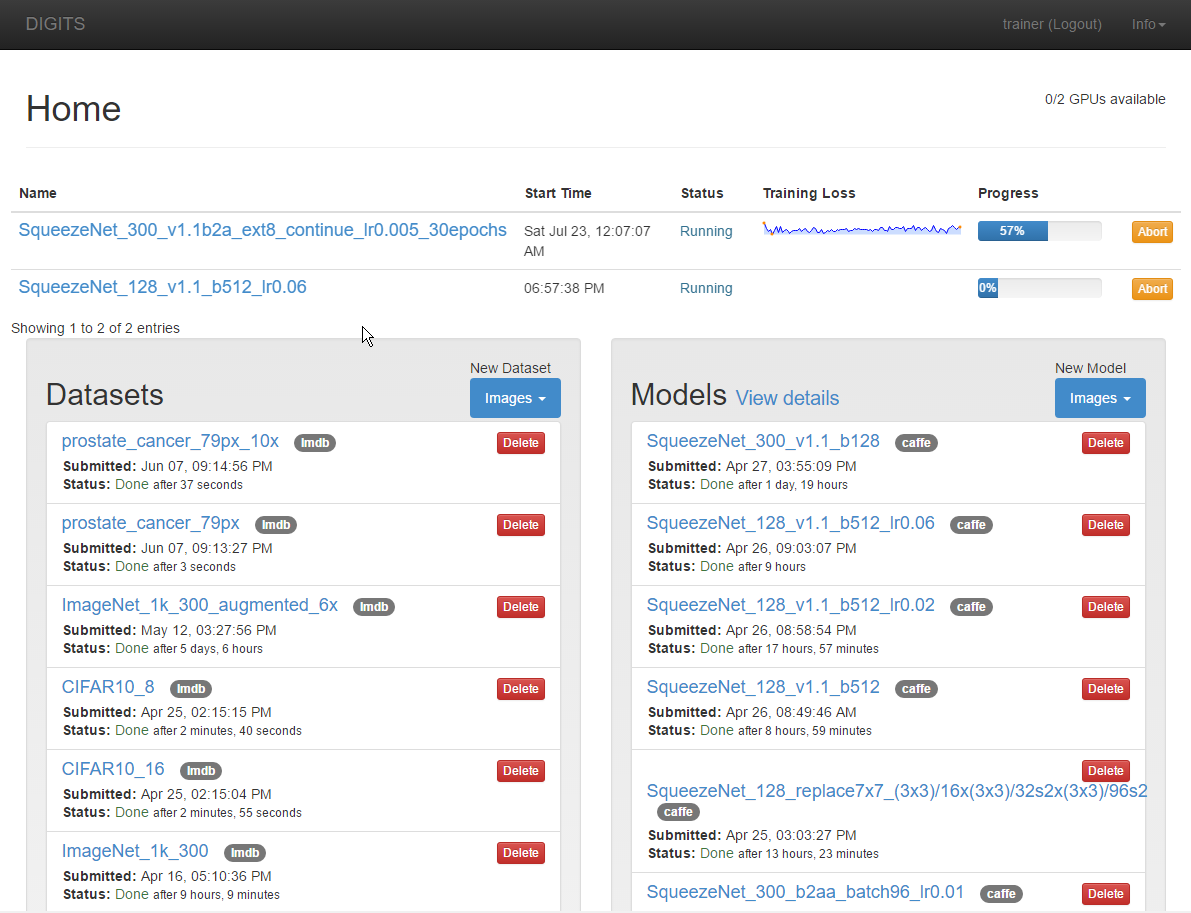}}\quad
  \framebox{\includegraphics[height = 5.25cm]{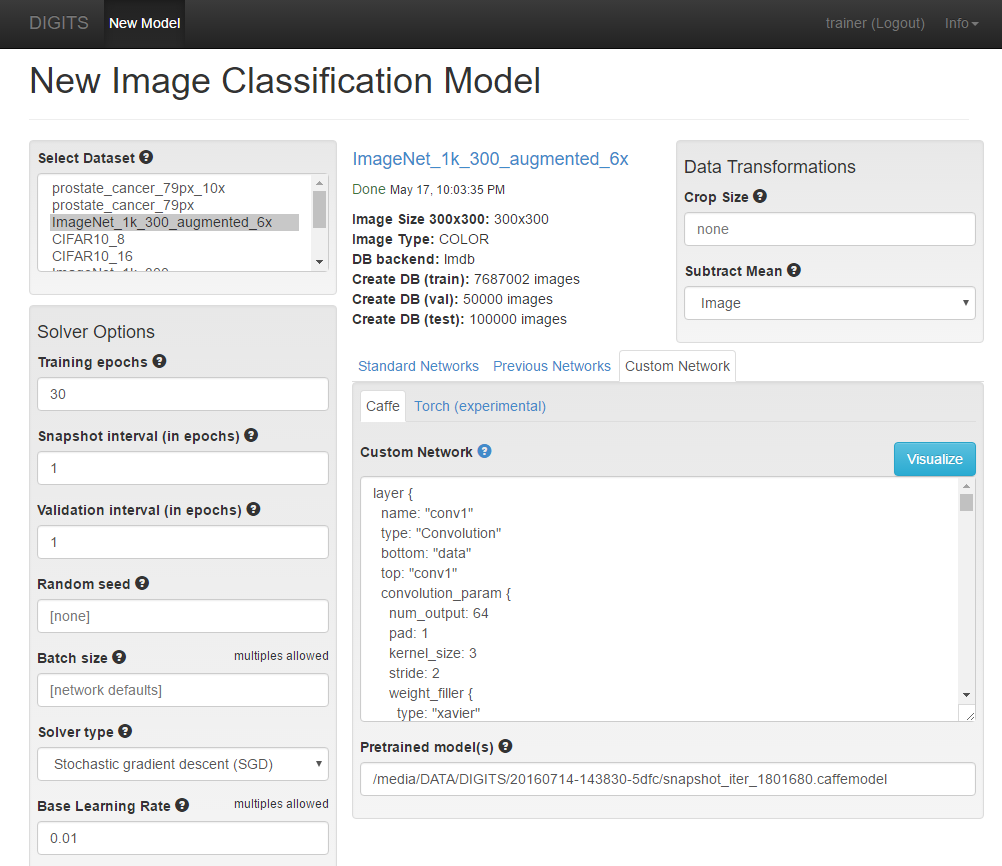}}\\[3mm]

  \framebox{\includegraphics[height = 7.5cm]{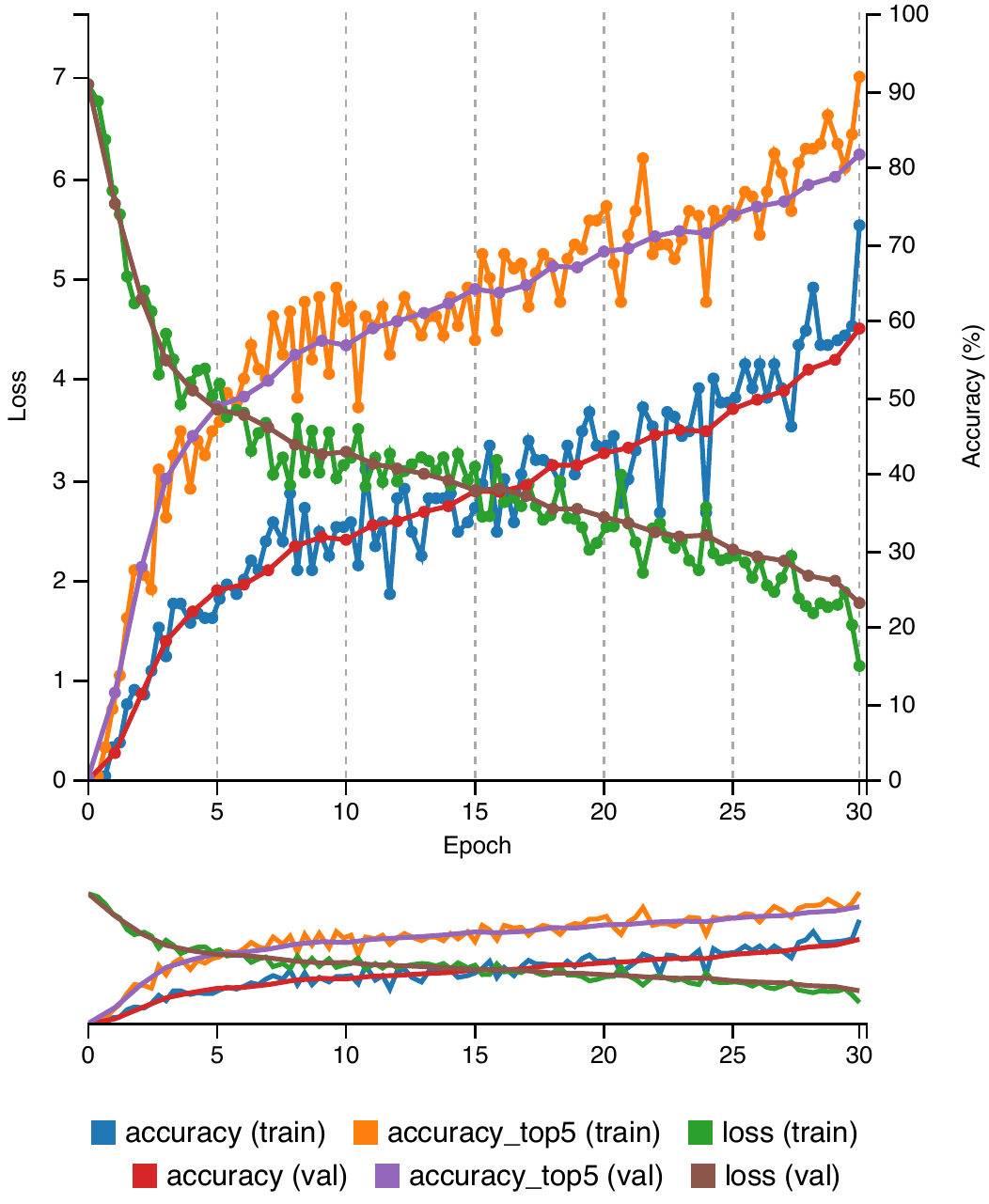}}\quad
  \framebox{\includegraphics[height = 7.5cm]{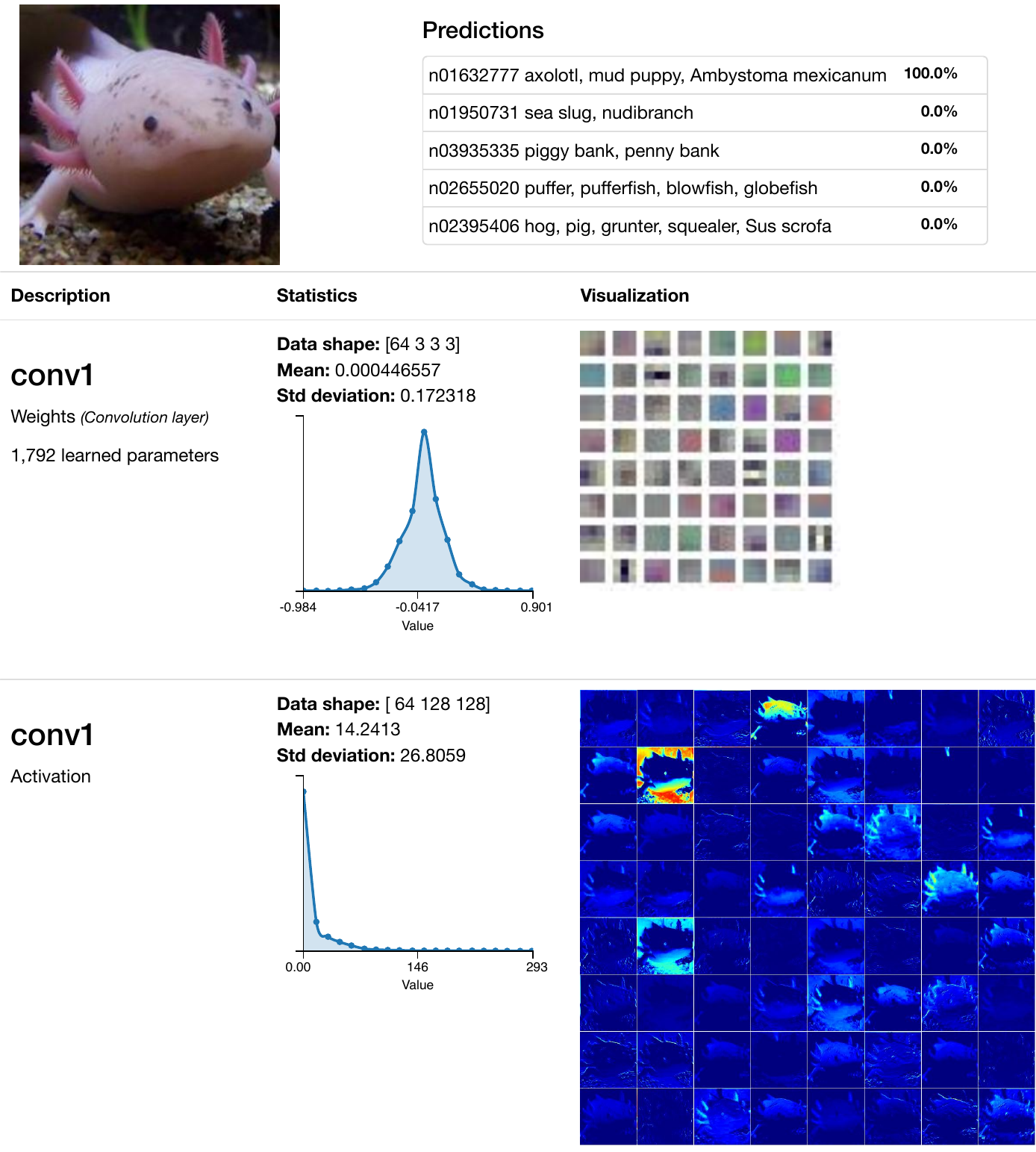}}
  \end{minipage}}}
  \caption[Screenshots from the DIGITS CNN Training Software]{Screenshots from the DIGITS v3.4 CNN Training Software, showing the job schedule and dataset/model management on the home page (top left), the model definition interface (top right), a training progress chart with decreasing loss and increasing accuracy figures (bottom left), as well as the visualization of weights and activations in a trained network (bottom right).}
  \label{fig:digits_screenshots}
\end{figure}

\section{Overview of all CNN Training Experiments}
\label{sec:appendix-training-overview}

\vspace{-4mm}

  \begin{table}[H]

    \makebox[\textwidth][c]{ 
      \begin{minipage}{1.1\linewidth} 

      \caption[Overview of all CNN Training Experiments]{Overview of all Experiments conducted during CNN Training. ZynqNet CNN is listed as \emph{SqueezeNet\_300\_v1.1\_b2a\_ext8}.}
      \label{tab:training-overview}

        \includegraphics[page=1,scale=0.6,trim={1.2cm 2cm 2cm 2cm},clip]{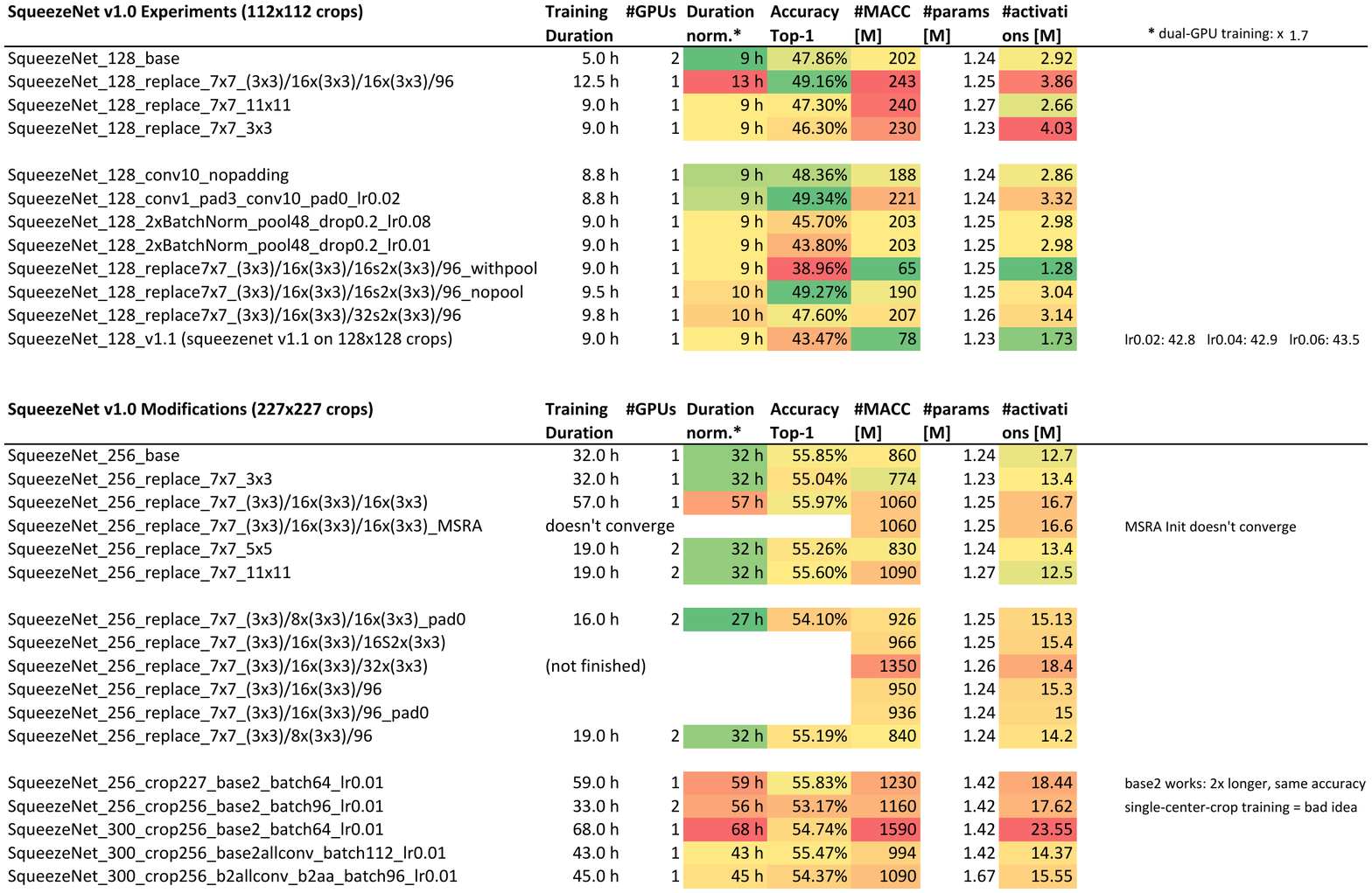}\\
        \includegraphics[page=2,scale=0.6,trim={1.2cm 2.95cm 2cm 2cm},clip]{figures/overview_trainings}

      \end{minipage}
    }

  \end{table}

\section{Layer Description Table for ZynqNet CNN}
\label{sec:appendix-squeezenet-table}
\vspace{-4mm}

\begin{table}[H]
\centering
\caption[Layer Description Table for ZynqNet CNN]{Detailed Description of all ZynqNet CNN Layers and their Parameters.}
\label{tab:zynqnet-description}
\scalebox{0.75}{
\def\arraystretch{0.8}
  \begin{tabular}{@{}lllrrrrcrcl@{}}
  \toprule
  ID & Name                   & Type        & Kernel & Stride & Pad & CH in        & W\x H in   & CH out       & W\x H out   & Notes               \\ \midrule
  1  & data                   & Data        &        &        &     &              &            & 3            & 256\x256    &                     \\
  2  & conv1                  & Convolution & 3\x3   & 2      & 1   & 3            & 256\x256   & 64           & 128\x128    &                     \\
  3  & relu\_conv1            & ReLU        &        &        &     & 64           & 128\x128   & 64           & 128\x128    &                     \\
  4  & fire2/squeeze3x3       & Convolution & 3\x3   & 2      & 1   & 64           & 128\x128   & 16           & 64\x64      &                     \\
  5  & fire2/relu\_squeeze3x3 & ReLU        &        &        &     & 16           & 64\x64     & 16           & 64\x64      &                     \\
  6  & fire2/expand1x1        & Convolution & 1\x1   & 1      & 0   & 16           & 64\x64     & 64           & 64\x64      &                     \\
  7  & fire2/relu\_expand1x1  & ReLU        &        &        &     & 64           & 64\x64     & 64           & 64\x64      &                     \\
  8  & fire2/expand3x3        & Convolution & 3\x3   & 1      & 1   & 16           & 64\x64     & 64           & 64\x64      &                     \\
  9  & fire2/relu\_expand3x3  & ReLU        &        &        &     & 64           & 64\x64     & 64           & 64\x64      &                     \\
  10 & fire2/concat           & Concat      &        &        &     & 128          & 64\x64     & 128          & 64\x64      &                     \\
  11 & fire3/squeeze1x1       & Convolution & 1\x1   & 1      & 0   & 128          & 64\x64     & 16           & 64\x64      &                     \\
  12 & fire3/relu\_squeeze1x1 & ReLU        &        &        &     & 16           & 64\x64     & 16           & 64\x64      &                     \\
  13 & fire3/expand1x1        & Convolution & 1\x1   & 1      & 0   & 16           & 64\x64     & 64           & 64\x64      &                     \\
  14 & fire3/relu\_expand1x1  & ReLU        &        &        &     & 64           & 64\x64     & 64           & 64\x64      &                     \\
  15 & fire3/expand3x3        & Convolution & 3\x3   & 1      & 1   & 16           & 64\x64     & 64           & 64\x64      &                     \\
  16 & fire3/relu\_expand3x3  & ReLU        &        &        &     & 64           & 64\x64     & 64           & 64\x64      &                     \\
  17 & fire3/concat           & Concat      &        &        &     & 128          & 64\x64     & 128          & 64\x64      &                     \\
  18 & fire4/squeeze3x3       & Convolution & 3\x3   & 2      & 1   & 128          & 64\x64     & 32           & 32\x32      &                     \\
  19 & fire4/relu\_squeeze3x3 & ReLU        &        &        &     & 32           & 32\x32     & 32           & 32\x32      &                     \\
  20 & fire4/expand1x1        & Convolution & 1\x1   & 1      & 0   & 32           & 32\x32     & 128          & 32\x32      &                     \\
  21 & fire4/relu\_expand1x1  & ReLU        &        &        &     & 128          & 32\x32     & 128          & 32\x32      &                     \\
  22 & fire4/expand3x3        & Convolution & 3\x3   & 1      & 1   & 32           & 32\x32     & 128          & 32\x32      &                     \\
  23 & fire4/relu\_expand3x3  & ReLU        &        &        &     & 128          & 32\x32     & 128          & 32\x32      &                     \\
  24 & fire4/concat           & Concat      &        &        &     & 256          & 32\x32     & 256          & 32\x32      &                     \\
  25 & fire5/squeeze1x1       & Convolution & 1\x1   & 1      & 0   & 256          & 32\x32     & 32           & 32\x32      &                     \\
  26 & fire5/relu\_squeeze1x1 & ReLU        &        &        &     & 32           & 32\x32     & 32           & 32\x32      &                     \\
  27 & fire5/expand1x1        & Convolution & 1\x1   & 1      & 0   & 32           & 32\x32     & 128          & 32\x32      &                     \\
  28 & fire5/relu\_expand1x1  & ReLU        &        &        &     & 128          & 32\x32     & 128          & 32\x32      &                     \\
  29 & fire5/expand3x3        & Convolution & 3\x3   & 1      & 1   & 32           & 32\x32     & 128          & 32\x32      &                     \\
  30 & fire5/relu\_expand3x3  & ReLU        &        &        &     & 128          & 32\x32     & 128          & 32\x32      &                     \\
  31 & fire5/concat           & Concat      &        &        &     & 256          & 32\x32     & 256          & 32\x32      &                     \\
  32 & fire6/squeeze3x3       & Convolution & 3\x3   & 2      & 1   & 256          & 32\x32     & 64           & 16\x16      &                     \\
  33 & fire6/relu\_squeeze3x3 & ReLU        &        &        &     & 64           & 16\x16     & 64           & 16\x16      &                     \\
  34 & fire6/expand1x1        & Convolution & 1\x1   & 1      & 0   & 64           & 16\x16     & 256          & 16\x16      &                     \\
  35 & fire6/relu\_expand1x1  & ReLU        &        &        &     & 256          & 16\x16     & 256          & 16\x16      &                     \\
  36 & fire6/expand3x3        & Convolution & 3\x3   & 1      & 1   & 64           & 16\x16     & 256          & 16\x16      &                     \\
  37 & fire6/relu\_expand3x3  & ReLU        &        &        &     & 256          & 16\x16     & 256          & 16\x16      &                     \\
  38 & fire6/concat           & Concat      &        &        &     & 512          & 16\x16     & 512          & 16\x16      &                     \\
  39 & fire7/squeeze1x1       & Convolution & 1\x1   & 1      & 0   & 512          & 16\x16     & 64           & 16\x16      &                     \\
  40 & fire7/relu\_squeeze1x1 & ReLU        &        &        &     & 64           & 16\x16     & 64           & 16\x16      &                     \\
  41 & fire7/expand1x1        & Convolution & 1\x1   & 1      & 0   & 64           & 16\x16     & 192          & 16\x16      &                     \\
  42 & fire7/relu\_expand1x1  & ReLU        &        &        &     & 192          & 16\x16     & 192          & 16\x16      &                     \\
  43 & fire7/expand3x3        & Convolution & 3\x3   & 1      & 1   & 64           & 16\x16     & 192          & 16\x16      &                     \\
  44 & fire7/relu\_expand3x3  & ReLU        &        &        &     & 192          & 16\x16     & 192          & 16\x16      &                     \\
  45 & fire7/concat           & Concat      &        &        &     & 384          & 16\x16     & 384          & 16\x16      &                     \\
  46 & fire8/squeeze3x3       & Convolution & 3\x3   & 2      & 1   & 384          & 16\x16     & 112          & 8\x8        &                     \\
  47 & fire8/relu\_squeeze3x3 & ReLU        &        &        &     & 112          & 8\x8       & 112          & 8\x8        &                     \\
  48 & fire8/expand1x1        & Convolution & 1\x1   & 1      & 0   & 112          & 8\x8       & 256          & 8\x8        &                     \\
  49 & fire8/relu\_expand1x1  & ReLU        &        &        &     & 256          & 8\x8       & 256          & 8\x8        &                     \\
  50 & fire8/expand3x3        & Convolution & 3\x3   & 1      & 1   & 112          & 8\x8       & 256          & 8\x8        &                     \\
  51 & fire8/relu\_expand3x3  & ReLU        &        &        &     & 256          & 8\x8       & 256          & 8\x8        &                     \\
  52 & fire8/concat           & Concat      &        &        &     & 512          & 8\x8       & 512          & 8\x8        &                     \\
  53 & fire9/squeeze1x1       & Convolution & 1\x1   & 1      & 0   & 512          & 8\x8       & 112          & 8\x8        &                     \\
  54 & fire9/relu\_squeeze1x1 & ReLU        &        &        &     & 112          & 8\x8       & 112          & 8\x8        &                     \\
  55 & fire9/expand1x1        & Convolution & 1\x1   & 1      & 0   & 112          & 8\x8       & 368          & 8\x8        &                     \\
  56 & fire9/relu\_expand1x1  & ReLU        &        &        &     & 368          & 8\x8       & 368          & 8\x8        &                     \\
  57 & fire9/expand3x3        & Convolution & 3\x3   & 1      & 1   & 112          & 8\x8       & 368          & 8\x8        &                     \\
  58 & fire9/relu\_expand3x3  & ReLU        &        &        &     & 368          & 8\x8       & 368          & 8\x8        &                     \\
  59 & fire9/concat           & Concat      &        &        &     & 736          & 8\x8       & 736          & 8\x8        &                     \\
  60 & drop9                  & Dropout     &        &        &     & 736          & 8\x8       & 736          & 8\x8        & p = 0.5             \\
  61 & conv10/split1          & Convolution & 1\x1   & 1      & 0   & 736          & 8\x8       & 512          & 8\x8        &                     \\
  62 & conv10/split2          & Convolution & 1\x1   & 1      & 0   & 736          & 8\x8       & 512          & 8\x8        &                     \\
  63 & conv10                 & Concat      &        &        &     & 1024         & 8\x8       & 1024         & 8\x8        &                     \\
  64 & pool10                 & Pooling     & 8\x8   &        &     & 1024         & 8\x8       & 1024         & 1\x1        & global avg. pooling \\
  65 & loss                   & Softmax     &        &        &     & 1024         & 1\x1       & 1024         & 1\x1        &                     \\ \bottomrule
  \end{tabular}
}
\end{table}

\clearpage
\section{Tips and Trick for the Training of CNNs}
\label{sec:appendix-training-howto}

The following section gives an overview of the training process with DIGITS v3.4 and NVidia's \caffe fork v0.14 \cite{digits}. The subsequent section lists a number of tips and tricks for the successful training of Convolutional Neural Networks.

\vspace{-1.6mm}\subsubsection{Training with \caffe and DIGITS}
\label{sec:training-howto}

\paragraph{Dataset Preparation} The first step before training is the creation of a \emph{dataset}. For the standard datasets \emph{MNIST} (28\x28 grayscale hand-written digits) and \emph{CIFAR} (32\x32 color images in 10 or 100 classes), a download script is provided with DIGITS. Other datasets have to be provided by the user. For datasets that are structured using separate subfolders for each class, DIGITS automatically recognizes the classes. In the case of ImageNet, the training and validation data has to be downloaded from the ILSVRC 2012 website~\cite{imagenet-2012}. The dataset has not changed since 2012 and consists of \SI{138}{\giga\byte} training and \SI{6}{\giga\byte} validation images. Additionally, the files \texttt{train.txt} and \texttt{val.txt} containing the mapping from image names to class numbers, as well as \texttt{synset\_words.txt} with the mapping from class numbers to class names, are needed and can be downloaded using a tool supplied with \caffe~\cite{caffe-aux}.

\paragraph{Dataset Creation} In DIGITS, a new dataset is created by choosing\ \texttt{Datasets > New Dataset Images > Classification}. Subfolder-structured datasets can be created by pointing to the root folder and choosing the percentage of images which should be used as validation images (the default of \SI{25}{\%} is reasonable in many cases). For other datasets, the paths to the image folders are set individually, and the \texttt{train.txt} and \texttt{val.txt} text files are uploaded. It is important to set the option \texttt{Shuffle Lines} to \texttt{Yes} to randomize the training set, and to upload \texttt{synset\_words.txt} under \texttt{Labels}. The image size can be chosen freely,\footnote{Most CNNs use 256\x256 pixel images, with random 224 or 227 pixel crops during training. Inception networks use 299 pixel crops and therefore need larger training images.}
the transformation type should be set to \texttt{half crop, half fill} for best results. LMDB is the default backend and allows fast image fetching during training. JPEG image compression slightly increases the runtime and completely loads the CPU during training, but significantly reduces the database size on disk.\footnote{An ImageNet dataset with size 256\x256 pixel images and JPEG compression occupies \SI{43}{\giga\byte} disk space, the same dataset with 128\x128 pixel images and lossless PNG compression \SI{51}{\giga\byte}.}

\paragraph{Data Augmentation} Especially for small datasets (where CNNs have a high risk of overfitting), data augmentation can improve the quality of results by creating additional artificial training samples either on-the-fly or during dataset creation. DIGITS and \caffe do not natively support on-the-fly augmentation yet~\cite{digits-augmentation}, but for this project, we added basic data augmentation support during dataset creation to DIGITS based on patch~\cite{digits-augmentation-patch}. The user can create multiple copies of each training image and apply random rotations, hue modulations, contrast modulations and translations with a chosen probability and in a chosen modulation range.

\paragraph{Model Definition} In DIGITS, a new CNN model is created by choosing \texttt{Models > New Model Images > Classification}.
There is a choice of three preset networks (LeNet, AlexNet and GoogleNet), which can be adapted, as well as the option to enter a custom network definition in \caffe\ \texttt{.prototxt} or Torch format.\footnote{
   Many additional CNN models can be found in the \caffe Model Zoo~\cite{model-zoo} and on Github~\cite{resnet-model,inceptionv3-model,inceptionv4-model}.}
DIGITS uses a custom fork of \caffe for training, which supports all layer types defined in \texttt{caffe.proto}~\cite{caffe-proto} and is mostly compatible with the official \caffe.\footnote{
   One known incompatibility concerns Batch Normalization layers, which use a different syntax and rely on different libraries underneath.}
However, most models require slight adaptations of their \emph{data}, \emph{softmax} and \emph{accuracy} layers to match the DIGITS style (refer to the given networks for examples).
Networks using {custom layer types}, such as \emph{Highway Networks}~\cite{highway-networks}, even require a recompilation of the underlying \caffe binaries.
After entering the network description and selecting the previously created dataset, a crop size $C$ may be specified.
This causes a random $C$\x$C$ pixel crop of each example image to be used during training, which helps the model to develop translational invariance.
During testing and inference, the $C$\x$C$ center-crop is used.
Typically the mean of all pixels in the training set (the so-called ``mean pixel'') is subtracted from input images to help with training convergence.\footnote{The helpfulness of mean subtraction is being debated in~\cite{digits-augmentation}. The researchers come to the conclusion that mean image subtraction is seldomly useful, and often one can even omit mean pixel subtraction.}
In addition to these settings, DIGITS allows a pre-trained model (\texttt{.caffemodel} file) to be specified for fine-tuning instead of training the CNN model from scratch.

\paragraph{Solver Configuration} With the dataset and the CNN model fully specified, only the \emph{Solver} is left to be configured.
Unfortunately, there exist no unique valid settings, and both model performance and training convergence are highly dependent on these \emph{hyperparameters}.
Besides the choice of the \emph{Base Learning Rate} and the \emph{Batch Size}, the \emph{Solver Algorithm}, the \emph{Learning Rate Schedule}, and the \emph{Number of Training Epochs} can be changed.
The duration of a training run is directly proportional to the \emph{Number of Training Epochs}. Shorter trainings are usually welcome and allow more experiments to be made, but longer training runs usually converge to slightly more ideal solutions.
The \emph{Learning Rate} also strongly influences how well trainings converge, by scaling the weight updates made in each training step.
If the learning rate is chosen too low, the training converges quickly, but to a suboptimal solution. And if the learning rate is set too high, the training may diverge.
The learning rate starts at the \emph{Base Learning Rate} and is then annealed over time according to the \emph{Learning Rate Policy}.
The default \emph{Solver Algorithm} is \emph{Stochastic Gradient Descent} (SGD), but other solver types are also available.
The \emph{Batch Size} determines how many training examples are pushed to the GPU at once.
Larger Batch Sizes results in faster training, but may outgrow the available graphics memory.
The final settings are the optional \emph{Random Seed}, which enables reproducible weight initializations, as well as the \emph{Snapshot and Validation Intervals}.

\paragraph{Training Launch} DIGITS makes multi-GPU training as simple as selecting the desired number of GPUs to be used for the job.
The scheduler then queues the task and waits until enough GPUs become available.
Once the job transitions from \emph{waiting} to \emph{running}, an estimate for the remaining time is calculated.
The training can be monitored in the progress chart, which tracks the CNN's loss and accuracy for both the training and the validation set.
For a training from scratch, the loss curve should start decreasing within the first epoch, otherwise the learning rate was probably set too high.
If the validation accuracy starts significantly drifting away from the training accuracy, the CNN model is overfitting and data augmentation or increased model regularization should be considered.

\vspace{-1.6mm}
\subsubsection{Training Tips and Tricks}
\label{sec:training-tips}

The successful training of CNNs requires persistence, good intuition and experience. The following rules of thumb worked for most of our experiments, which mainly consisted of the training of SqueezeNet variants on ImageNet with 128\x128 or 256\x256 pixel crops using one or two NVidia GeForce GTX Titan X GPUs:

\begin{description}
  \item[Batch Size] Choose the maximum Batch Size that still fits onto the GPU to speed up the training significantly. Batch Sizes are often chosen as powers of 2 to fit well onto the GPU's CUDA cores, but multiples of 32 seem to work just as well. The Batch Size further influences training convergence, because weight updates are deferred and averaged over each batch. When multiplying the Batch Size by a factor of $k$, the Base Learning Rate should also be changed by a factor of $\sqrt{k}$ (although a factor of $k$ usually works just as well)~\cite{alex-parallelizing-cnns}.

  \item[Learning Rate and Batch Size Sweeps] The ideal Learning Rate depends on the Batch Size, the dataset, as well as the CNN model. Base Learning Rates typically lie between 0.0001 and 0.01 when training from scratch, and even smaller learning rates may be used for finetuning. The \texttt{solver.prototxt} file supplied with most pre-trained networks can give a hint, otherwise a trial and error approach with a geometric series works best. DIGITS accepts lists in the format \texttt{[0.001, 0.002, 0.004]} for Batch Size and Base Learning Rate and automatically generates jobs for each permutation.

  \item[Learning Rate Policies] As mentioned, the learning rate is usually lowered over time during the training, which helps the optimization to settle into an optimum. The classic approach steps down the learning rate every few epochs by a fixed factor $k$. Other approaches include exponential decay, sigmoidal decay and polynomial decay. Mishkin et al. thoroughly explored many optimizations and found a \emph{Linear Learning Rate Policy} (polynomial decay with power 1) to work best for AlexNet-like CNNs~\cite{ducha-aiki}.

  \item[Solver Types] DIGITS supports a number of different optimization algorithms, including Nesterov's Accelerated Gradient (NAG), Adaptive Gradient (AdaGrad), RMSprop, AdaDelta, and Adam which should in theory all lead to a faster training. These algorithms improve convergence for example by adding momentum to the optimization steps, or by adaptively tuning the learning rate for each individual weight (see \cite{cs231n-learning-schedule} for details and a beautiful illustration). Despite their appeal, these solvers require a new trial-and-error hyperparameter search and quick tests led to inferior optimization results in our case. Therefore, we used the basic Stochastic (Mini-Batch) Gradient Descent algorithm in this project.

  \item[Batch Normalization] ResNet and all Inception variants use Batch Normalization Layers. Unfortunately, the implementation and syntax of BN layers differs between the original \caffe and the NVidia fork, as well as between CPU and GPU-based computation. See \cite{batchnorm-tip0,batchnorm-tip1,batchnorm-tip2} for hints if you want to experiment with BN layers.

  \item[Cloning existing Jobs] The fastest way to create a new design iteration is to clone a previous job, which copies the network description and all previous settings into a new job, ready for customization. The list of ``previous networks'' then optionally allows the pre-trained network weights to be loaded as initialization or for finetuning.

  \item[Running Time] There are two peculiarities with regard to the running time. First, when starting a new job, the estimated time remaing is initially very high and completely wrong. The value takes a few minutes to settle to a realistic estimate. Second, when looking for the runtime of an active or completed job, the correct value is found in the job page under \texttt{Job Status > Train Caffe Model > Running}. All other values (including the \texttt{runtime} field in table \texttt{Details}) wrongly add the job's \emph{waiting} time.

\end{description}




\chapter{FPGA Accelerator Details}

\section{Analysis of the Pipeline Flushing Issue}
\label{sec:appendix-slowdown-calculation}


\begin{table}[H]
  \makebox[\textwidth][c]{ 
    \begin{minipage}{1.1\linewidth} 

    \caption[Analysis of the Pipeline Flushing Issue]{Calculation of the Slow-Down Factor caused by the Pipeline Flushing Issue in Vivado HLS 2016.2.}
    \label{tab:slow-down-calculation-flushing-issue}
      \includegraphics[width=\linewidth]{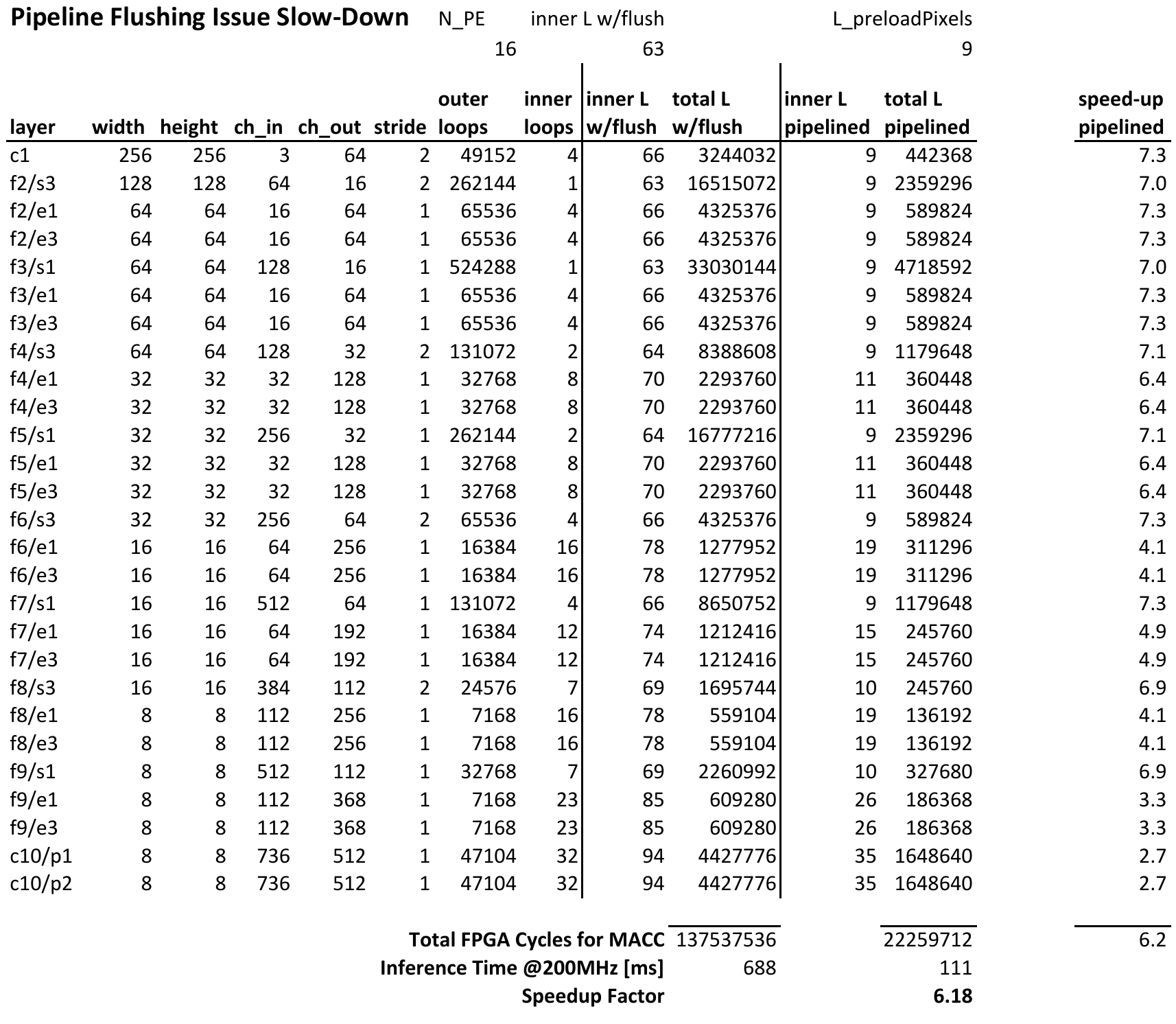}\\
    \end{minipage}
  }
\end{table}


\section{Detailed Block Diagram for the ZynqNet FPGA Accelerator}
\label{sec:appendix-block-diagram}
\begin{figure}[H]
  \centering
  \includegraphics[angle=90,width=\linewidth]{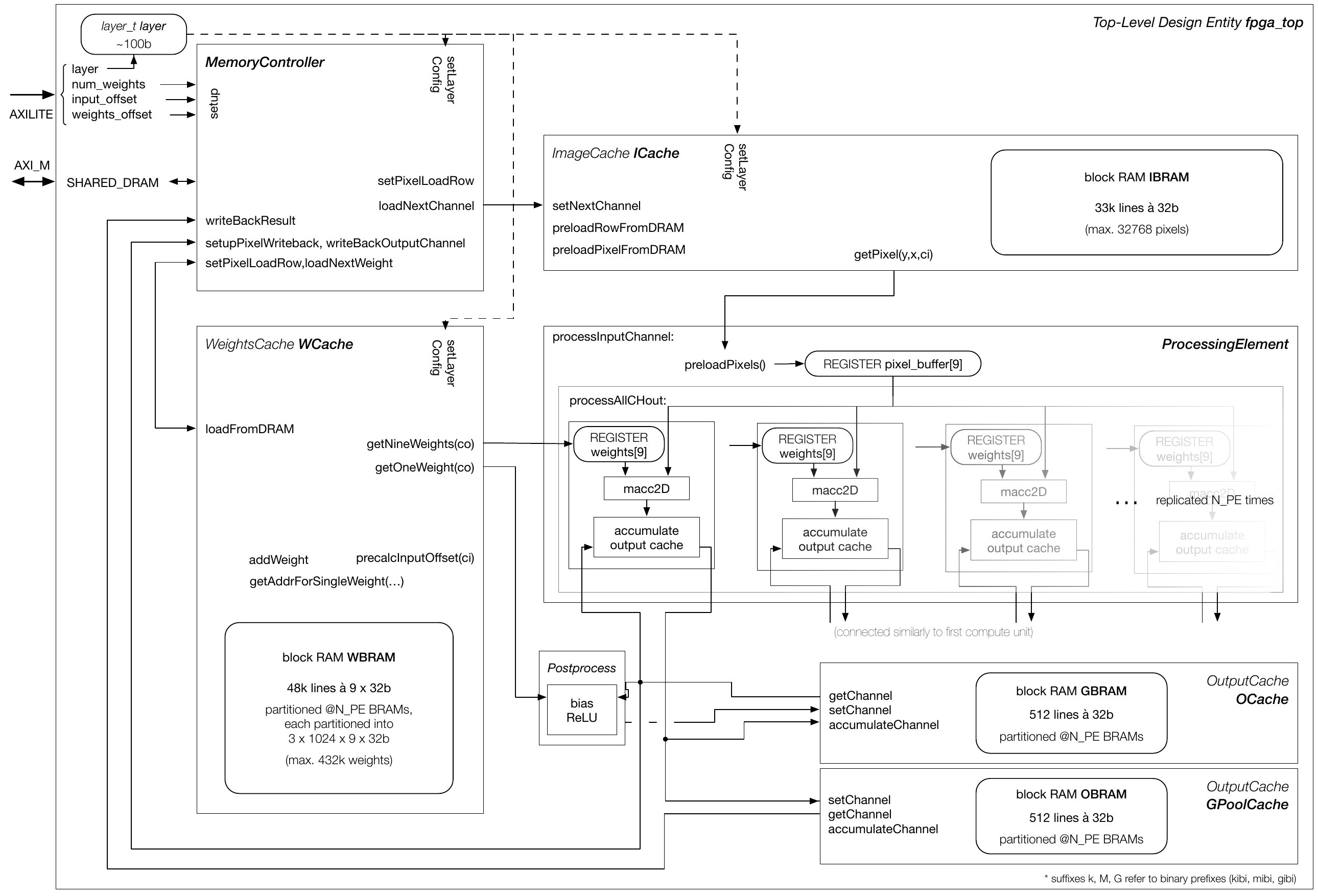}
  \caption[Detailed Block Diagram for the ZynqNet FPGA Accelerator]{Detailed Block Diagram for the ZynqNet FPGA Accelerator, including actual Cache Sizes and References to the \cpp Software Implementation.}
  \label{fig:detail-blockdiagram}.
\end{figure}

\end{appendices}

{%
\setstretch{1.1}
\renewcommand{\bibfont}{\normalfont\small}
\setlength{\biblabelsep}{0pt}
\setlength{\bibitemsep}{0.5\baselineskip plus 0.5\baselineskip}
\printbibliography[nottype=online]
\printbibliography[heading=subbibliography,title={Websites},type=online,prefixnumbers={@}]
}

\end{document}